%% file: main.tex
\pdfoutput=1
\documentclass[sigconf,nonacm]{acmart}
\usepackage{xspace}
\usepackage{enumitem}

\usepackage{bbm}
\usepackage{xcolor}
\usepackage{cleveref}
\usepackage[normalem]{ulem}
\usepackage{subcaption}
\usepackage{mathtools}

\def\shapley{Shapley-MC\xspace}
\def\kernel{Shapley-KS\xspace}
\def\single{Single Column Ablation\xspace}
\def\mask{Mask Learning\xspace}
\def\datatt{$\text{dat}(\mathcal{S})$\xspace}
\def\datatteq{\text{dat}(\mathcal{S})}

\newenvironment{repeatresult}[2]
{\vskip0.5em\par\textsc{#1} #2.\em}
{\vskip1em}

\newenvironment{reptheorem}[1]{\begin{repeatresult}{Theorem}{#1}}{\end{repeatresult}}

\AtBeginDocument{%
  }

\begin{document}

\title{A Comparative Study of Importance Attribution to Columns \\ in Deep Learning over Relational Databases}

\author{Hila Levi-Yosefi}
\affiliation{%
  \institution{Technion}
  \city{Haifa}
  \country{Israel}
}
\email{hilalevi@campus.technion.ac.il}

\author{Shunit Agmon}
\affiliation{%
  \institution{Technion}
  \city{Haifa}
  \country{Israel}
}
\email{shunita@campus.technion.ac.il}

\author{Ester Livshits}
\orcid{0000-0003-3485-9887}
\affiliation{%
  \institution{Technion}
  \city{Haifa}
  \country{Israel}
}
\email{esterliv@technion.ac.il}

\author{Agapi Rissaki}
\affiliation{%
  \institution{Northeastern University}
  \city{Boston}
  \country{USA}
}
\email{rissaki.a@northeastern.edu}

\author{Benny Kimelfeld}
\affiliation{%
  \institution{Technion}
  \city{Haifa}
  \country{Israel}
}
\email{bennyk@cs.technion.ac.il}

\renewcommand{\shortauthors}{Levi-Yosefi et al.}

\begin{abstract}
Deep learning over relational databases has led to increasingly effective prediction models, motivating a growing need to explain their behavior. We study explanation methods that assign importance scores to individual database attributes, focusing on a fundamental question: to what extent does the choice of importance-attribution method matter? We develop a methodology to systematically compare importance-attribution methods and analyze their similarities and differences. In particular, we study two aspects of the explanations: which attributes they deem important, captured by differences in attribute rankings, and how importance scores vary across attributes, captured by differences in score distributions. We further investigate whether these differences are reflected in the conventional measure of explanation quality, fidelity.

We apply our methodology in an extensive empirical study of four methods: Monte Carlo Shapley, Kernel SHAP, single-attribute impact, and mask learning. We experiment with multiple datasets and tasks using two RDL architectures: heterogeneous GraphSAGE and Relational Graph Transformers. Our experiments show that explanation methods behave differently depending on the prediction type: regression or classification. In regression, most methods exhibit high agreement in attribute rankings, whereas classification yields considerably less consensus. Notably, while mask learning frequently diverges in its importance distributions, Shapley-based methods and single-attribute impact often converge to similar explanations. This suggests that, in many settings, single-attribute impact can replace Shapley at substantially lower computational cost. Overall, our study provides methodological tools for comparing importance-attribution methods, and empirical evidence that their choice has important consequences for interpreting relational deep learning models.
\end{abstract}

\maketitle
\pagestyle{plain}

\input{introduction}
\input{related}

\input{preliminaries}

\input{methods}
\input{experiments}
\input{conclusions}

\clearpage

\bibliographystyle{ACM-Reference-Format}
\bibliography{references}

\newpage
\appendix

\input{appendix_methods}

\input{appendix_plots}
\input{appendix_rt}

\end{document}

%% file: introduction.tex
\section{Introduction}
Modern machine learning (ML) is applied to relational databases through various approaches. One approach represents a relational database as a graph, where nodes represent entities (tuples) and edges capture their relationships (typically induced by foreign-key constraints); then, graph-based deep learning models (e.g., graph neural networks) are trained directly on this representation~\cite{DBLP:conf/sigmod/CappuzzoPT20, DBLP:conf/icde/TonshoffFGK23, fey2024position, 4dbinfer_wang2024, cvitkovic2020supervised,DBLP:conf/icml/WangWGWYWZ25}. A second approach designs neural architectures that operate directly over the database, explicitly modeling tuples, attributes, and their relationships; examples include Relational Transformers~\cite{DBLP:journals/corr/abs-2510-06377}, KumoRFM-2~\cite{hudovernik2026kumorfm2scalingfoundationmodels}, and neuro-relational programs~\cite{DBLP:journals/corr/abs-2605-24207,DBLP:journals/corr/abs-2606-11946}. A third approach transforms the database into a single flat table, using techniques such as deep feature synthesis~\cite{DBLP:conf/dsaa/KanterV15} and the ACORA architecture~\cite{DBLP:journals/ml/PerlichP06}, and applies a model designed for tabular data~\cite{grinsztajn2026tabpfn3technicalreport,Meyer2026RelationalIL}, such as XGBoost~\cite{DBLP:journals/corr/ChenG16} and TabPFN~\cite{Hollmann2022TabPFNAT}. 

Importantly, the above approaches rely on modeling complex interactions among database components (e.g., records and cells), often using deep learning architectures that incorporate mechanisms like attention and transformers. While these interactions allow the models to exploit information distributed across the relational structure, they also make it difficult to understand how individual components of the database contribute to a prediction.
To this end, the field of \emph{Explainable Artificial Intelligence} (XAI) studies methods for clarifying the behavior of complex models~\cite{DBLP:journals/widm/AngelovSJAA21, marques2023logic, molnar2020interpretable,guidotti2019survey,samek2019explainable}. Explanations can be \emph{local}, seeking to explain a prediction for a particular input, or \emph{global}, seeking to describe the behavior of the model more broadly. They may also differ in the type of information they provide. \emph{Contrastive} (or \emph{counterfactual}) explanations identify changes to the input that would lead to a different outcome~\cite{barcelo2025explaining}, while \emph{abductive} explanations identify a restricted part of the input that is sufficient to support the observed outcome~\cite{marques2023logic}.

We present an experimental study of a particular class of global explanations---those that assign an \emph{importance score} to each data attribute (i.e., column) in a relational database, quantifying its contribution to the model's predictions~\cite{rissaki2025database}. Such explanations can help data owners understand what the model relies on when making predictions, which parts of the data are most influential in decisions concerning individuals, and whether unexpectedly high importance assigned to certain attributes may reveal data leakage or other artifacts in the dataset's construction.
This setting differs fundamentally from feature-importance analysis for single-table models~\cite{breiman2001_pfi,lundberg2017unified,fisher2019_pfi}. In the latter, each column typically corresponds directly to a feature of the target instance being predicted. In a relational database, by contrast, an attribute may belong to a relation that is several foreign-key hops away from the target relation. Its values may thus influence a target prediction only indirectly, through the relational structure and the model's aggregation mechanisms. Moreover, the same attribute value may contribute to the predictions for one, several, or many target instances. Consequently, attributing importance to database columns sheds light not only on the predictive effect of a feature, but also on how information propagates and is reused across the relational structure.

\emph{But how should the importance of an attribute be measured?} There are multiple approaches to this task, drawing on techniques developed in both ML and databases, each embodying a different notion of what it means for an attribute to be important~\cite{covert2021explaining}. Since all of these methods aim to quantify an attribute's contribution to the model's predictions, one might expect them to agree, at least broadly, on which attributes matter and how much. If, however, well-founded methods produce substantially different insights, then the choice of explanation method is not merely a technical detail: the same model may admit conflicting accounts of what drives its predictions, depending on the explanation method of choice, potentially leading to different operational or scientific conclusions~\cite{krishna2022disagreement}. Our study examines this question. We compare explanation methods along three dimensions: whether they induce similar \emph{rankings} of the attributes, whether they produce similar \emph{distributions of importance scores}, and how these differences relate to the conventional notion of \emph{fidelity}~\cite{yuan2022explainability, amara2022graphframex, longa2025explaining}, which evaluates the effect of retaining or removing (or masking) database attributes. 

We study four explanation methods that embody different notions of importance. The first, \emph{single-column ablation}, is a direct instantiation of 
removal-based explanations~\cite{breiman2001_pfi,covert2021explaining}, and quantifies the impact of an attribute by masking its column and quantifying the resulting change in the model's predictions. Two more methods are based on the \emph{Shapley value}~\cite{Shapley}---a traditional contribution measure from cooperative game theory that has been applied in a plethora of domains, including ML~\cite{rozemberczki2022shapley} and databases~\cite{DBLP:journals/sigmod/BertossiKLM23}. We show that the exact Shapley value is computationally intractable even for extremely limited architectures (\Cref{thm:shapley_hardness}), and employ two distinct approximation strategies:
Monte Carlo, which is the direct sampling approach to estimating the Shapley value's expectation~\cite{rozemberczki2022shapley}, and
\emph{Kernel SHAP}~\cite{lundberg2017unified}, which estimates the Shapley values through a weighted linear surrogate model fitted over sampled coalitions. The last method is \emph{Mask Learning}, a recently proposed approach for relational deep learning that optimizes a sparse mask over the database attributes so as to preserve the model's predictions while retaining only the most important attributes~\cite{rissaki2025database}.

Our study considers two graph-based relational deep-learning architectures: \emph{Heterogeneous GraphSAGE} (HGS)~\cite{hamilton2017inductive, fey2019fast} and the \emph{Relational Graph Transformer} (RelGT)~\cite{dwivedi2025relational}. 
We also validate our findings on a third, non-graph architecture, the \emph{Relational Transformer} (RT)~\cite{ranjan2026rtjlargescalepretrainingrelational}, on the removal-based explainers. (Results are provided in ~\Cref{app:rt}.)
Using these architectures as experimental testbeds, we develop quantitative analyses and visualizations to characterize agreement among explanation methods along two dimensions: the rankings they induce over attributes and the distributions of the importance scores they assign (\Cref{sec:experiments}).

Our study leads to several conclusions. First, addressing the concern that different explanation methods might yield conflicting insights, we find that the extent and pattern of agreement are dominated by the nature of the predictive task rather than the underlying relational deep-learning architecture. In regression tasks, most methods exhibit high agreement, isolating a tight core of critical attributes. In contrast, explanations for classification tasks exhibit much more disagreement between the methods. 

Within this task-driven context, we observe consistent behavioral patterns among the explanation methods themselves. Single-column ablation and the two Shapley-based approximations (Monte Carlo and Kernel SHAP) often converge to highly similar explanations, exhibiting strong alignment in both their attribute rankings and score distributions. This agreement suggests that practitioners can often replace Shapley approximations with single-column ablation, reducing computational cost without sacrificing explanation quality.
Mask Learning, on the other hand, frequently diverges from this consensus in how it distributes importance scores. This is likely because it aims to identify a \emph{set} of attributes that are jointly sufficient for the prediction by treating each attribute as either included or excluded. This contrasts with approaches that assign each attribute an individual contribution score to produce a fine-grained ranking.
This indicates that the choice of an explanation method should be carefully aligned with the specific notion of importance the practitioner seeks to capture.

These findings suggest that importance-attribution methods should not be regarded as interchangeable ways of recovering a single explanation. Rather, the choice of method and the notion of importance it captures should be made explicit when interpreting the resulting scores. When conclusions depend on attribute importance, examining more than one explanation method may provide useful evidence about the extent to which these conclusions are robust to this choice. Our findings also reveal that, when associating an explanation with a model, it is important to examine the properties of the explainers with respect to different scenarios.

\subsubsection*{Contributions.}
In summary, our contributions are as follows:
\begin{enumerate}[leftmargin=*]
\item We present a framework for comparing attribute-importance explanations in relational deep learning, and study four representative methods within this framework: single-column ablation, Monte Carlo Shapley, Kernel SHAP, and mask learning. 

\item We develop methodologies and visualizations for systematically comparing importance attribution methods along two complementary dimensions: the rankings they induce over database attributes, and the distributions of the scores they assign. We further examine how agreement along these dimensions relates to the conventional fidelity-based evaluation of explanations.

\item We conduct an extensive empirical study of the four methods across multiple datasets and prediction tasks, using two graph-based
architectures for relational deep learning over databases, Heterogeneous GraphSAGE and Relational Graph Transformer, alongside a complementary evaluation of the removal-based explainers on the Relational Transformer architecture.

\item We observe that an explainer's behavior depends heavily on the predictive task type (e.g., regression versus classification) and its underlying mathematical notion of importance (e.g., individual attribute ranking versus collective subset sufficiency). These results motivate practitioners to carefully consider the explanation's objective and the nature of the predictive task when choosing a method.

\end{enumerate}

\subsubsection*{Organization.}
The remainder of the paper is organized as follows. We review related work in \Cref{sec:related}. \Cref{sec:preliminaries} introduces our relational database setting, prediction models, and attribute-importance explanations. \Cref{sec:methods} presents the four explanation methods considered in our study and establishes the computational intractability of the exact Shapley value. \Cref{sec:experiments} describes our experimental setup and presents our analysis of attribute rankings, importance-score distributions, and fidelity. We conclude in \Cref{sec:conclusions}.

%% file: related.tex
\section{Related Work}\label{sec:related}

\paragraph{Feature-importance explanations}

We focus on global feature importance methods, which assign an importance score to each attribute, summarizing its influence across all predictions rather than on a single instance. A direct realization is Permutation Feature Importance (PFI)~\cite{breiman2001_pfi, fisher2019_pfi, debeer2020cpi, molnar2024conditional_subgroups}, which breaks the association between an attribute and the prediction by shuffling its values, and reports the performance degradation, evaluating each attribute on its own and thus ignoring feature interaction. The Shapley value~\cite{lundberg2017unified, rozemberczki2022shapley} addresses this limitation by evaluating an attribute's marginal effect across every coalition of the others, at a computational cost exponential in the number of attributes. Widely used approximations are employed in practice: Kernel SHAP~\cite{lundberg2017unified} recovers the values as coefficients of a weighted linear regression over sampled coalitions, while Monte Carlo estimation samples permutations and averages marginal contributions~\cite{rozemberczki2022shapley}. We experiment with both of these approximations, which are discussed in detail in \Cref{sec:methods}.

These methods belong to a broader family of removal-based explanations~\cite{covert2021explaining}, which simulate the absence of an attribute and observe the effect, and which differ in how the attribute is removed, which aspect of model behavior is examined, and how the effects are summarized. LIME~\cite{ribeiro2016should_lime} is another prominent method in this family; however, it is inherently local, fitting a surrogate to a neighborhood of a single instance. 
Another line of work quantifies importance through the gradient of the output with respect to the input, with integrated gradients~\cite{sundararajan2017axiomatic_IG} being the standard representative. Such methods are not well suited for tabular and relational data because the feature space is heterogeneous, i.e., categorical features, timestamps, and text do not belong to a real-valued differentiable space. 

Finally, importance may be expressed as membership in a subset rather than as a score, identifying a small set of attributes that alone suffice to preserve the model's output. L2X~\cite{chen2018learning} learns such a selection through an information-theoretic objective, while an alternative is to learn a sparse mask directly~\cite{fong2017masking, ying2019gnnexplainer}. The latter has been adapted to relational deep learning~\cite{rissaki2025database}, which is to our knowledge the only explanation method developed specifically for relational deep learning, which we include alongside the three removal-based methods above.

\paragraph{Explanations in databases}

Database research has studied explanations as well, where the goal is to explain a query result with explanations found in the input or specific aspects of computation~\cite{glavic2021trends}. Different approaches explain the presence or absence of specific tuples in query results~\cite{whynot}, outliers, trends, and bias in query outputs~\cite{miao2019cape_outliers, wu2013scorpion_outliers, roysuciu2014, Salimi23}, and performance anomalies~\cite{kalmegh2016iQCAR, zhang2017exstream}. Various techniques have been employed, including provenance~\cite{Glavic21, greentannen2017semiring}, targeted interventions on the input~\cite{wu2013scorpion_outliers, Roy15a}, and concise summaries~\cite{gebaly2018explanation_tables, Lee20}. For a comprehensive survey, see~\cite{glavic2021trends}.

In parallel to the feature importance literature, the Shapley value has been applied for database explanations, for instance to quantify the contribution of a fact \cite{Livshits20, deutch2022computing} or a parameter \cite{gilad2024importance} to a query result. The complexity landscape has been studied extensively: for conjunctive queries without self-joins the exact computation admits a dichotomy separating polynomial-time from \#P-hard cases \cite{Livshits20}, and tractability is governed by the complexity of probabilistic query evaluation \cite{deutch2022computing}. On the machine learning side the analogous question has been tackled for classes of models rather than classes of queries, where exact computation is \#P-hard over the empirical distribution and already for logistic regression under fully-factorized distributions \cite{vandenbroeck2021}, and remains tractable only under restricted model structure \cite{arenas2022}. Our setting differs in that the players are attributes rather than facts and the game is a learned model rather than a query, and the payoff aggregates disagreement over an \emph{entire} evaluation set. In~\Cref{app:shapley_hardness},
we show that computing the exact Shapley value in our setting is \#P-hard even for a linear model over a single-table database, and remains so when each table carries a bounded number of attributes. 

\paragraph{Evaluating and comparing explanations}

XAI literature has documented multiple modes of failure for explanations, e.g.,~attributions may be insensitive to randomization of the model's parameters or training data~\cite{adebayo2018sanity} or biased by the distribution shift caused by feature ablation~\cite{hooker2019roar}. These findings justify the need to systematically evaluate explanations under common metrics~\cite{agarwal2022openxai, hedstrom2023quantus, bhatt2020evaluating}. Among them are the fidelity measures we adopt, which assess whether the attributes an explanation deems as \emph{important} are \emph{sufficient} to preserve the prediction and whether they are \emph{necessary} for it~\cite{yuan2022explainability, amara2022graphframex, longa2025explaining}. 
The two are distinct properties, and a method optimizing one need not perform well on the other. Testing methods individually leaves open whether they agree with each other. Comparing explanations for tabular models has found that well-motivated explanation approaches often disagree on which features are important, with no principled basis for preferring one ranking over another~\cite{krishna2022disagreement}. We answer the same question for the case of relational deep learning.

%% file: preliminaries.tex
\section{Preliminaries}
\label{sec:preliminaries}

We first establish the formal foundations of our framework.

\subsubsection*{Relational databases.} 
A database schema $\mathcal{S}$ is a collection of relation names. Each relation name $R$ in $\mathcal{S}$ is associated with a sequence $\text{att}(R) = (A_1, \dots, A_k)$ of attributes, together with a designated sequence $\text{key}(R)$ of primary key attributes from $\text{att}(R)$.
Each attribute $A_i$ has an associated domain $\text{dom}(A_i)$. A tuple $t$ over $R$ is a sequence $(c_1, \dots, c_k)$ of values such that
$c_i \in \text{dom}(A_i)$ for each $i \in \{1, \dots, k\}$.
We write $t[A_i] = c_i$ for the value assigned to attribute $A_i$, and,
for a sequence $X = (A_{j_1}, \dots, A_{j_\ell})$ of attributes, we write
$t[X] = (c_{j_1}, \dots, c_{j_\ell})$ for the projection of $t$ onto $X$. An instance $r$ over $R$ is a finite set of tuples over $R$ such that
$t[\text{key}(R)] \neq t'[\text{key}(R)]$ for all distinct
tuples $t,t' \in r$.

Furthermore, a database schema $\mathcal{S}$ contains a set of foreign-key (FK)
dependencies, denoted by $\text{FK}(\mathcal{S})$. An FK dependency from a
relation $R$ to a relation $R'$ is of the form
$R[X] \sqsubseteq \text{key}(R')$, where $X$ is a sequence of
attributes of $R$ of the same length as $\text{key}(R')$.
For instances $r$ over $R$ and $r'$ over $R'$, the dependency is satisfied if,
for every $t \in r$, there exists $t' \in r'$ such that
$t[X] = t'[\text{key}(R')]$.

A database $D$ over a schema $\mathcal{S}$ assigns to each relation name
$R$ in $\mathcal{S}$ an instance $D(R)$ over $R$. The database $D$ is valid
if, for every FK dependency
$R[X] \sqsubseteq \operatorname{key}(R')$ in $\operatorname{FK}(\mathcal{S})$,
the instances $D(R)$ and $D(R')$ satisfy the dependency.

\subsubsection*{Prediction model.} 
Formally, let $\mathcal{D}(\mathcal{S})$ denote the space of all valid databases over the schema $\mathcal{S}$, let $R$ be a relation name of $\mathcal{S}$, and let $Y$ be the target variable domain (e.g., $Y = \mathbb{R}$ for regression, or a discrete set of classes for classification).
A \emph{prediction model} $f$ is a function that takes a database $D \in \mathcal{D}(\mathcal{S})$ and any tuple $t \in D(R)$ as input, and maps them to an estimated target value $f(t,D) \in Y$ based on the underlying data in $D$.

For the purpose of evaluation, we let $D_{\text{target}} \subseteq D(R)$ denote an \emph{evaluation set} containing specific target tuples for which a target value needs to be predicted and assessed against ground truth labels.

\subsubsection*{Attribute-importance explanations.}
In this work, we focus on a specific class of explanations known as \emph{attribute-importance explanations}~\cite{sundararajan2017axiomatic_IG}.
This form of explanation quantifies the contribution of individual attributes to the model's prediction by assigning a numeric score to each attribute. 
Structural keys (primary and foreign keys) are explicitly excluded from the explanations; since these keys dictate the structural topology of the relational schema, masking or perturbing them alters the underlying relational structure instead of modifying a data attribute value. Therefore, we focus on the remaining attributes, which we refer to henceforth as \emph{data attributes}. 
We denote this set of data attributes in all relations of the schema $\mathcal{S}$ by \datatt and its size by $N=|\datatteq|$.

In this paper, by an \emph{explanation method} we refer to a function $E$ that, given a prediction model $f$ and a database $D \in \mathcal{D}(\mathcal{S})$, assigns a numerical importance score to each attribute in \datatt. Formally, $E(f, D)$ maps the data attributes to an attribution function $W:\datatteq \rightarrow \mathbb{R}$ where $W(A_i)$ represents the estimated contribution of the attribute $A_i \in \datatteq$ to the model's prediction.

%% file: methods.tex
\section{Explanation Methods}\label{sec:methods}

In this section, we describe the explanation methods used in our evaluation. We selected four diverse methods originating from different interpretability frameworks. The methods we describe are fundamentally model-agnostic, and treat the underlying predictive model as a black box. Consequently, the proposed explanation framework is generalizable to any Relational Deep Learning (RDL) architecture that operates on structured, multi-table data.

\subsection{Single Column Ablation}
\label{sub:generic_ablation}
The \single explainer is a black-box framework grounded in the concept of \textit{Permutation Feature Importance} (PFI)~\cite{molnar2020interpretable}. PFI quantifies the importance of an input attribute by measuring the change in a model's prediction output after permuting the attribute's values, which effectively breaks the attribute's relationship with the target outcome. We generalize this principle to relational data in two ways. First, while standard PFI relies exclusively on shuffling values (permutation), we expand the ablation operator $\alpha$ to structurally support a variety of type-specific strategies, such as mean imputation and noise injection. 
Second, we measure the direct prediction sensitivity of the model consistently across both classification and regression tasks by quantifying the average magnitude of the shift in the continuous output (raw prediction probabilities for classification and target predictions for regression).

\subsubsection*{Formal mechanism}
For each attribute $A_i \in \datatteq$, we define an ablation operator $\alpha(A_i)$ that replaces the values of the column with a predefined baseline value. Let $D_{\text{target}}$ be the evaluation set, and let $D_{i,\alpha}$ be the modified version of the database where only the $i$-th attribute has been replaced by $\alpha(A_i)$.

We define the attribution function $W:\datatteq \rightarrow \mathbb{R}$ by setting the importance score $W(A_i)$ as the average absolute change in the model's prediction output across all evaluated target tuples:
\begin{equation*}
W(A_i) = \frac{1}{|D_{\text{target}}|} \sum_{t \in D_{\text{target}}} \left| f(t, D) - f(t_{i,\alpha}, D_{i,\alpha}) \right|
\end{equation*}
where $f$ is the prediction model, $t \in D_{\text{target}}$ represents an original target tuple, and $f(t, D)$ denotes the prediction for this tuple under the original database $D$.
Accordingly, let $t_{i,\alpha}$ denote the perturbed target tuple where the $i$-th attribute has been replaced by $\alpha(A_i)$.
In both task regimes, $W(A_i)$ quantifies the dataset-level prediction sensitivity, where a higher value indicates greater overall importance of attribute $A_i$ across the entire evaluation set.

\subsection{Shapley Value}
\label{sub:shapley_explanations}

The Shapley value is a concept originating from cooperative game theory, providing a unique and axiomatic solution for fair credit assignment among players (e.g.,~input attributes) contributing to a total payout (e.g.,~the model's prediction)~\cite{rozemberczki2022shapley, lundberg2017unified}. In the domain of relational databases, the Shapley value has been effectively employed to quantify the importance of specific database facts or query parameters~\cite{deutch2022computing, gilad2024importance,kimelfeld21query_games}.
We next describe an application of the Shapley value to attribute importance.

Formally, let \datatt be the set of $N$ data attributes as defined in \Cref{sec:preliminaries}. To capture the impact of these attributes on the model's predictions, we define a dataset-wide characteristic function $\nu(S): \mathcal{P}(\datatteq) \to \mathbb{R}$ over the evaluation set $D_{\text{target}}$.
For a given coalition $S \subseteq \datatteq$ of attributes, the attributes excluded from the coalition (i.e.,~$\datatteq \setminus S$) are systematically ablated across the database. 

For both classification and regression tasks, the global characteristic function $\nu(S)$ measures the global mean absolute prediction shift over all target tuples. Let $t \in D_{\text{target}}$ represent an individual target tuple, and let $f(t, D)$ be the prediction of the original black-box model $f$ for tuple $t$ given the original database $D$. Let $D_S$ denote the perturbed database generated by replacing all the attributes of $\datatteq \setminus S$ with their predefined baseline values. 
Accordingly, let $t_S$ denote the perturbed target tuple where the same excluded attributes have been replaced by their baseline values. The characteristic function $\nu(S)$ is formulated as:
\begin{equation}
\label{eq:shapley_characteristic}
\nu(S) = \frac{1}{|D_{\text{target}}|} \sum_{t \in D_{\text{target}}} \left|f(t, D) - f(t_S, D_S)\right|.
\end{equation}
Given this global characteristic function, we follow the definition of~\cite{rozemberczki2022shapley} for the global Shapley value $W(A_i)$ for a specific input attribute $A_i \in \datatteq$, which is formulated by averaging its marginal contribution over all possible permutations of the attributes:
\begin{equation*}
W(A_i) = \frac{1}{N!} \sum_{\pi \in \Pi} \left[ \nu(S_\pi^{(i)} \cup \{A_i\}) - \nu(S_\pi^{(i)}) \right]
\end{equation*}
where $\Pi$ is the set of all $N!$ permutations of the attributes in $\datatteq$, and $S_\pi^{(i)}$ represents the set of attributes that precede attribute $A_i$ in permutation $\pi$.
The resulting value $W(A_i)$ is the average marginal contribution of attribute $A_i$ to the model's global prediction shift.

\subsubsection*{Computational intractability}
While this approach offers a robust and theoretically sound foundation, naively calculating the Shapley value directly from its definition requires evaluating the characteristic function across all $N!$ possible permutations.
In fact, computing the exact Shapley value under our characteristic function $\nu$, which measures the global mean absolute prediction shift, is computationally
intractable even in highly restricted settings.

\def\hardnessshapley{Computing the Shapley value of an attribute under the characteristic function $\nu$ is \#P-hard already for a linear model over a single-relation schema. This also holds when the schema consists of multiple relations, all of bounded arity, and the model is a message-passing GNN with a single layer and no non-linear activations.}

\begin{theorem}\label{thm:shapley_hardness}
\hardnessshapley
\end{theorem}
The hardness result is also of independent interest beyond our specific setting: since the reduction applies to a single tuple in a single relation, it establishes \#P-hardness of exact local/global attribute-level Shapley computation under natural ablation-based characteristic functions, even for a linear model. 
The complete proof is in~\Cref{app:shapley_hardness}.

To overcome their intractability, we approximate the Shapley values via two methods: Monte Carlo permutation sampling (denoted as \shapley) and Kernel SHAP (denoted as \kernel).

\subsubsection{Monte Carlo Shapley (\shapley)}  
Since evaluating all $N!$ permutations is computationally prohibitive,
\shapley approximates the Shapley values by sampling $P$ random permutations and averaging the marginal contributions over them~\cite{deutch2022computing, rozemberczki2022shapley}: 
\begin{equation*}
W(A_i) \approx \frac{1}{P} \sum_{p=1}^{P} \left[ \nu(S_p^{(i)} \cup \{A_i\}) - \nu(S_p^{(i)}) \right]
\end{equation*}
where $S_p^{(i)}$ represents the set of attributes that precede attribute $A_i$ in the $p$-th random permutation.
By using our characteristic function $\nu$ combined with a unified baseline ablation strategy (\Cref{sub:explainer_configs}), this \shapley method provides a consistent and unbiased statistical estimation of the exact Shapley values.

\subsubsection*{Convergence and empirical termination} 
\begin{figure}[t]
    \centering    \includegraphics[width=\columnwidth]{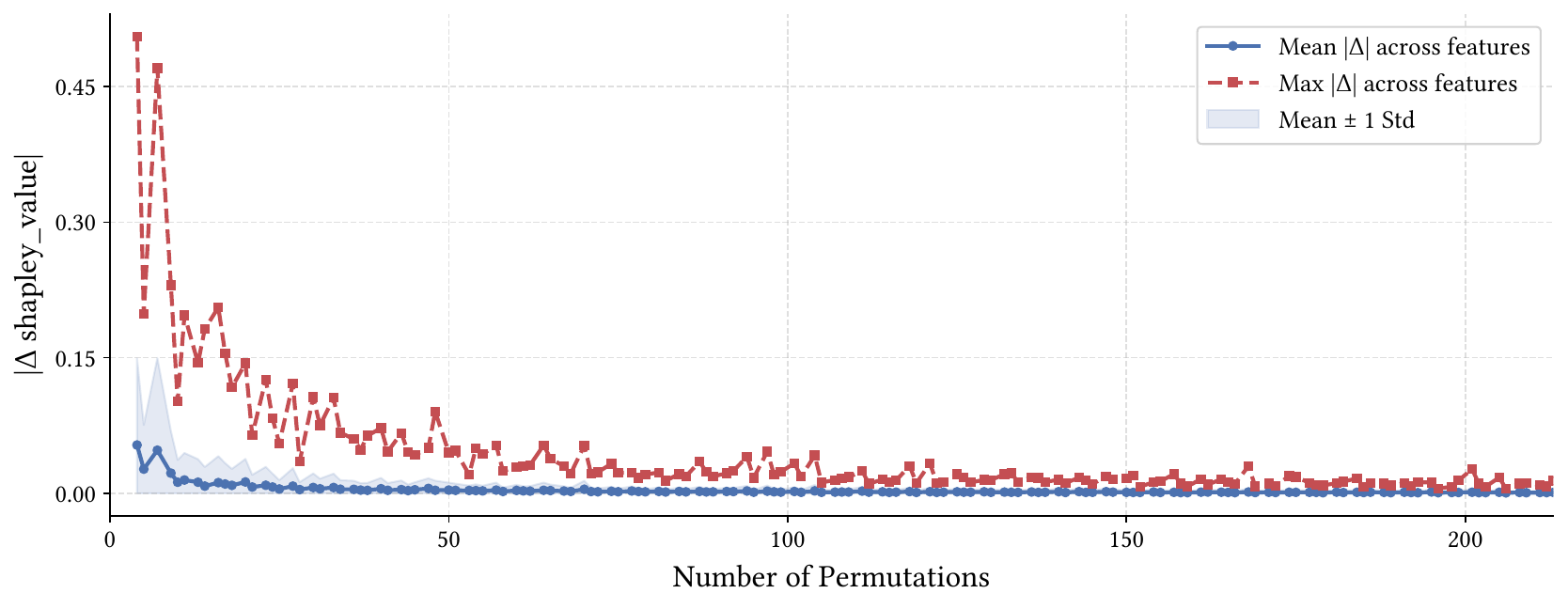}
    \caption{A representative empirical convergence curve (shown for HGS on the \texttt{rel-trial} dataset for the \texttt{study-adverse} task). The plot tracks the absolute difference in estimated scores ($|\Delta \text{shapley\_value}|$) between consecutive permutation checkpoints. 
    }
    \label{fig:shapley_convergence}
\end{figure}

Theoretical convergence guarantees (such as Hoeffding-Chernoff bounds~\cite{gerbessiotis2025surveychernoffhoeffdingbounds}) for \shapley require a computationally prohibitive number of permutations. In our experiments (\Cref{sec:experiments}), we adopt an empirical termination strategy based on stability over time.
During execution, we track the running mean of the Shapley estimates utilizing Welford's online algorithm~\cite{schubert2018numerically}, a numerically stable method for computing the running variance and mean. To evaluate convergence, we measure step-by-step stability by calculating the absolute difference in the estimated scores between consecutive checkpoints for all attributes:
\begin{equation*}
\Delta W_{\text{\shapley}}^{(t)}(A_i) = |W_{\text{\shapley}}^{(t)}(A_i) - W_{\text{\shapley}}^{(t-1)}(A_i)|
\end{equation*}
where $W_{\text{\shapley}}^{(t)}(A_i)$ represents the estimated importance score at elapsed hour $t$. 
We terminate once the convergence curve flattens out.
A representative convergence curve is illustrated in \Cref{fig:shapley_convergence} (while others are in~\Cref{app:shapley_convergence}.

\subsubsection{Kernel SHAP (\kernel)}
\label{sub:kernel_shap}

An alternative approximation method is \kernel~\cite{lundberg2017unified}, which frames
the explanation of a black-box model as a weighted linear regression. 
Unlike exact Shapley computations that require evaluating an exponentially large number of permutations, \kernel estimates the importance scores by fitting an additive surrogate model on a bounded set $C \subseteq \mathcal{P}(\datatteq)$ of sampled attribute coalitions. 

To construct the sample set $C$, we adopt a structured sampling strategy aligned with the original \kernel implementation~\cite{lundberg2017unified}, and adapted for our global evaluation setting. Coalitions are constructed by fully enumerating small  (e.g., $|S| \in \{1, 2\}$) and large (e.g., $|S| \in \{N-1, N-2\}$) subset sizes that naturally carry the highest SHAP kernel weights $\delta(S)$.
These weights $\delta(S)$ are theoretically derived to recover the original Shapley values and are defined as:
\begin{equation*}
\delta(S) = 
\frac
{(N-1)}{\binom{N}{|S|}\cdot |S| \cdot (N-|S|)}
\end{equation*}
The remaining evaluation budget is filled by randomly sampling coalitions such that the probability of drawing a coalition $S$ is proportional to its weight $\delta(S)$. Since $\delta(S)$ depends only on the coalition size $|S|$, this is efficiently implemented without explicitly enumerating the entire space: the explainer first samples a subset size $k$ according to the aggregate weights of each size, and then draws a combination of $k$ attributes uniformly at random.

For a sampled coalition $S \in C$, we evaluate the characteristic function $\nu(S)$ as defined in \Cref{eq:shapley_characteristic}.
The surrogate explanation model approximates the global Shapley values $W(A_i)$ for each attribute $A_i$ via the optimized coefficients $\phi_i$, which are obtained by solving the following weighted least-squares optimization problem:
\begin{equation*}
(\phi_0,\dots,\phi_N) = \arg\min_{\phi'_0,\dots,\phi'_N} \sum_{S\in C} \left[ \nu(S)-\left(\phi'_0+\sum_{A_i\in S}\phi'_i\right) \right]^2 \delta(S)
\end{equation*} 

Comparing the two approximation strategies, \kernel is generally more computationally efficient than \shapley. 
This efficiency stems from \kernel optimizing the variance of the estimates over a constrained sample space using weighted least-squares, whereas \shapley averages marginal contributions over sampled permutations. \kernel provides a consistent Shapley value estimator, but whether it is unbiased remains unknown~\cite{DBLP:conf/aistats/CovertL21}.

\subsection{Mask Learning} 
The \mask explainer~\cite{rissaki2025database} is an optimization-based explanation method designed to identify a small subset of attributes that are sufficient to maintain the model's original prediction. Unlike sampling-based methods (such as \shapley and \kernel), it formulates the explanation task as a mask-learning problem. For a given set of relational tables, the explainer learns a continuous mask vector $m \in [0, 1]^{N}$ corresponding to the attributes in \datatt, where each element $m_i$ represents the importance of attribute $A_i \in \datatteq$ across the entire relational schema.  Upon convergence, these learned mask values directly serve as our attribution function, setting $W(A_i) = m_i$.

\subsubsection*{Optimization objective}
Following~\cite{rissaki2025database}, the mask $m$ is learned by solving a continuous optimization problem that minimizes the prediction error while encouraging $m$ to promote only a small set of attributes. 
Let $t$ be a target tuple, and let $f(t, D)$ be the prediction of the model $f$ on the original input database $D$. Let $f(t, D_{m, \omega})$ be the prediction on the masked database $D_{m, \omega}$ under a random perturbation distribution $\omega$, which we describe in detail below. The objective employs a loss function to penalize the discrepancy between the original and masked predictions, alongside an $L_1$ regularization weighted by a hyperparameter $\lambda$ to encourage the mask $m$ to be sparse. Formally, the objective function is defined as:
\begin{equation}
\label{eq:mask_learning}
\min_{m \in [0,1]^N} \mathbb{E}_{t} \mathbb{E}_\omega \left[ \text{loss}(f(t, D), f(t, D_{m, \omega})) \right] + \lambda \|m\|_1
\end{equation}

\subsubsection*{Relational masking strategy}
The masked database $D_{m, \omega}$ is obtained through a mapping function $h(m)$ that applies the learned masks to the relational attributes. This mapping interpolates between the original value $t[A_i]$ of an attribute $A_i$ in a tuple $t$ and an ``uninformative'' alternative value. At each optimization step, we randomly sample for each tuple $t$ an alternative tuple $t'$ from the same relation, and use $t'[A_i]$ as the alternative value for every attribute $A_i$.  
The continuous mask value $m_i$ dictates the weighting between the original and alternative values to produce the masked representation $\widetilde{t[A_i]}$. The masking operation for each attribute is computed as $\widetilde{t[A_i]} = m_i \cdot t[A_i] + (1 - m_i) \cdot t'[A_i]$.
This masking operation is performed on the real-valued learned feature embeddings inside the model's forward pass, ensuring a fully differentiable formulation. For all graph-based architectures, the masks are applied within their shared tuple-level feature encoder, as formalized in~\cite{rissaki2025database}.

\subsubsection*{Implementation and hyperparameter selection}
We incorporate an automated selection mechanism for the $\lambda$ hyperparameter 
aiming to balance the two components of \Cref{eq:mask_learning} at the beginning of optimization.
The optimization is performed using the Adam optimizer over a fixed number of iterations, adopting the optimization hyperparameters reported in~\cite{rissaki2025database}.

%% file: experiments.tex
\section{Experimental Evaluation}\label{sec:experiments}

We present comprehensive experiments to evaluate the alignment, consistency, and faithfulness of explanation methods applied to relational deep learning architectures. We aim to address three main research questions: ($Q_1$) How do explanation methods differ in the \emph{ranking} of the attributes? ($Q_2$) How do they differ in the distribution of \emph{importance scores} to the attributes? ($Q_3$) How do they differ in their fidelity scores, and to what extent are these differences reflected in the ordinal and distributional comparisons?

\paragraph{Summary of results.} 
Our empirical evaluation reveals that explainer consensus depends heavily on the nature of the predictive task. 
Across both ranking (Section~\ref{subsec:rq1}) and score distribution (Section~\ref{subsec:rq2}) evaluations, we observe that regression problems generally yield a unified subset of critical attributes across different methods, whereas classification tasks produce more varied importance assignments.
Furthermore, while standard fidelity metrics (Section~\ref{subsec:rq3}) 
broadly corroborate our direct comparisons, we also observe cases where divergent explanations achieve comparable fidelity. This demonstrates that comparing explainers by standard faithfulness evaluations alone can mask disagreements, making our alignment analysis a required complement to standard evaluations.

\subsection*{Experimental Setup}

Next, we describe the general setup of our empirical study. The implementation is in Python 3.12 and publicly available.\footnote{\url{https://github.com/Hila-Levi/Comparing-Explanation-Methods-in-RDL}}
\subsubsection{Datasets and tasks}
We focus on predictive tasks from RelBench, spanning datasets from both its initial release~\cite{relbench} and the updated 
version~\cite{gu2026relbenchv2largescalebenchmark}. 
We use entity-level tasks where the goal is to predict a specific outcome (classification or regression) for a target entity by aggregating information across multiple tables. 

\subsubsection*{rel-trial}
This dataset~\cite{relbench} stores clinical trial records in 15 interconnected tables, including \textit{studies}, \textit{interventions}, and \textit{outcomes}. Together, the tables contain 140 attributes and approximately 5.5M rows. This dataset contains a rich set of attributes, including categorical study types, numerical participant counts, and textual descriptions of medical procedures. We use two distinct tasks: 
\begin{enumerate}
\item \texttt{study-outcome} (classification): whether a trial will achieve its primary outcome (defined as reaching a p-value $< 0.05$).
\item \texttt{study-adverse} (regression):  the number of affected patients with severe adverse events during the trial.
\end{enumerate}

\subsubsection*{rel-arxiv}
This dataset~\cite{gu2026relbenchv2largescalebenchmark} stores physics publications on arXiv from 2018 to 2023 and consists of 6 interconnected tables, including \textit{papers}, \textit{authors}, and \textit{categories}. The schema has 21 attributes with approximately 2M rows. This dataset contains diverse attributes such as paper titles and abstracts (represented as pretrained text embeddings).
We use two tasks within this dataset:
\begin{enumerate}
\item 
\texttt{paper-citation} (classification): whether a given paper will receive at least one citation within the next 6 months.
\item \texttt{author-publication} (regression): number of papers a given author will publish within the next 6 months.
\end{enumerate}

\subsubsection*{rel-f1}
This dataset~\cite{relbench} stores over five decades of Formula 1 racing data in 9 interconnected tables including \textit{races}, \textit{drivers}, \textit{constructors}, and \textit{results}. It has 67 attributes with approximately 98K rows across all tables.
We use two tasks: 
\begin{enumerate}
\item \texttt{driver-dnf} (classification): whether a driver will get a ``Did Not Finish'' (DNF) status in a race within the next 1 month.
\item 
\texttt{driver-position} (regression): the average finishing position of each driver across all races in the next 2 months.
\end{enumerate}

\subsubsection*{rel-event} This dataset~\cite{relbench} stores user activity on the Hangtime event application, consisting of 5 interconnected tables including \textit{users}, \textit{events}, and \textit{event\_attendees}. The dataset has 128 attributes and approximately 41.3M rows.
We use two tasks: 
\begin{enumerate}
\item 
\texttt{user-repeat} (classification): whether a user who attended a recent event will attend a future one (respond yes/maybe).
\item \texttt{user-attendance} (regression): the total number of events to which a user will respond yes/maybe in the next 7 days.
\end{enumerate}

\subsubsection{Prediction Models}

Following~\citet{fey2024position}, a database is represented as a graph where each tuple corresponds to a node, and edges are established based on FK/PK relationships. We experiment with two models of different paradigms: a message passing GNN and a transformer-based architecture.

\begin{itemize}
    \item Heterogeneous GraphSAGE (\textbf{HGS})~\cite{hamilton2017inductive, fey2019fast}: the 
    message-passing GNN used in RelBench. 
    \item Relational graph transformer (\textbf{RelGT})~\cite{dwivedi2025relational}: a state-of-the-art transformer-based GNN.
\end{itemize}
In both architectures, following the RelBench methodology~\cite{relbench}, initial node embeddings are generated using a ResNet-based tabular encoder~\cite{gorishniy2021revisiting}. 
To support attribute-level explanations via masking, we utilize the implementation of~\cite{rissaki2025database}, which augments the tuple-level feature encoder shared across graph-based architectures with real-valued masking capabilities.
In addition to HGS and RelGT, we conducted a complementary evaluation on a third architecture, the \textbf{Relational Transformer (RT)}~\cite{ranjan2026rtjlargescalepretrainingrelational}, using the three removal-based explainers (\single, \shapley, and \kernel). The full RT results, which corroborate our main findings across all three research questions, are provided in~\Cref{app:rt}.

\subsubsection{Explainer Configurations}
\label{sub:explainer_configs}

To enable a methodological alignment among the removal-based explainers (\single, \shapley, and \kernel), we utilize a unified, deterministic ablation operator: mean imputation for numerical and text attributes (computed over their pretrained embeddings), median imputation for temporal attributes, and the Most Frequent Value (MFV) for categorical attributes.
This design choice is motivated by the foundational approximations of \kernel~\cite{lundberg2017unified}, which simulates missing attributes using expected marginal values under the assumption of feature independence. 
\mask inherently relies on a permutation-based masking strategy during its optimization, and is therefore not affected by this choice.

For the \kernel explainer, we follow the standard sample-complexity heuristic~\cite{lundberg2017unified} recommending $2N+2048$ samples. In our schemas $N\leq 140$, and we round up to enable robust empirical approximation, resulting in a fixed budget of 3,000 samples.

\subsection{Explainer Ranking Alignment}\label{subsec:rq1}

\subsubsection*{Objectives} 
We address \(Q_1\) by comparing attribute rankings produced by the explanations, independently of their absolute scores.

\subsubsection*{Evaluation Methodology} 

\begin{figure*}[t]
    \centering
    \includegraphics[width=\linewidth]{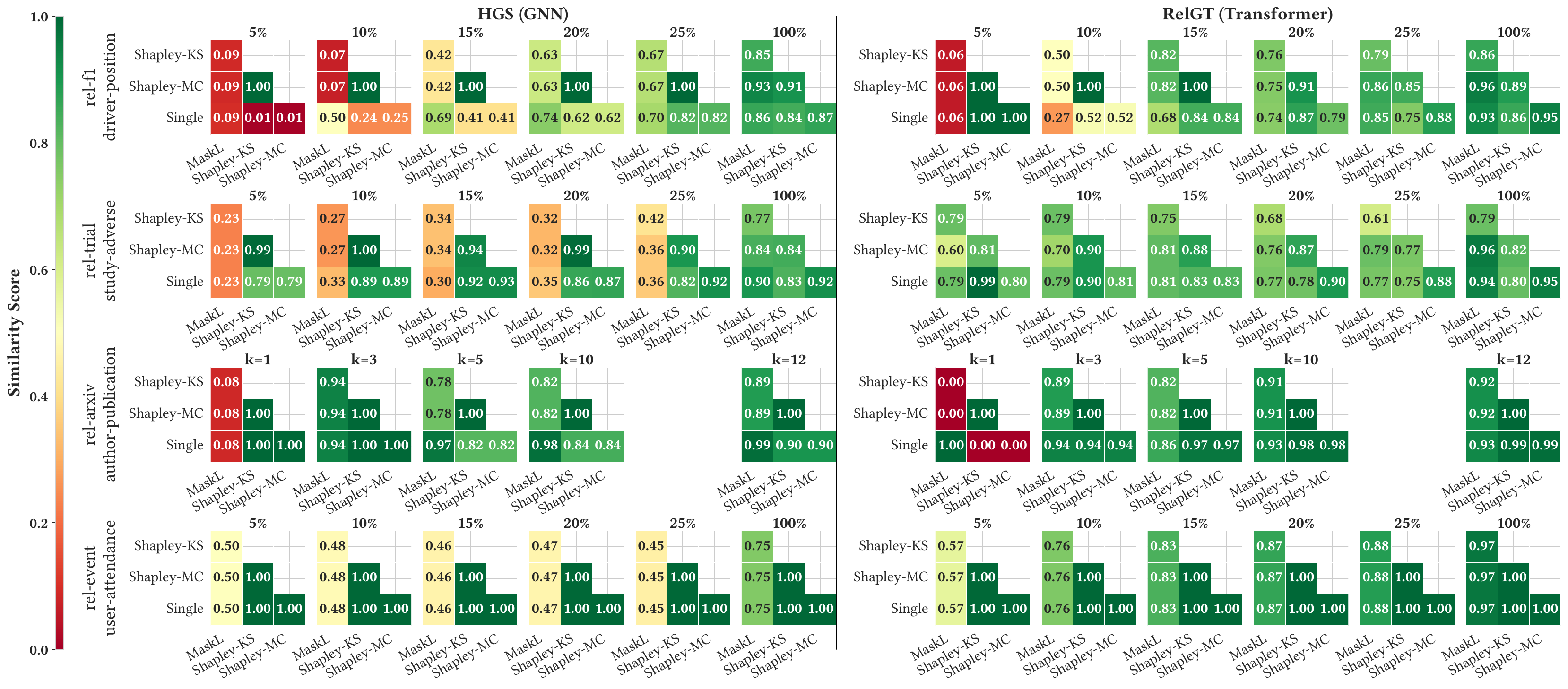}
    \caption{Pairwise explainer similarity for regression, showing alignment on different prefixes ($k$) for HGS and RelGT.}
    \label{fig:regression_heatmaps}
\end{figure*}

\begin{figure*}[t]
    \centering
    \includegraphics[width=\linewidth]{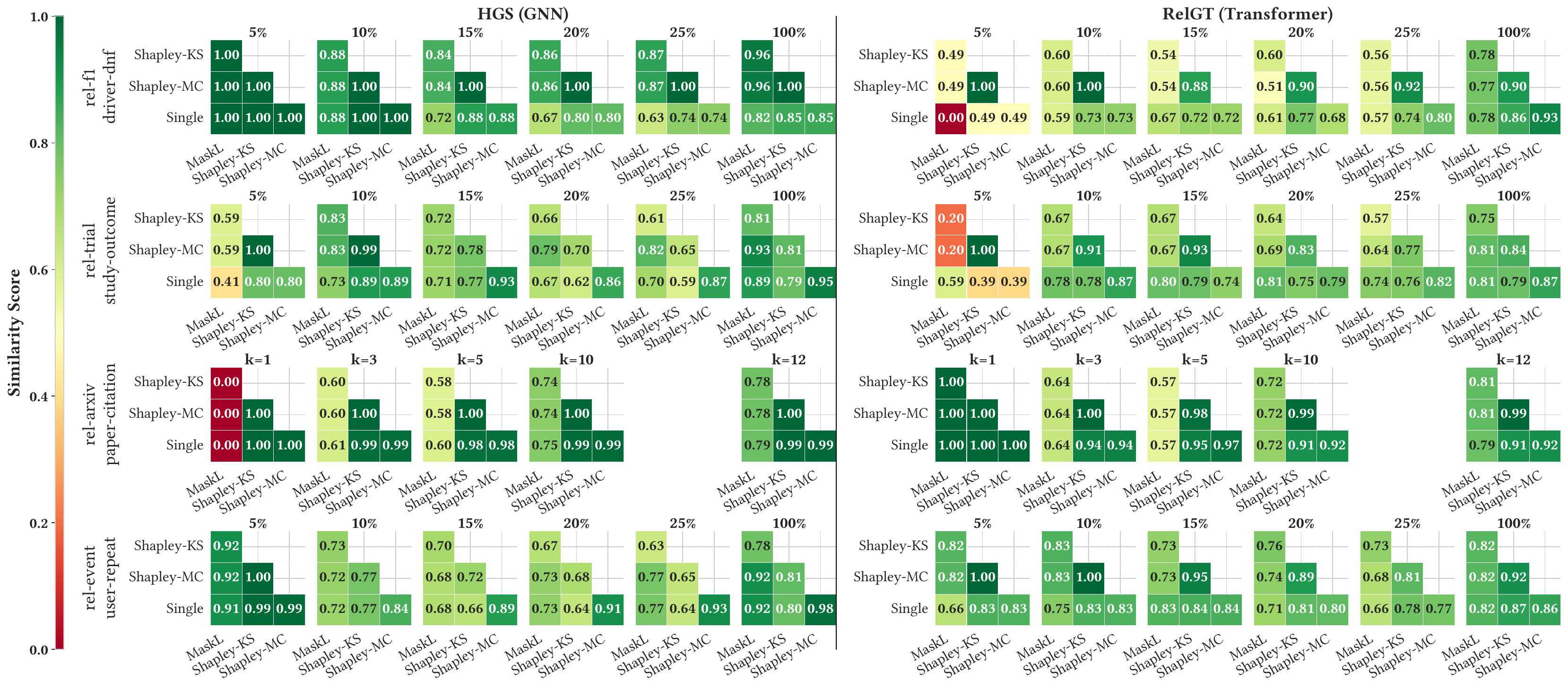}
    \caption{Pairwise explainer similarity for classification, showing alignment on different prefixes ($k$) for HGS and RelGT.}
    \label{fig:classification_heatmaps}
\end{figure*}

Explainers produce an importance score for each attribute.
To compare top-$k$ rankings with ties, we use the framework of~\citet{fagin2003comparing, 10.1145/1055558.1055568}.
There, tied elements form equivalence classes (buckets) and share their bucket's average rank.
Our importance scores may be nearly identical without being exactly tied. 
Our custom quantization step therefore groups attributes into equivalence classes using a dynamic, range-based threshold computed separately for each explainer's score distribution, $\tau = (w_{\max} - w_{\min}) / 100$.
We greedily scan attributes in descending score order, starting a new bucket only when a score is more than $\tau$ below the highest score in the current bucket.
Following Fagin's methodology, we assign each bucket its average real-valued rank and project the scores into a unified partial (top-$k$) ranking $\sigma$, where $\sigma(A_i)$ is the final rank of attribute $A_i \in \datatteq$.
For each $A_i \in \datatteq$, we set $\sigma(A_i) = \mbox{rank}(A_i)$ if its sorted position (from 1 to $N$) after alphabetical tie-breaking falls within the top-$k$ window.
Attributes outside the top-$k$ window receive a tied rank of $\ell = (N + k + 1) / 2$, the neutral average rank of all unranked items.

To quantify pairwise alignment between two explainers, we follow~\citet{10.1145/1055558.1055568} and use the $L_1$ distance (generalized Spearman footrule) between the rankings $\sigma_1$ and $\sigma_2$.
Specifically, for a prefix size $k$, define a normalized similarity score using the top-$k$ distance:
\begin{equation*}
\text{Similarity}(\sigma_1, \sigma_2) = 
1 - \frac{\sum_{A_i \in \datatteq} |\sigma_1(A_i) - \sigma_2(A_i)|}{2k\ell - k(k+1)}
\end{equation*}
\begin{figure*}[t]
    \centering
    \begin{subfigure}[b]{0.48\linewidth}
        \centering
        \includegraphics[width=\linewidth]{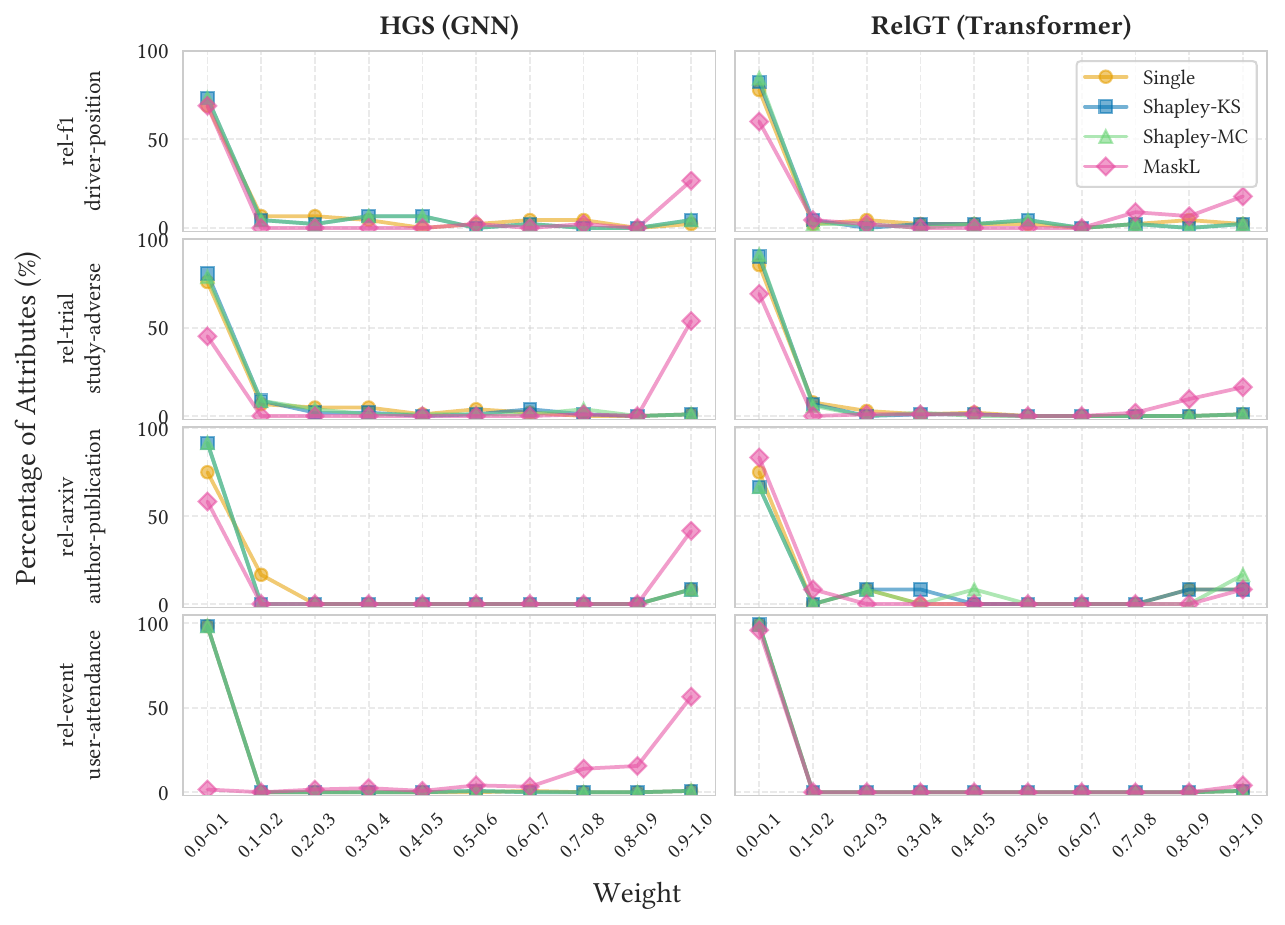}
        \caption{Regression tasks.}
        \label{fig:distribution_regression}
    \end{subfigure}
    \hfill
    \begin{subfigure}[b]{0.48\linewidth}
        \centering
        \includegraphics[width=\linewidth]{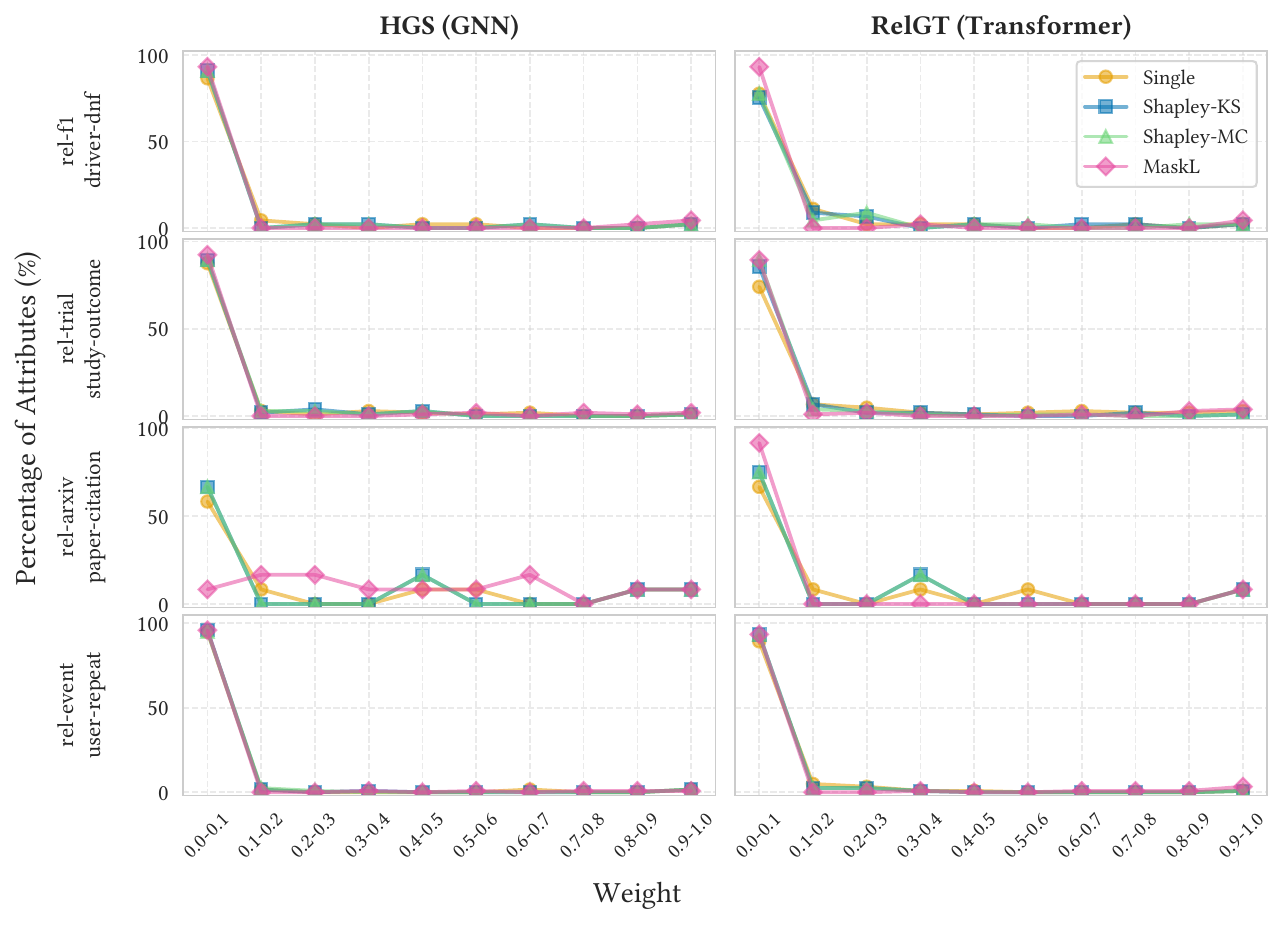}
        \caption{Classification tasks.}
        \label{fig:distribution_classification}
    \end{subfigure}
    \caption{Score distribution (PDF). Dots show the percentage of attributes within the normalized weight range. 
    }
    \label{fig:distribution_grid}
\end{figure*}
As expressed in the formula, the disagreement distance is the total sum of absolute rank differences across all attributes. 
The distance is normalized to $[0,1]$ by the maximum possible distance, $2k\ell - k(k+1)$, which occurs when the top-$k$ sets are disjoint.
We transform the distance to similarity by subtracting it from 1. The results are between $0$ (complete disagreement) and $1$ (full agreement).

We visualize alignments using pairwise similarity heatmaps (\Cref{fig:regression_heatmaps,fig:classification_heatmaps}).
As pairwise similarity is symmetric and the diagonal is always $1.0$, we display only the lower-triangular matrix to reduce visual redundancy. To maintain structural relevance across varying schema scales, large schemas are evaluated using percentage-based prefixes ($5\%$ to $100\%$ of attributes), while compact schemas are tracked via absolute attribute counts. Darker green cells indicate higher agreement, whereas red configurations signal disagreement. 

\subsubsection*{Results}

Our analysis reveals that agreement is primarily driven by the nature of the task rather than the choice of the neural architecture. In regression (\Cref{fig:regression_heatmaps}), we observe high agreement among most explanation methods, growing stronger at larger $k$ values. Specifically, \single, \shapley, and \kernel demonstrate strong alignment in their global and top-$k$ rankings while \mask frequently diverges from this consensus.
Conversely, classification tasks (\Cref{fig:classification_heatmaps}) exhibit markedly less agreement among the explainers, as evidenced by the consistently lower similarity scores (weaker green intensities) across all $k$ values. 

Since \kernel and \shapley are both strategies designed to estimate the exact Shapley values, we often observe a perfect agreement score between them, especially over the top attributes.
However, this is not always the case: for example, the methods disagreed on the \textit{study-adverse} task under RelGT, \textit{study-outcome} and \textit{user-repeat} for both architectures. To verify that these discrepancies are not simply artifacts of insufficient sampling, we empirically validated the convergence of both methods (convergence plots for \kernel are provided in~\Cref{app:kernel_convergence} alongside those for \shapley in~\Cref{app:shapley_convergence}). Therefore, these differences stem inherently from the distinct estimation mechanisms of the two explainers.
Hence, while the two approximations reach strong agreement on the highest-ranked attributes (i.e., at lower $k$ values), their rankings frequently diverge as  $k$ increases.

While \single, \kernel, and \shapley often behave similarly in their importance scores, their execution times differ significantly. The runtime of \single depends directly on the number of database columns, whereas \kernel and \shapley also depend on the number of samples.
In our empirical evaluation, when measuring execution times at the point of empirical convergence for each method, \kernel completed its execution a median of 37$\times$ faster than Monte Carlo sampling.
However, the \single explainer remains the most computationally efficient, completing its execution a median of 9.3$\times$ faster than \kernel. A more comprehensive evaluation of run times is left for future work. These runtime performance trade-offs highlight the practical advantages of the \single explainer in large-scale relational schemas.

\begin{figure*}[t]
    \centering
    \begin{subfigure}[b]{0.48\linewidth}
        \centering
        \includegraphics[width=\linewidth]{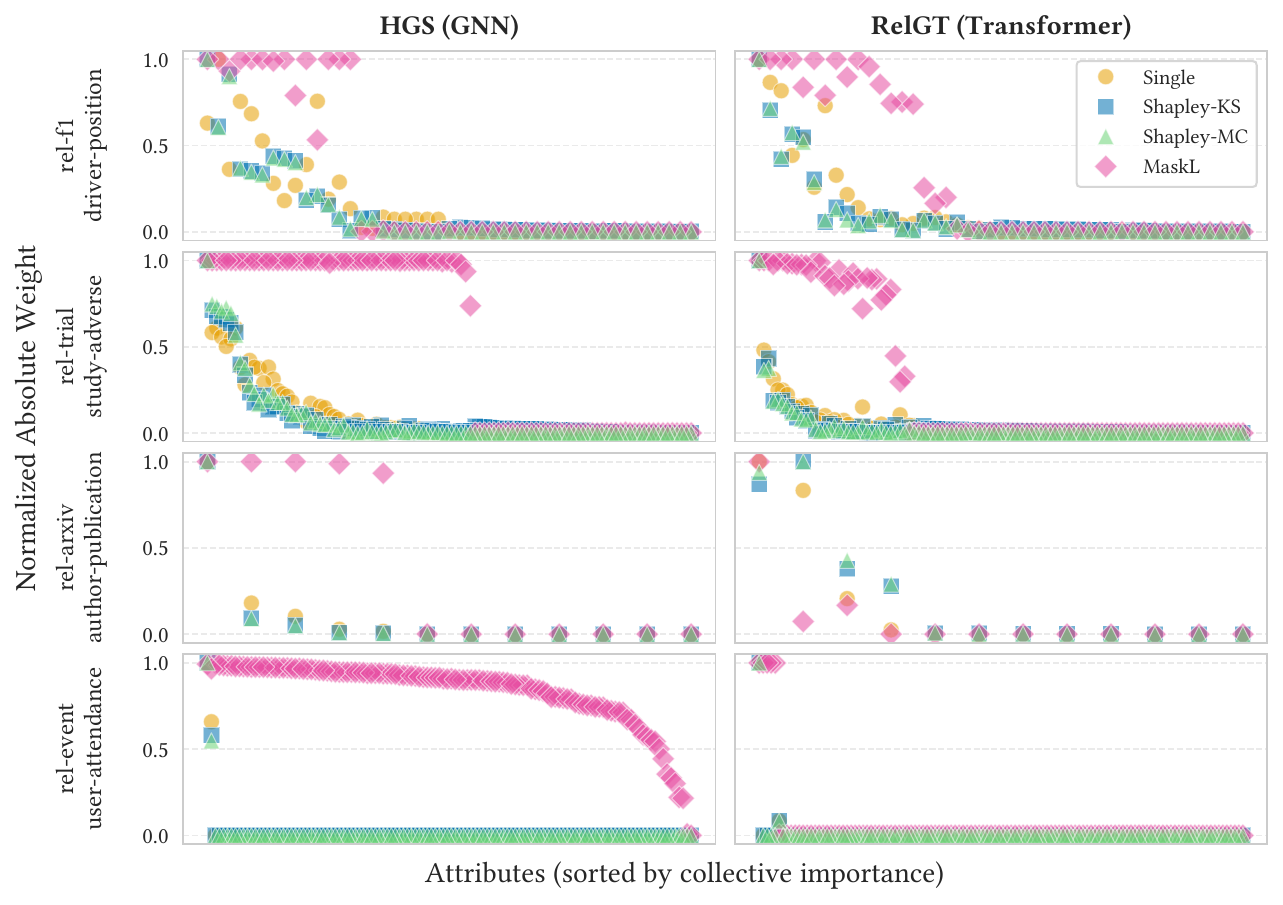}
        \caption{Regression tasks.}
        \label{fig:scatter_regression}
    \end{subfigure}
    \hfill
    \begin{subfigure}[b]{0.48\linewidth}
        \centering
        \includegraphics[width=\linewidth]{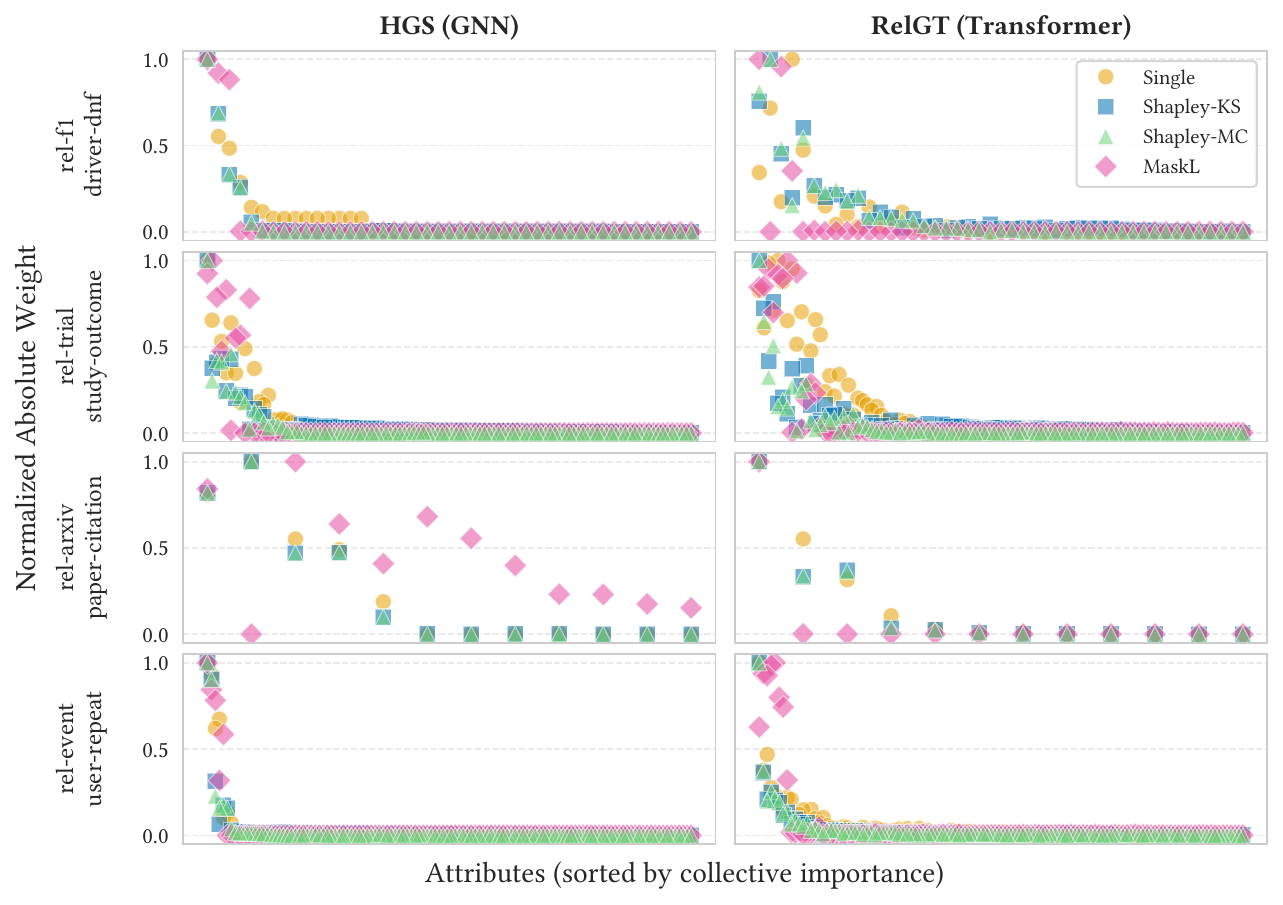}
        \caption{Classification tasks.}
        \label{fig:scatter_classification}
    \end{subfigure}
    \caption{Normalized score scatter grid. Attributes along the x-axis are sorted by collective importance.}    
    \label{fig:scatter_grid}
\end{figure*}

\subsubsection*{Conclusions} 
Our findings indicate that regression tasks have significantly higher explainer agreement than classification. When methods produce highly similar rankings, they differ primarily by computational efficiency. There, \single emerges as an efficient proxy, delivering rankings that align closely with the computationally prohibitive \shapley and \kernel methods. 

Additionally, the persistent divergence of \mask highlights that an explainer's behavior is heavily dictated by its mathematical formulation. Specifically, \mask evaluates \emph{collaborative impact} by optimizing all attributes simultaneously to discover the most predictive joint subset, rather than isolating individual contributions. Furthermore, \mask employs permutation-based masking instead of deterministic baseline imputation. These differences in core objectives and underlying ablation operators play a critical role in shaping the resulting importance rankings.

\subsection{Explainer Score Distribution}\label{subsec:rq2} 

\subsubsection*{Objectives} 
To address \(Q_2\), we compare the importance-score distributions, retaining information about score magnitudes that is discarded by the ranking analysis of the previous section. 

\subsubsection*{Evaluation methodology} 
To enable meaningful comparisons across explainers with different score distributions, we apply min-max normalization to map absolute attribution weights onto a standardized continuous scale $[0, 1]$: 
\begin{equation*}
\tilde{W}(A_i) = \frac{|W(A_i)| - \min_{A_j \in \datatteq} |W(A_j)|}{\max_{A_j \in \datatteq} |W(A_j)| - \min_{A_j \in \datatteq} |W(A_j)|}
\end{equation*}
After the transformation, the most vital attribute receives a score of $1.0$ and the least important one receives a score of $0.0$. We compare the different explainers by inspecting their normalized score distributions and the individual attributes with the highest scores.

\subsubsection*{Results}
We first examine the overall score distributions using the probability density function (PDF) of attribution weights (\Cref{fig:distribution_grid}), by dividing the normalized range of weights into discrete bins and measuring the percentage of attributes in each bin. 
Across both task types, most methods assign very low importance scores to the vast majority of attributes, reserving high scores for only a select few. However, in regression tasks (\Cref{fig:distribution_regression}), \mask exhibits a notably different behavior: it is more ``generous,'' assigning maximum importance values (in the 0.9--1.0 bin) to a considerably larger proportion of attributes compared to the other explainers.

To understand how these scores behave across individual attributes, we plot the individual normalized importance scores as a scatter (\Cref{fig:scatter_grid}), where the attributes along the x-axis are ordered by their cumulative normalized weight over all explainers. We observe a general consensus among all methods regarding which attributes are highly important versus those that are negligible. In regression tasks (\Cref{fig:scatter_regression}), the individual explainer scores track closely together, forming a tight and steep decay curve, though \mask maintains high scores for a wider range of attributes before decaying. Conversely, in classification tasks (\Cref{fig:scatter_classification}), the scores are noticeably more dispersed, indicating lower explainer agreement on the exact importance magnitudes of the top attributes.

Finally, we construct an ensemble consensus score grid (\Cref{fig:consensus_grid_regression,fig:consensus_grid_classification}) to inspect the specific attributes driving the predictions. 
To reduce noise from zero-weight attributes whose relative ranking would depend solely on alphabetical tie-breaking, we first filter out attributes whose normalized score falls below $0.001$ across all explainers. From the remaining attributes,
we isolate the union of attributes that rank within the top $k=15$ of at least one method and display the top 10 attributes from this union, ordered by the number of explainers that included them. Within each panel cell, we present the explainer's ordinal rank position ($\#$) and the normalized score (encoded by the bar width).
Each attribute is preceded by a gray circle indicating the hop distance of its table from the target table ($0$ for the target table), where HGS aggregates over $2$-$3$ hops depending on the task and RelGT operates over $2$ hops.
As previously noted, alphabetical tie-breaking is applied when attributes share identical importance scores. It is critical to keep this tie-breaking strategy in mind, as it heavily influences the exact top-$k$ choices when an explainer (such as \mask in regression tasks) assigns the identical maximum score of 1.0 to a large number of attributes. These grids corroborate our previous findings while providing semantic insights into the model's logic.

Across tasks and models, the top-ranked attributes span all layers of each model's receptive field, drawing from both the target table ($0$-hop) and distant tables up to the maximum hop depth ($2$-hop in RelGT and up to $3$-hop in HGS). This confirms that the models aggregate predictive signal across the full relational graph rather than relying solely on the target table.

An inspection of these top-ranked attributes suggests that the explainers tend to prioritize historical indicators that directly relate to the target. For instance, in the \texttt{rel-f1} \texttt{driver-position} regression task under the RelGT architecture, all explainers agree that \texttt{res.posOrder} (the finishing position in previous races) is the most critical attribute (\mask also assigns it the maximum score of 1.0, placing it at $\#4$ due to alphabetical tie-breaking). This aligns with the intuition that a driver's past finishing positions are strong indicators of their future average position. 

Similarly, in the \texttt{rel-event} \texttt{user-attendance} task under RelGT, we observe a strong preference for behavioral signals. In particular, \texttt{evAttend.stat}, representing the user's historical RSVP responses (\emph{invited}, \emph{yes}, \emph{maybe}, \emph{no}), is unanimously identified as the leading attribute. This aligns with the intuition that a participant's future event attendance is strongly influenced by their past attendance.

{
\def\consensuswidth{0.485\linewidth}
\begin{figure*}[t]
    \centering
    \begin{subfigure}[b]{\consensuswidth}
        \centering
        \includegraphics[width=\linewidth]{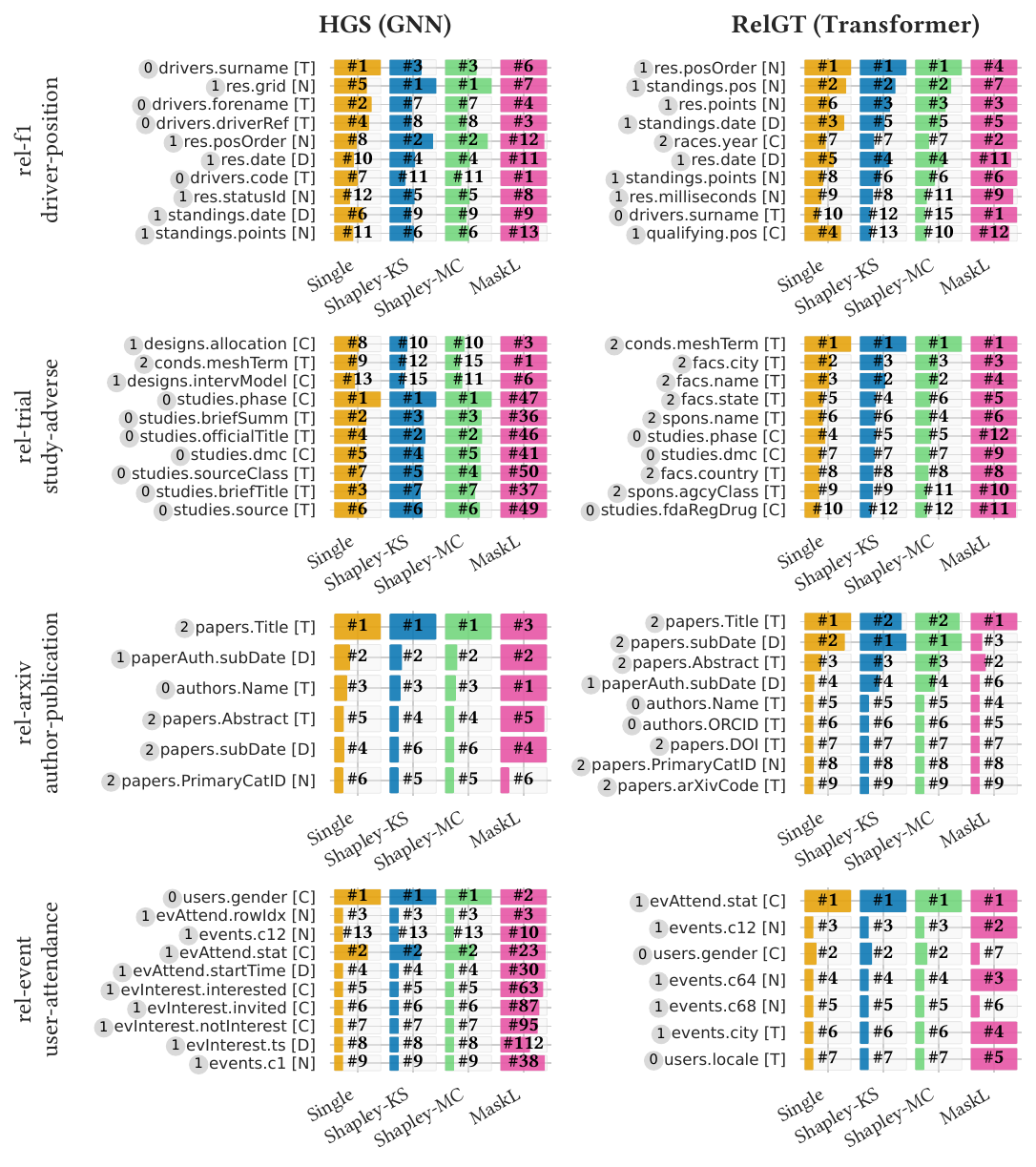}
        \caption{Regression tasks.}
        \label{fig:consensus_grid_regression}
    \end{subfigure}
    \hfill
    \begin{subfigure}[b]{\consensuswidth}
        \centering
        \includegraphics[width=\linewidth]{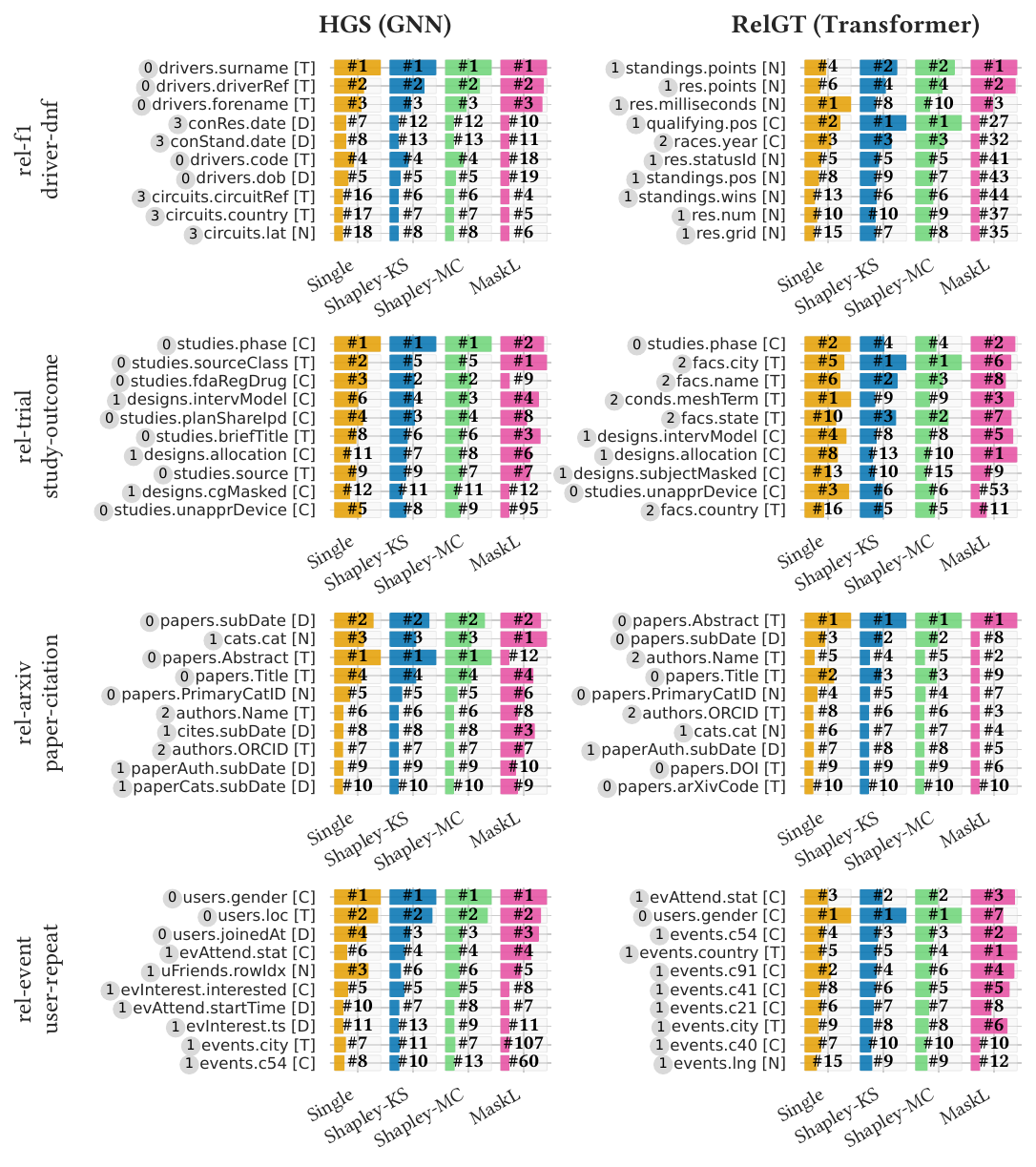}
        \caption{Classification tasks.}
        \label{fig:consensus_grid_classification}
    \end{subfigure}
    \caption{Consensus min-max normalized score grids across all tasks (top-15 evaluation, displaying top 10 consensus attributes after filtering out attributes with normalized scores below $0.001$ across all explainers). Gray circles denote each attribute's hop distance from the target table ($0$ is the target table).
    Cells display individual ranks ($\#$) and bar widths represent normalized scores. Data types are in brackets: \texttt{[N]} Numerical, \texttt{[C]} Categorical, \texttt{[D]} Date, and \texttt{[T]} Text. 
    Alphabetical tie-breaking may affect the exact top-$k$ choices (See Section~\ref{subsec:rq1}).
    }
    \label{fig:consensus_grid}
\end{figure*}}

\subsubsection*{Conclusions}
The manner in which explainers distribute importance scores is driven by the predictive task. Regression tasks foster a strong consensus where multiple explainers agree on both the identity and the steep decay of the most critical attributes, providing practitioners with high confidence in the derived explanations. While \mask tends to be more generous with its score distributions in these settings, its ordinal alignment across the top attributes remains consistent with the other methods.
In contrast, classification tasks yield more scattered and inconsistent score distributions, reflecting higher uncertainty among the explanations. 
Across both settings, the leading attributes span multiple tables and hop distances ($0$-$2$ hops in RelGT and up to $3$ hops in HGS), showing that the models combine target-table attributes with multi-hop relational context.
Despite the distribution differences, when methods converge on specific semantic attributes, such as a driver's historical finishing order or a user's past RSVP behavior, they reveal intuitive insights into the underlying model's predictive mechanisms.

\subsection{Fidelity Metrics Evaluation}\label{subsec:rq3}

\subsubsection*{Objectives} 
Fidelity metrics are a widely used method to evaluate AI explainability tools~\cite{yuan2022explainability}. fidelity aims to quantify explanation faithfulness, or, how well do the explanations capture and reflect the prediction model's decisions. Following standard conventions~\cite{yuan2022explainability},
$\text{Fidelity}^+$ evaluates the \emph{necessity} of the top-ranked attributes by measuring the absolute prediction shift when those specific attributes are ablated. Conversely, $\text{Fidelity}^-$ assesses the \emph{sufficiency} of the top attributes by measuring the prediction shift when \emph{only} those attributes are retained and the rest are ablated. 
We now investigate to what extent the differences and similarities among explanation methods are reflected in their fidelity scores ($Q_3$).

\subsubsection*{Evaluation methodology} 
Across both classification and regression, fidelity (of either type) is evaluated via the absolute prediction shift induced on model output predictions.
To enable the comparison over schemas with varying numbers of attributes, we evaluate the fidelity across relative $k$ values ($5\%$ to $100\%$) for schemas with a large number of data attributes, and with absolute counts for schemas with fewer attributes. It is important to note that our fidelity evaluation employs the same deterministic ablation strategy used by all explanation methods, with the exception of \mask, which relies on its own permutation-based ablation during optimization.

\subsubsection*{Results} 
We examine how explanation subsets impact model performance by comparing 
$\text{Fidelity}^+$ and $\text{Fidelity}^-$ scores across varying top-$k$ subsets in all combinations of tasks and models (\Cref{fig:fidelity_grid_combined}). 

Overall, the fidelity scores are consistent with the results of
our ranking analyses (\Cref{subsec:rq1}). Methods that produce similar attribute rankings generally yield similar fidelity scores. 
For example, in the \texttt{rel-trial} \textit{study-adverse} regression task under HGS, \single, \shapley, and \kernel achieve high ranking similarity, while \mask receives very low agreement scores. This behavior is reflected in the fidelity metrics: the three aligned methods yield closely matched $\text{Fidelity}^+$ and $\text{Fidelity}^-$ scores, whereas \mask produces significantly different fidelity scores. A similar pattern emerges in the \texttt{rel-arxiv} \textit{paper-citation} classification task under RelGT, where \mask diverges significantly in both ranking and fidelity across most $k$ values. The only exception occurs at $k=1$, where all four explainers reach full ranking agreement, which is directly mirrored by identical fidelity scores across all methods at that specific point.

\subsubsection*{Conclusions} 
Fidelity metrics provide a valuable perspective broadly consistent with our comparisons: \single and the Shapley-based methods often group together in their predictive degradation profiles, while \mask exhibits distinct behavior. 
Nevertheless, fidelity metrics evaluate predictive utility rather than exact content. 
As aforesaid, \mask's tendency to assign identical maximum scores can trigger tie-breaking, causing apparent fidelity deviations at small $k$ values (e.g., \texttt{rel-arxiv} \textit{author-publication} at $k=1$). Furthermore, different subsets can have the same predictive power. For instance, in the \texttt{rel-trial} \textit{study-adverse} regression task under RelGT at $k=20\%$, we observe a ranking disagreement between \kernel and \mask (with a similarity score of 0.68), yet their fidelity is tightly clustered.

{
\begin{figure*}[t]
    \centering
    \includegraphics[width=\linewidth]{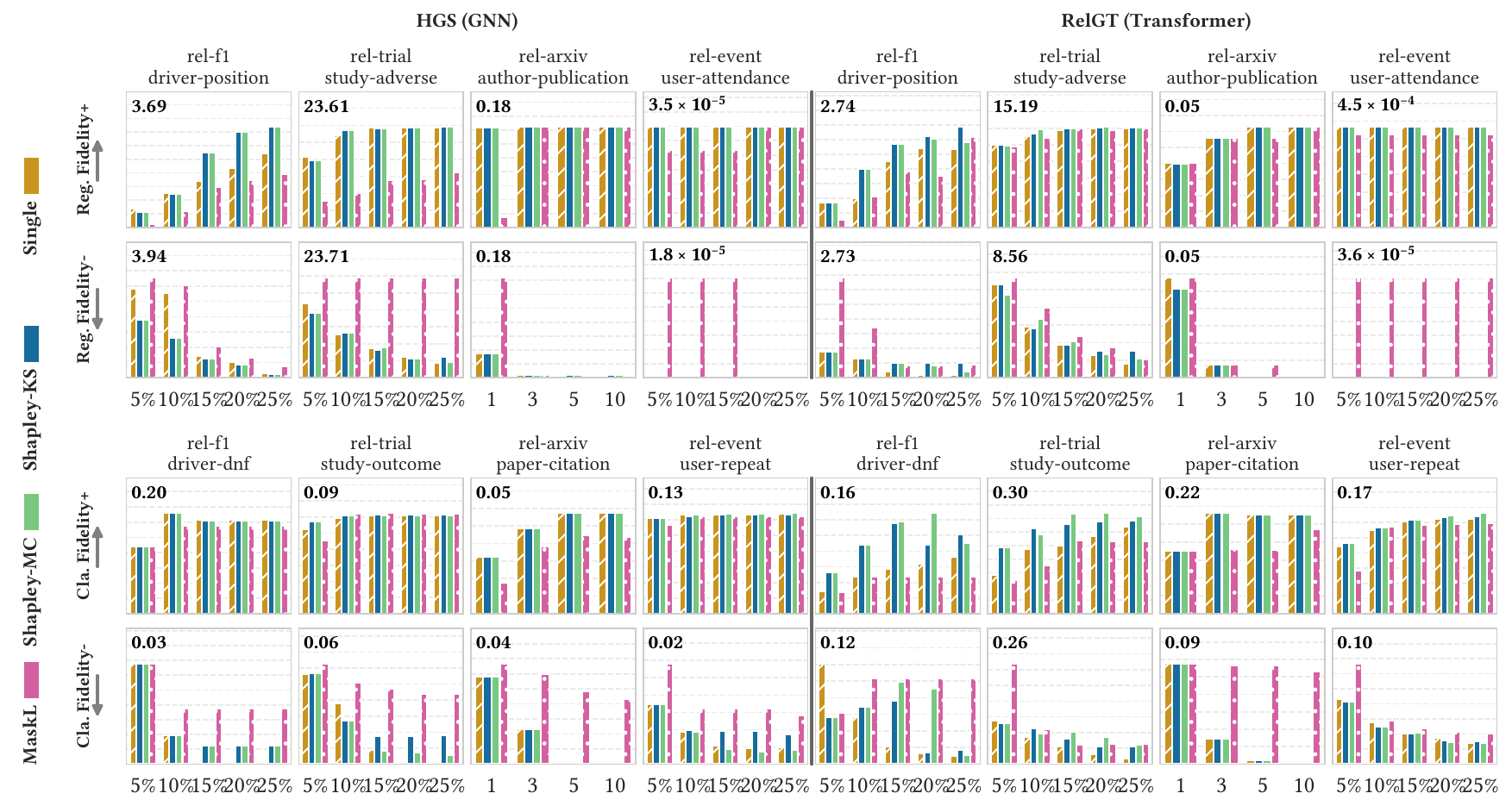}
    \caption{Explainer fidelity in regression (top 2 rows) and classification (bottom 2 rows) tasks, over HGS (left) and RelGT (right).
    Bars represent $\text{Fidelity}^+$ (necessity, higher is better) and $\text{Fidelity}^-$ (sufficiency, lower is better) across varying top-$k$ attribute subsets. 
    The maximum y-axis value for each subplot is annotated in its top-left corner. 
    }
    \label{fig:fidelity_grid_combined}
\end{figure*}
}

%% file: conclusions.tex
\section{Conclusions}\label{sec:conclusions}

We presented a systematic study of importance attribution methods for database columns in relational deep learning. We considered four representative methods and developed quantitative analyses and visualizations for comparing both the rankings they induce and the distributions of their importance scores. We applied these methods across multiple datasets and prediction tasks using two relational deep learning architectures, HGS and RelGT, 
and examined their relationship to conventional fidelity measures. 

Our results demonstrate that the extent of agreement among explanation methods is context-dependent. Specifically, explainer agreement is driven by the nature of the predictive task, with regression tasks yielding a much higher consensus than classification tasks. Furthermore, the agreement depends on the underlying mathematical objective of the explainer. While \single and the two Shapley approximations often align closely, \mask frequently diverges, reflecting its distinct focus on collective subset sufficiency rather than individual marginal contributions.

Our findings do not imply that any one explanation method is universally preferable. Rather, they suggest that importance scores should be interpreted \emph{in light of the method that produces them}. In cases where explanations play a substantive role, a system may benefit from presenting results from more than one method, or at least from clarifying the essence and notion of importance captured by the selected method. Similarly, scientific studies that draw conclusions from attribute-importance scores may strengthen their conclusions by examining whether they persist across alternative explanation methods. Finally, our results motivate considering the stability of explanations across different predictive tasks as a complementary aspect of model evaluation, alongside predictive performance. We view these as directions for practice and future investigation, rather than as general requirements.

Our study opens several directions for future investigation. First, while our empirical results often showed strong alignment between computationally expensive Shapley approximations and the highly efficient single-column ablation, a critical avenue for future work is identifying the specific dataset characteristics, model complexities, or problem domains where the rigorous game-theoretic properties of the Shapley value provide a tangible advantage that justifies its computational overhead. Second, as the ultimate goal of explainability is to provide actionable insights to users, our current evaluation is limited to quantitative algorithmic metrics. Future research should incorporate human-centric case studies to determine which mathematical notions of importance best align with user intuition and domain expertise. Finally, while our evaluation framework is broadly applicable to any predictive model operating over relational data (with the exception of Mask Learning, which requires architecture-specific integration), our experiments focused on graph-based neural networks. An important next step is to expand this methodology to encompass emerging architectures, such as relational foundation models and advanced tabular approaches (e.g., TabPFN), across a wider spectrum of relational datasets.

%% file: appendix_methods.tex
\section{Proof of Theorem~\ref{thm:shapley_hardness}}
\label{app:shapley_hardness}

We now prove \Cref{thm:shapley_hardness}.
We consider combined complexity, where the schema is part of the input, as under data complexity, the schema, and, in particular, the number of data attributes, is fixed; hence, the exact Shapley value can be computed in polynomial time in data complexity by enumerating all attribute subsets.

In the proofs below, we use the following equivalent subset-based
formula for the Shapley value:
\begin{equation*}
    W(A_i)
    =
    \sum_{S\subseteq\datatteq\setminus\{A_i\}}
    \frac{|S|!(N-|S|-1)!}{N!}
    \left[\nu(S\cup\{A_i\})-\nu(S)\right].
\end{equation*}

For simplicity, the reductions below use the Zero ablation strategy, in which every data attribute excluded from a coalition is replaced by $0$. The proofs also apply to mean imputation, which is the strategy used in our experiments for numerical attributes; we explain the required adjustment after each proof.

\begin{reptheorem}{\ref{thm:shapley_hardness}}
    \hardnessshapley
\end{reptheorem}

We prove Theorem~\ref{thm:shapley_hardness} by considering the two restricted schema settings separately. We first establish hardness for a schema consisting of a single relation.

\begin{lemma}\label{lem:shapley_hardness_single}
    Computing the Shapley value of an attribute under the global
    characteristic function $\nu$ is \#P-hard when the prediction
    model is linear and the schema consists of a single relation.
\end{lemma}
\begin{proof}
    We give a polynomial-time Turing reduction from the \#P-hard problem
    \#SUBSETSUM. An instance consists of a multiset
    $B=\{\!\{b_1,\dots,b_n\}\!\}$ of positive integers and a target integer $T$,
    and the task is to count the subsets $B'\subseteq B$ whose elements
    sum to $T$. We may assume that $\sum_{i=1}^n b_i\ge T$, since
    otherwise the answer is $0$.

    For each $m\in\{0,\dots,n\}$, we construct a database $D_m$ over
    the same schema $\mathcal{S}$, which consists of a single relation $R$.
    The relation $D_m(R)$ contains one tuple $t^{(m)}$; its primary-key
    attribute has an arbitrary value, and there are no foreign-key
    dependencies. The fixed set of data attributes contains $2n+2$
    numerical attributes:
    \begin{enumerate}
        \item \textbf{Item attributes:} $A_1,\dots,A_n$, where
        $t^{(m)}[A_i]=b_i$.
        \item \textbf{Probe attribute:} $A_{\mathsf{probe}}$, where
        $t^{(m)}[A_{\mathsf{probe}}]=1$.
        \item \textbf{Offset attribute:} $A_T$, where
        $t^{(m)}[A_T]=T$.
        \item \textbf{Heavy and padding attributes:} $A_{H_1},\dots,A_{H_n}$,
        where
        \[
            t^{(m)}[A_{H_j}]
            =
            \begin{cases}
                b_H=\sum_{i=1}^n b_i+T+100, & j\le m,\\
                0, & j>m.
            \end{cases}
        \]
    \end{enumerate}
    Thus, the first $m$ of these attributes are heavy and the remaining
    $n-m$ are padding attributes. The evaluation set is
    $V^{(m)}=\{t^{(m)}\}$, and let $D_{m,S}$ denote the perturbed
    database obtained by applying Zero ablation to every data attribute
    outside the coalition $S$.

    We use the same linear prediction model $f$ for every $D_m$.
    Specifically, $f$ assigns weight $+1$ to every item attribute and to
    the probe attribute, and weight $-1$ to the offset and all
    heavy/padding attributes. The padding attributes have value $0$, so
    they do not affect any prediction. Deleting them from a uniformly
    random permutation leaves a uniformly random permutation of the
    remaining $N_m:=n+m+2$ attributes, so the Shapley calculation below
    is unchanged. Let $\nu^{(m)}$ denote the characteristic function
    induced by $D_m$, $f$, and $V^{(m)}$. Since
    $V^{(m)}=\{t^{(m)}\}$, it is simply the absolute prediction shift
    on $t^{(m)}$. Hence, for every coalition $S$ after omitting the
    padding attributes,
    \begin{align*}
        \nu^{(m)}(S)
        &=
        \left|
            f(t^{(m)},D_m)
            -
            f(t^{(m)},D_{m,S})
        \right|\\
        &=
        \Bigg|
            \left(
                \sum_{i=1}^n b_i
                +1
                -T
                -m b_H
            \right)\\
            &-
            \left(
                \sum_{A_i\in S} b_i
                +\mathbbm{1}[A_{\mathsf{probe}}\in S]
                -T\cdot\mathbbm{1}[A_T\in S]
                -\sum_{A_{H_j}\in S} b_H
            \right)
        \Bigg|\\
        &=
        \Bigg|
            \sum_{A_i\notin S} b_i
            +\mathbbm{1}[A_{\mathsf{probe}}\notin S]
            -T\cdot\mathbbm{1}[A_T\notin S]
            -\sum_{A_{H_j}\notin S} b_H
        \Bigg|.
    \end{align*}
    Define
    \[
        X_S^{(m)}
        :=
        \sum_{A_i\notin S} b_i
        -T\cdot\mathbbm{1}[A_T\notin S]
        -\sum_{A_{H_j}\notin S} b_H.
    \]
    Thus,
    \[
        \nu^{(m)}(S)
        =
        \left|
            X_S^{(m)}
            +
            \mathbbm{1}[A_{\mathsf{probe}}\notin S]
        \right|.
    \]

    We now compute the Shapley value of $A_{\mathsf{probe}}$. Fix a
    predecessor coalition
    \[
        S\subseteq
        \{A_1,\dots,A_n,A_{\mathsf{probe}},A_T,A_{H_1},\dots,A_{H_m}\}
        \setminus\{A_{\mathsf{probe}}\}.
    \]
    Its marginal contribution is
    \[
        \Delta_{\mathsf{probe}}(S)
        =
        \nu^{(m)}(S\cup\{A_{\mathsf{probe}}\})
        -
        \nu^{(m)}(S)
        =
        |X_S^{(m)}|-|X_S^{(m)}+1|.
    \]
    Since $X_S^{(m)}$ is an integer,
    \[
        \Delta_{\mathsf{probe}}(S)
        =
        \begin{cases}
            -1, & X_S^{(m)}\ge 0,\\
            \phantom{-}1, & X_S^{(m)}\le -1.
        \end{cases}
    \]

    We next characterize exactly when $X_S^{(m)}\ge0$. If any heavy
    attribute is excluded from $S$, then
    \[
        X_S^{(m)}
        \le
        \sum_{i=1}^n b_i-b_H
        <0.
    \]
    Thus, all heavy attributes must belong to $S$. Subject to this
    condition, if $A_T\in S$, then
    \[
        X_S^{(m)}
        =
        \sum_{A_i\notin S}b_i
        \ge0.
    \]
    If instead $A_T\notin S$, then
    \[
        X_S^{(m)}
        =
        \sum_{A_i\notin S}b_i-T,
    \]
    which is nonnegative exactly when the excluded item attributes
    have total value at least $T$. Therefore, $X_S^{(m)}\ge0$ if and
    only if all heavy attributes are in $S$ and either
    (i) $A_T\in S$, or
    (ii) $A_T\notin S$ and the excluded item attributes sum to at
    least $T$.

    For a predecessor coalition $S$ as above, the Shapley formula assigns its
    marginal contribution the weight
    \[
        \gamma^{(m)}(S)
        :=
        \frac{|S|!(N_m-|S|-1)!}{N_m!}.
    \]
    Equivalently, $\gamma^{(m)}(S)$ is the probability that, in a
    uniformly random permutation of all $N_m$ attributes, the set of
    attributes appearing before $A_{\mathsf{probe}}$ is exactly $S$.
    Hence, the weights $\gamma^{(m)}(S)$ over all predecessor
    coalitions sum to $1$.

    Let
    \[
        \Gamma_{\ge T}^{(m)}
        :=
        \sum_{\substack{
            S\subseteq
            \{A_1,\dots,A_n,A_T,A_{H_1},\dots,A_{H_m}\}\\
            X_S^{(m)}\ge 0
        }}
        \gamma^{(m)}(S)
    \]
    be the total Shapley weight of the predecessor coalitions for which
    $X_S^{(m)}\ge0$. As shown above, the marginal contribution of
    $A_{\mathsf{probe}}$ is $-1$ for precisely these coalitions and is
    $1$ for all remaining predecessor coalitions. Therefore,
    \[
        W^{(m)}(A_{\mathsf{probe}})
        =
        -\Gamma_{\ge T}^{(m)}
        +
        \bigl(1-\Gamma_{\ge T}^{(m)}\bigr)
        =
        1-2\Gamma_{\ge T}^{(m)},
    \]
    and therefore
    \[
        \Gamma_{\ge T}^{(m)}
        =
        \frac{1-W^{(m)}(A_{\mathsf{probe}})}{2}.
    \]

    For $k\in\{0,\dots,n\}$, let $c_{\ge T}^k$ denote the number of
    subsets $B'\subseteq B$ of size $k$ whose elements sum to at least
    $T$. We now express $\Gamma_{\ge T}^{(m)}$ in terms of these
    unknown counts.

    First, consider predecessor coalitions that contain $A_T$ and all
    $m$ heavy attributes. For such a coalition, let $k$ be the number
    of item attributes excluded from $S$. Then $S$ contains the
    remaining $n-k$ item attributes, and hence
    \[
        |S|=m+n-k+1.
    \]
    Its Shapley weight is therefore
    \[
        \frac{(m+n-k+1)!k!}{(n+m+2)!}.
    \]
    There are $\binom{n}{k}$ choices for the $k$ excluded item
    attributes. Moreover, when $A_T\in S$ and all heavy attributes
    belong to $S$, we have $X_S^{(m)}\ge0$ regardless of which item
    attributes are excluded. Hence, the total contribution of these
    coalitions to $\Gamma_{\ge T}^{(m)}$ is the known quantity
    \[
        K_{\mathsf{base}}^{(m)}
        :=
        \sum_{k=0}^n
        \binom{n}{k}
        \frac{(m+n-k+1)!k!}{(n+m+2)!}.
    \]

    Now consider predecessor coalitions that exclude $A_T$, contain
    all heavy attributes, and exclude exactly $k$ item attributes.
    Such a coalition has size $m+n-k$. Consequently, its Shapley
    weight is
    \[
        C_{m,k}
        :=
        \frac{(m+n-k)!(k+1)!}{(n+m+2)!}.
    \]
    By the characterization of $X_S^{(m)}\ge0$, precisely
    $c_{\ge T}^k$ choices of the $k$ excluded item attributes
    contribute. Therefore,
    \[
        \Gamma_{\ge T}^{(m)}
        =
        K_{\mathsf{base}}^{(m)}
        +
        \sum_{k=0}^n C_{m,k}\,c_{\ge T}^k.
    \]

    For each $m\in\{0,\dots,n\}$, we therefore obtain the equation
    \[
        \Gamma_{\ge T}^{(m)}
        -
        K_{\mathsf{base}}^{(m)}
        =
        \sum_{k=0}^n C_{m,k}\,c_{\ge T}^k.
    \]
    Evaluating the Shapley value of the probe for each
    $m=0,\dots,n$ gives $n+1$ linear equations in the $n+1$
    unknowns
    \[
        c_{0,\ge T},\dots,c_{n,\ge T}.
    \]
    Let $C$ be the $(n+1)\times(n+1)$ coefficient matrix whose
    $(m,k)$-entry is $C_{m,k}$. To recover the unknown counts
    $c_{\ge T}^k$, it remains to show that $C$ is invertible.
    Multiply row $m$ by $(n+m+2)!$, divide column $k$ by $(k+1)!$,
    and then reverse the order of the columns. These are all
    invertible operations, and they transform $C$ into the matrix
    $H$ with
    \[
        H_{m,k}=(m+k)!,
        \qquad 0\le m,k\le n.
    \]
    As shown in~\cite{determinants2002}, the determinant of $H$ is
    \[
        \det(H)
        =
        \prod_{i=0}^{n}(i!)^2 
        \neq 0.
    \]
    Hence $H$, and therefore $C$, is invertible. We can thus recover
    all the values $c_{\ge T}^k$.

    To recover the number of subsets of a given size summing
    \emph{exactly} to $T$, repeat the same computation with threshold
    $T+1$, obtaining $c_{k,\ge T+1}$, and set
    \[
        c_k
        =
        c_{\ge T}^k
        -
        c_{k,\ge T+1}.
    \]
    If $\sum_{i=1}^n b_i<T+1$, then all $c_{k,\ge T+1}$ are simply
    $0$. Therefore, the number of subsets of $B$ whose elements sum
    exactly to $T$ is
    \[
        \sum_{k=0}^n
        \left(
            c_{\ge T}^k
            -
            c_{k,\ge T+1}
        \right).
    \]
    The construction uses polynomially many oracle calls, all
    numerical values have polynomial bit length, and the resulting
    rational linear systems can be solved in polynomial time. This
    concludes our proof.
\end{proof}

\paragraph{Mean imputation.}
The same reduction works when an excluded numerical attribute is replaced by its mean rather than by $0$. For each database $D_m$ above, add two auxiliary tuples to $R$, neither of which belongs to the evaluation set $V^{(m)}$: in the first, every data attribute whose value in $t^{(m)}$ is $v$ is assigned the value $-v$, and in the second, every data attribute is assigned the value $0$. Give the auxiliary tuples fresh primary-key values. Thus, for every data attribute, its three values in $R$ are $v,-v,0$, and hence its mean is $0$. Mean imputation therefore induces exactly the same perturbed values on the evaluated tuple $t^{(m)}$ as Zero ablation. The two auxiliary tuples do not affect the prediction on $t^{(m)}$, because the linear model used in the reduction depends only on the attributes of the tuple being evaluated. Consequently, the characteristic function and the Shapley value of $A_{\mathsf{probe}}$ are unchanged.

\paragraph*{Hardness for bounded arity.} While the previous lemma establishes hardness for a schema
consisting of a single relation, we now show that computing the
Shapley value remains \#P-hard even when the number of attributes
in each relation is bounded by a constant, while the number of
relations is unbounded. For this result, the prediction model must
aggregate information across relations. We therefore first formalize
the message-passing GNN (MPGNN) architecture used in the
reduction.

Given a database $D$ over $\mathcal{S}$, we
represent $D$ as a heterogeneous graph in the standard manner:
every tuple is represented by a node, whose type is the relation
containing the tuple. For every foreign-key dependency
\[
    \lambda : R[X] \sqsubseteq \operatorname{key}(R')
\]
and every pair of tuples $u\in D(R)$ and $u'\in D(R')$ satisfying
\[
    u[X]=u'[\operatorname{key}(R')],
\]
the graph contains an edge from $u$ to $u'$. We allow message
passing both along this edge and in the reverse direction. Thus, a
message-passing edge type is determined by a foreign-key dependency
$\lambda$ together with an orientation
$o\in\{\mathsf{in},\mathsf{out}\}$.

An MPGNN consists of $K$ message-passing layers. For every tuple
$u$ and every layer $\ell\in\{0,\dots,K\}$, let
$z_u^{(\ell)}$ denote the representation of the node corresponding
to $u$. The initial representation is obtained using a
relation-specific encoder applied to the data attributes of the
tuple. Thus, for a tuple $u$ of relation $R$,
\[
    z_u^{(0)}=\mathsf{Enc}_R(u).
\]

For every layer $\ell\in\{1,\dots,K\}$, messages are aggregated
separately for each foreign-key dependency and orientation. Let
$\mathcal{N}_{\lambda}^{o}(u)$ denote the neighbors of $u$
connected through dependency $\lambda$ in orientation $o$. We
define
\[
    m_{\lambda,o}^{(\ell)}(u)
    =
    \mathsf{AGG}_{\lambda,o}^{(\ell)}
    \left(
        \{\!\{
            z_{u'}^{(\ell-1)}
            \mid
            u'\in\mathcal{N}_{\lambda}^{o}(u)
        \}\!\}
    \right),
\]
where $\mathsf{AGG}_{\lambda,o}^{(\ell)}$ is a
permutation-invariant aggregation function.

The updated representation of $u$ combines its
previous representation with the aggregated messages for the
distinct edge types:
\[
    z_u^{(\ell)}
    =
    \mathsf{UPD}_{R}^{(\ell)}
    \left(
        z_u^{(\ell-1)},
        \left\{
            m_{\lambda,o}^{(\ell)}(u)
        \right\}_{\lambda\in\operatorname{FK}(\mathcal{S}),
                   \,o\in\{\mathsf{in},\mathsf{out}\}}
    \right),
\]
where $R$ is the relation containing $u$ and
$\mathsf{UPD}_{R}^{(\ell)}$ is a learnable update function.

Finally, for a target tuple $t$ of the target relation, the
prediction is obtained from its final representation through a
task-specific prediction head:
\[
    f(t,D)=\mathsf{Head}\left(z_t^{(K)}\right).
\]

In the reduction below, we use a particularly simple instance of
this architecture: $K=1$, the encoders are identity maps, every
aggregation function is $\mathsf{SUM}$, and both the update
function and the prediction head are linear.

\begin{lemma}\label{lem:shapley_hardness_bounded}
   Computing the Shapley value of an attribute under the global
characteristic function $\nu$ is \#P-hard when the schema
consists of multiple relations, all of bounded arity, and the MPGNN
consists of a single layer without any non-linear activations.
\end{lemma}
\begin{proof}
    We reduce from \#SUBSETSUM. Let
    $B=\{\!\{b_1,\dots,b_n\}\!\}$ and $T$ be the input instance. For every
    $m\in\{0,\dots,n\}$, we construct a database $D_m$ over the same
    fixed star-shaped schema $\mathcal{S}$. The center is a
    target relation $R$ containing a single tuple $t^{(m)}$.
    In addition to its primary-key attribute, $R$ has one data
    attribute $A_{\mathsf{probe}}$, with
    \[
        t^{(m)}[A_{\mathsf{probe}}]=1.
    \]
    All remaining data attributes occur in separate relations, each
    containing a single tuple that references $t^{(m)}$ through a
    foreign-key dependency. Specifically, $D_m$ contains:
    \begin{enumerate}
        \item \textbf{Item relations:}
        for every $i\in\{1,\dots,n\}$, a relation $R_i$ whose
        unique tuple has a data attribute $A_i$ with value $b_i$;

        \item \textbf{Offset relation:}
        a relation $R_T$ whose unique tuple has a data attribute
        $A_T$ with value $T$;

        \item \textbf{Heavy and padding relations:}
        for every $j\in\{1,\dots,n\}$, a relation $R_{H_j}$ whose
        unique tuple has a data attribute $A_{H_j}$ with value $b_H$
        if $j\le m$ and value $0$ otherwise, where
        \[
            b_H=\sum_{i=1}^n b_i+T+100.
        \]
    \end{enumerate}
    Thus, the first $m$ of these relations are heavy and the remaining
    $n-m$ are padding relations. Each non-target relation has a
    single-attribute primary key and a single-attribute foreign key
    referencing the primary key of $R$. The foreign-key value of its
    unique tuple is set to the primary-key value of $t^{(m)}$. Hence
    $D_m$ is a valid database, and its tuple-level graph is a star
    centered at $t^{(m)}$. Each non-target relation has three attributes:
    a primary key, a foreign key, and one data attribute, while $R$ has
    two. Thus, every relation has at most three attributes, independently
    of $n$ and $m$, and the schema is identical for every $D_m$.

    The evaluation set is $V^{(m)}=\{t^{(m)}\}$, and let $D_{m,S}$
    denote the perturbed database obtained by applying Zero ablation to
    every data attribute outside the coalition $S$.
    In particular, structural key attributes are not ablated, so the
    relational topology is unchanged.

    We use the same one-layer MPGNN $f$ for every $D_m$. It has the
    form described above: its relation-specific encoders are identity
    maps and its aggregation functions are $\mathsf{SUM}$. Let
    $\lambda_i$, $\lambda_T$, and $\lambda_{H_j}$ denote the
    foreign-key dependencies from the corresponding non-target
    relations to $R$.

    Fix a coalition $S$, and consider the
    perturbed database $D_{m,S}$. Let $u_i$, $u_T$, and $u_{H_j}$
    denote the unique tuples of $R_i$, $R_T$, and $R_{H_j}$,
    respectively. Since each of these relations contains a single
    tuple, the identity encoders and $\mathsf{SUM}$ aggregation give
    \begin{align*}
        m_{\lambda_i,\mathsf{in}}^{(1)}(t^{(m)})
        &=
        z_{u_i}^{(0)}
        =
        b_i\,\mathbbm{1}[A_i\in S],
        && i\in\{1,\dots,n\},\\
        m_{\lambda_T,\mathsf{in}}^{(1)}(t^{(m)})
        &=
        z_{u_T}^{(0)}
        =
        T\,\mathbbm{1}[A_T\in S],\\
        m_{\lambda_{H_j},\mathsf{in}}^{(1)}(t^{(m)})
        &=
        z_{u_{H_j}}^{(0)}
        =
        \begin{cases}
            b_H\,\mathbbm{1}[A_{H_j}\in S], & j\le m,\\
            0, & j>m,
        \end{cases}
        && j\in\{1,\dots,n\}.
    \end{align*}
    Similarly, the initial representation of the target tuple is
    \[
        z_{t^{(m)}}^{(0)}
        =
        \mathbbm{1}[A_{\mathsf{probe}}\in S].
    \]

    We choose the linear update function at the target tuple to assign
    coefficient $+1$ to its previous representation and to the messages
    from the item relations, and coefficient $-1$ to the messages from
    the offset and heavy/padding relations. Hence,
    \begin{align*}
        z_{t^{(m)}}^{(1)}
        &=
        z_{t^{(m)}}^{(0)}
        +
        \sum_{i=1}^n
        m_{\lambda_i,\mathsf{in}}^{(1)}(t^{(m)})
        -
        m_{\lambda_T,\mathsf{in}}^{(1)}(t^{(m)})
        -
        \sum_{j=1}^n
        m_{\lambda_{H_j},\mathsf{in}}^{(1)}(t^{(m)})\\
        &=
        \mathbbm{1}[A_{\mathsf{probe}}\in S]
        +
        \sum_{A_i\in S} b_i
        -
        T\cdot\mathbbm{1}[A_T\in S]
        -
        \sum_{\substack{1\le j\le m\\ A_{H_j}\in S}} b_H.
    \end{align*}

    Finally, the prediction head applies unit weight. Therefore,
    \[
        f(t^{(m)},D_{m,S})
        =
        \sum_{A_i\in S} b_i
        +
        \mathbbm{1}[A_{\mathsf{probe}}\in S]
        -
        T\cdot\mathbbm{1}[A_T\in S]
        -
        \sum_{\substack{1\le j\le m\\ A_{H_j}\in S}} b_H.
    \]

    The zero-valued padding relations disappear from this expression.
    Thus, this is exactly the coalition-dependent prediction function used
    in the proof of Theorem~\ref{thm:shapley_hardness}. Hence, for every
    coalition $S$, the induced characteristic function $\nu^{(m)}(S)$
    is identical to that in the previous construction. The reduction
    now continues exactly as in the proof of
    Theorem~\ref{thm:shapley_hardness}.
\end{proof}

\paragraph{Mean imputation.}
The reduction also applies when excluded numerical attributes are
replaced by their mean values rather than by zero. For every database
$D_m$ constructed above, we add two auxiliary copies of each tuple,
neither of which belongs to the evaluation set $V^{(m)}$. In the first
copy, every data attribute whose value in the original tuple is $v$ is
assigned the value $-v$; in the second copy, every data attribute is
assigned the value $0$. All auxiliary tuples receive fresh primary-key
values.
We assign the foreign-key values of the auxiliary tuples so that each
negated tuple references the corresponding negated tuple in the
referenced relation, and each zero tuple references the corresponding
zero tuple. Hence, all foreign-key dependencies remain satisfied, and
the auxiliary tuples form two components that are disconnected from
the component containing the evaluated target tuple $t^{(m)}$.

For every numerical data attribute, the three values in its relation
are therefore $v$, $-v$, and $0$, so its mean is $0$. Consequently,
mean imputation has exactly the same effect as Zero ablation on the
component containing $t^{(m)}$. Moreover, since the two auxiliary
components are disconnected from $t^{(m)}$, they do not affect the
MPGNN prediction for $t^{(m)}$. Thus, for every coalition $S$, the perturbed prediction
$f(t^{(m)},D_{m,S})$, and therefore the characteristic function
$\nu^{(m)}(S)$ and the Shapley value of $A_{\mathsf{probe}}$, are
identical to those in the Zero-ablation construction. The same
reduction therefore establishes \#P-hardness under mean imputation.

%% file: appendix_plots.tex
\section{Monte Carlo Shapley Convergence Plots}
\label{app:shapley_convergence}

As discussed in Section~\ref{sub:shapley_explanations}, we employ an empirical termination strategy for the Monte Carlo permutation sampling. The sampling process is empirically terminated once the convergence curve visibly flattens out, indicating that the maximum absolute shift across all attributes has diminished to a negligible noise floor. 

\begin{figure*}[htbp]
    \centering
    
    \begin{subfigure}[b]{0.24\linewidth}
        \centering
        \includegraphics[width=\linewidth]{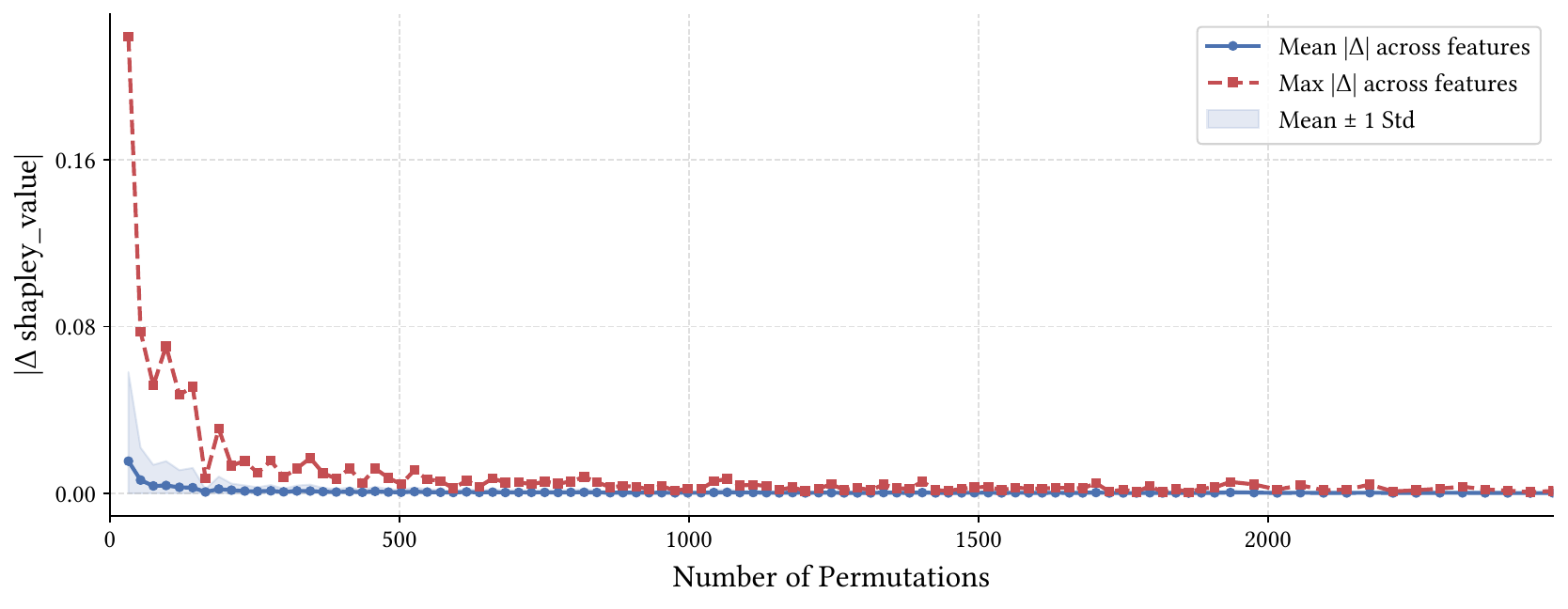}
        \caption{\texttt{rel-f1}, \textit{driver-position} (HGS)}
    \end{subfigure}
    \hfill
    \begin{subfigure}[b]{0.24\linewidth}
        \centering
        \includegraphics[width=\linewidth]{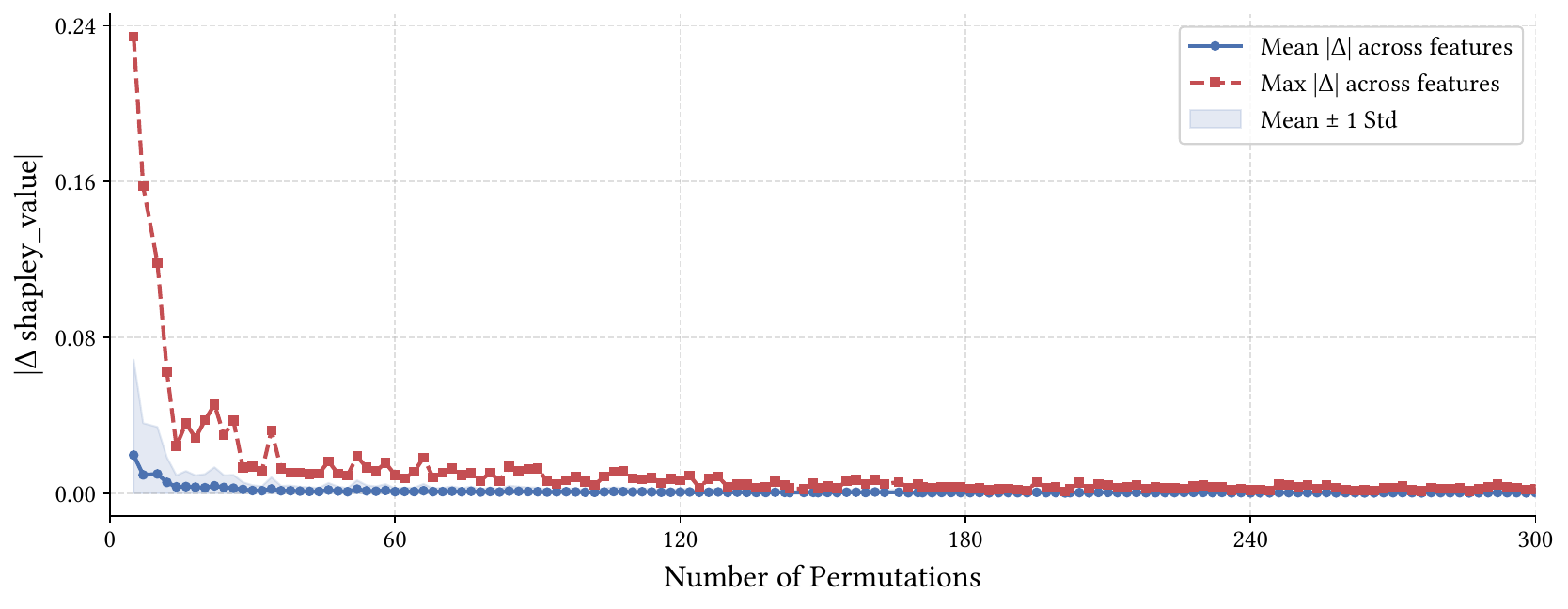}
        \caption{\texttt{rel-f1}, \textit{driver-position} (RelGT)}
    \end{subfigure}
    \hfill
    \begin{subfigure}[b]{0.24\linewidth}
        \centering
        \includegraphics[width=\linewidth]{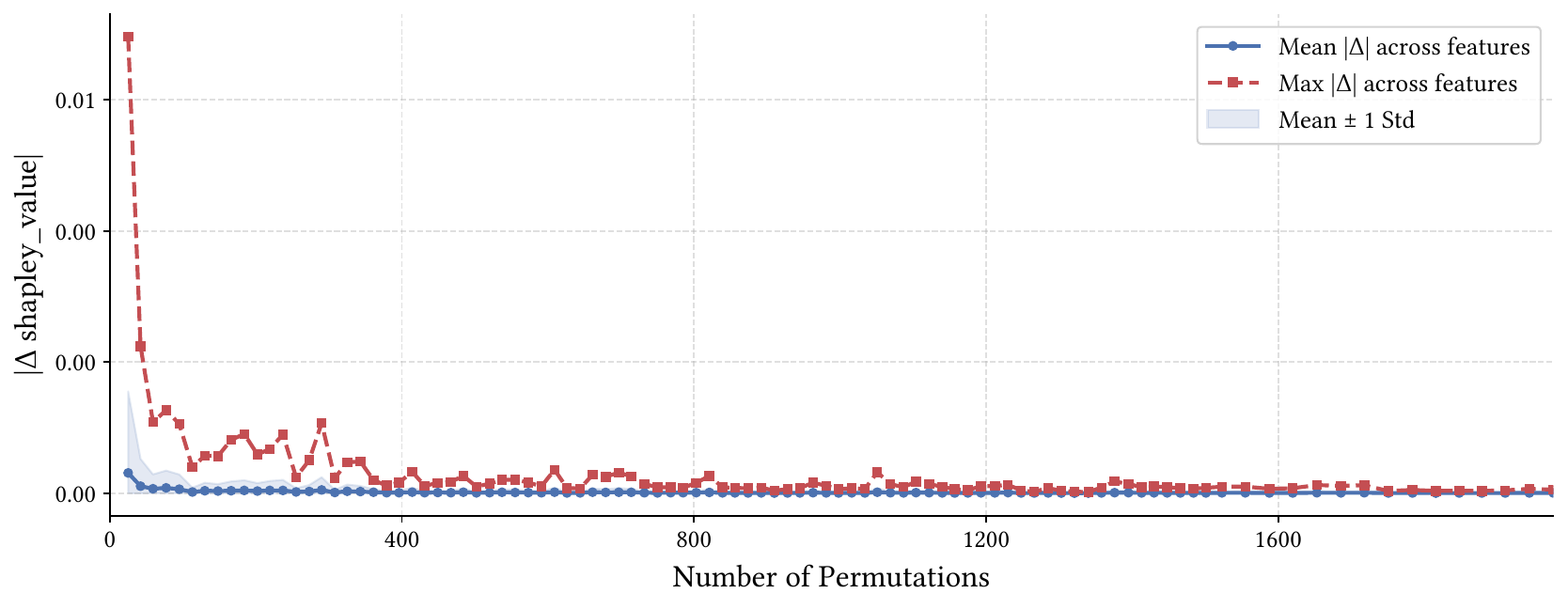}
        \caption{\texttt{rel-f1}, \textit{driver-dnf} (HGS)}
    \end{subfigure}
    \hfill
    \begin{subfigure}[b]{0.24\linewidth}
        \centering
        \includegraphics[width=\linewidth]{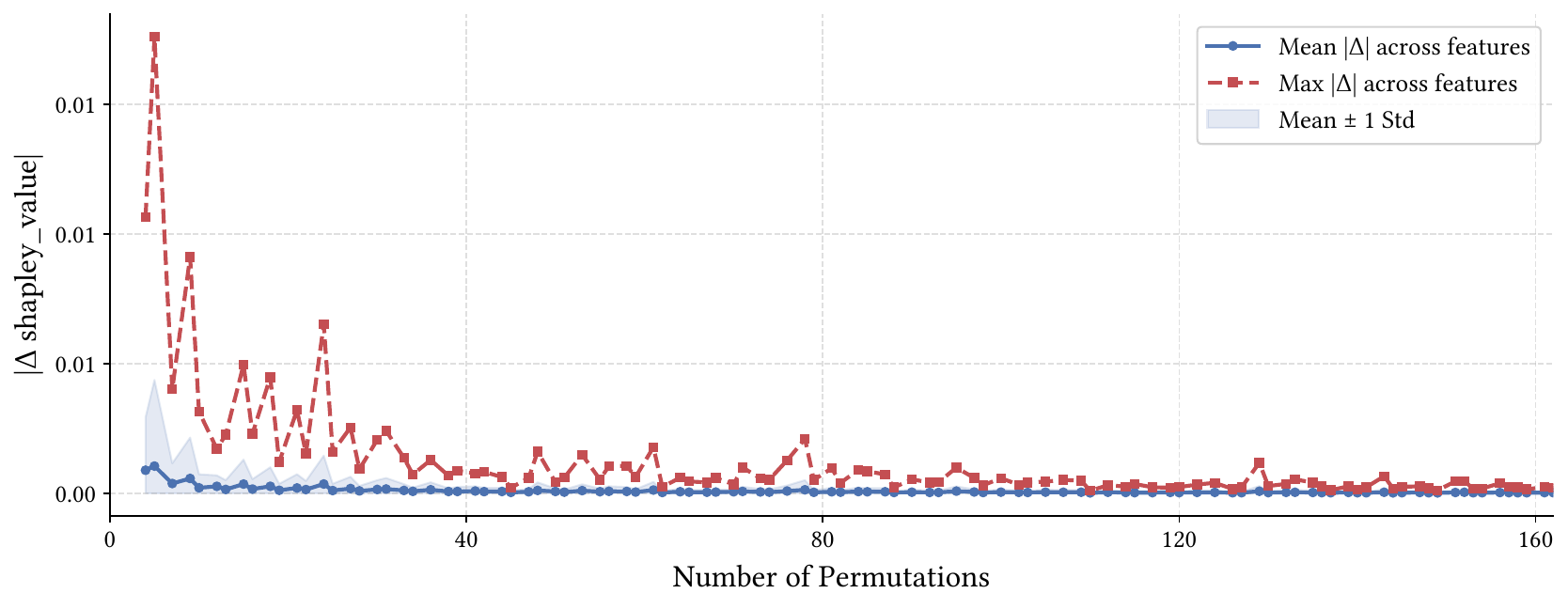}
        \caption{\texttt{rel-f1}, \textit{driver-dnf} (RelGT)}
    \end{subfigure}
    
    \vspace{1em}

    \begin{subfigure}[b]{0.24\linewidth}
        \centering
        \includegraphics[width=\linewidth]{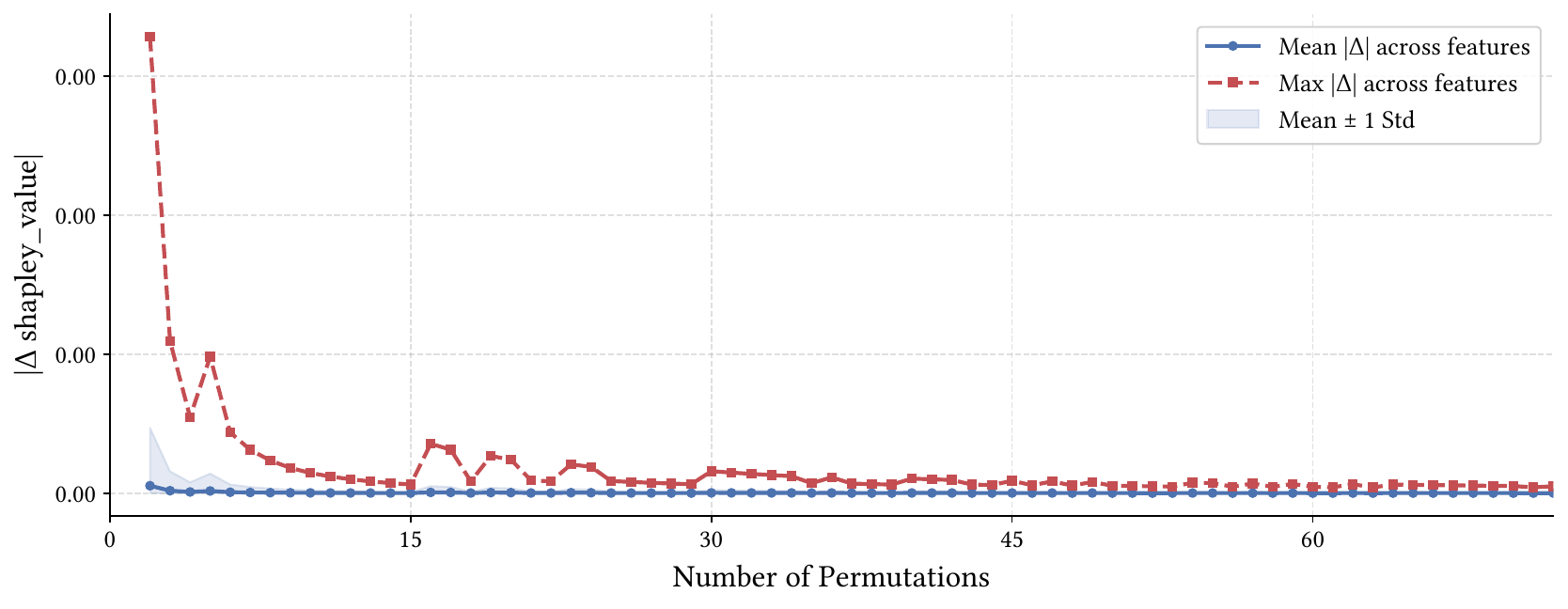}
        \caption{\texttt{rel-event}, \textit{user-attendance} (HGS)}
    \end{subfigure}
    \hfill
    \begin{subfigure}[b]{0.24\linewidth}
        \centering
        \includegraphics[width=\linewidth]{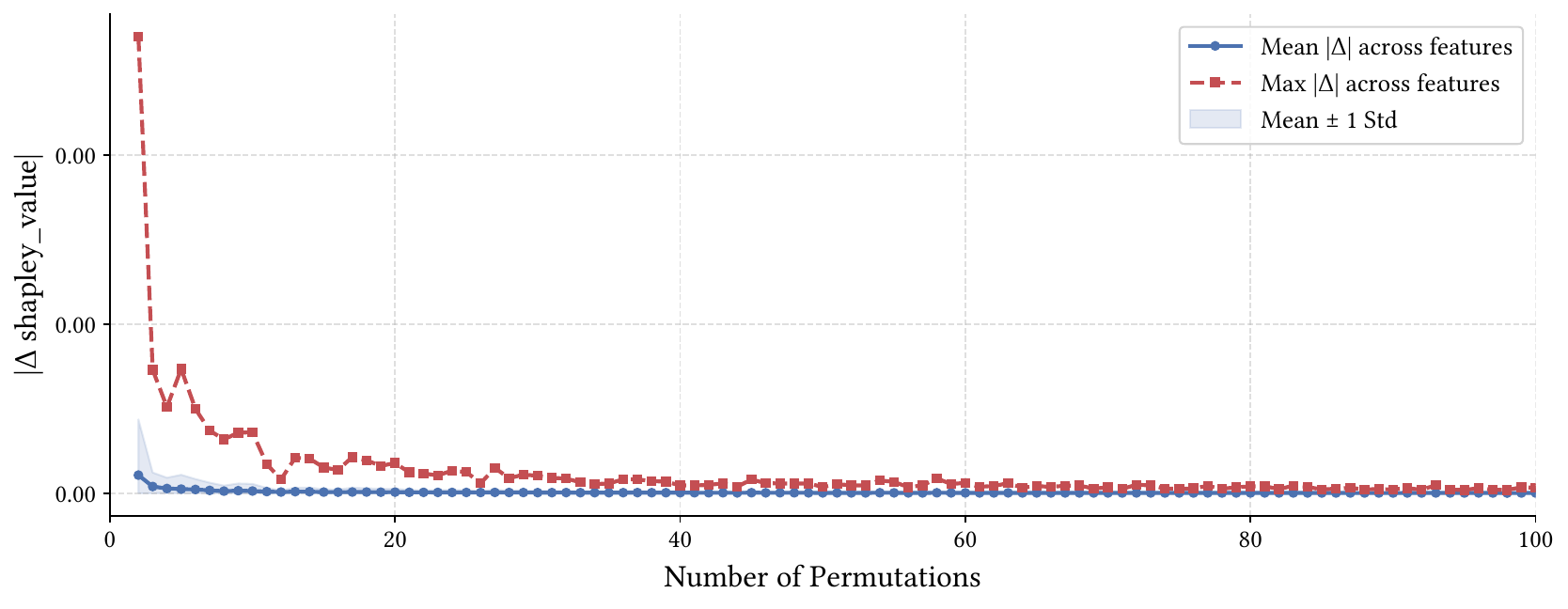}
        \caption{\texttt{rel-event}, \textit{user-attendance} (RelGT)}
    \end{subfigure}
    \hfill
    \begin{subfigure}[b]{0.24\linewidth}
        \centering
        \includegraphics[width=\linewidth]{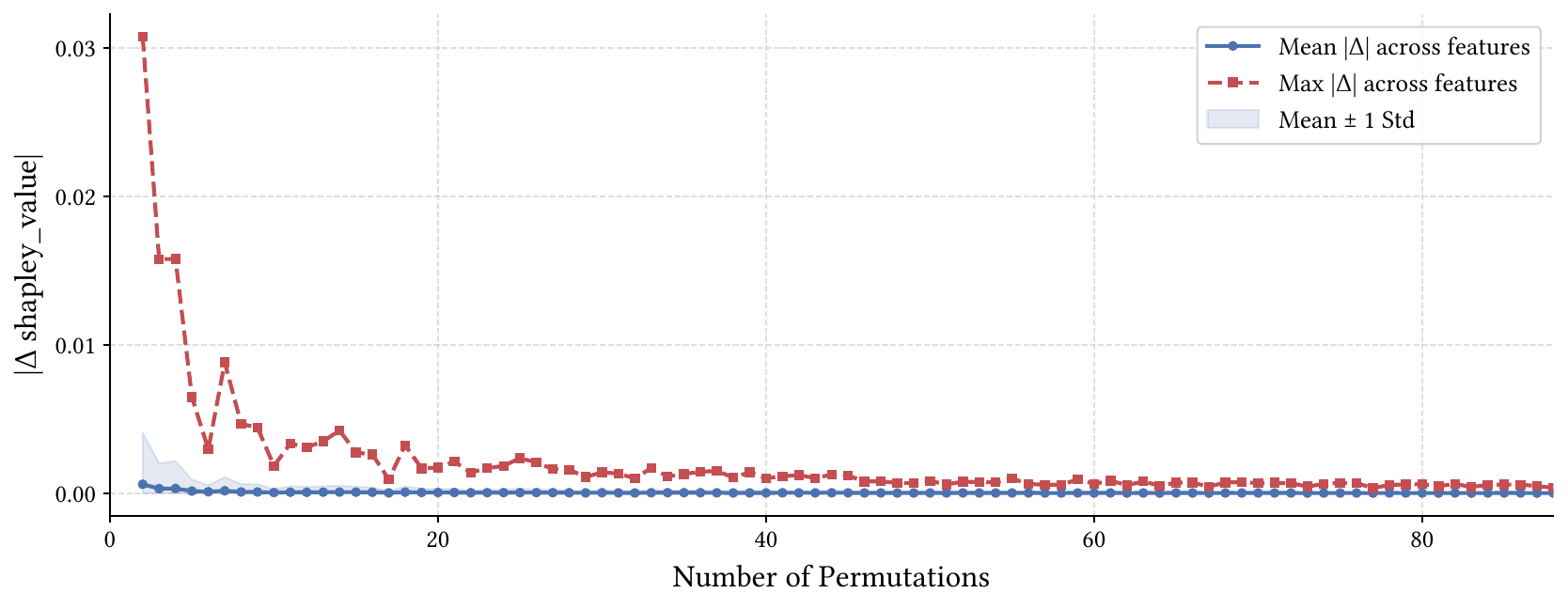}
        \caption{\texttt{rel-event}, \textit{user-repeat} (HGS)}
    \end{subfigure}
    \hfill
    \begin{subfigure}[b]{0.24\linewidth}
        \centering
        \includegraphics[width=\linewidth]{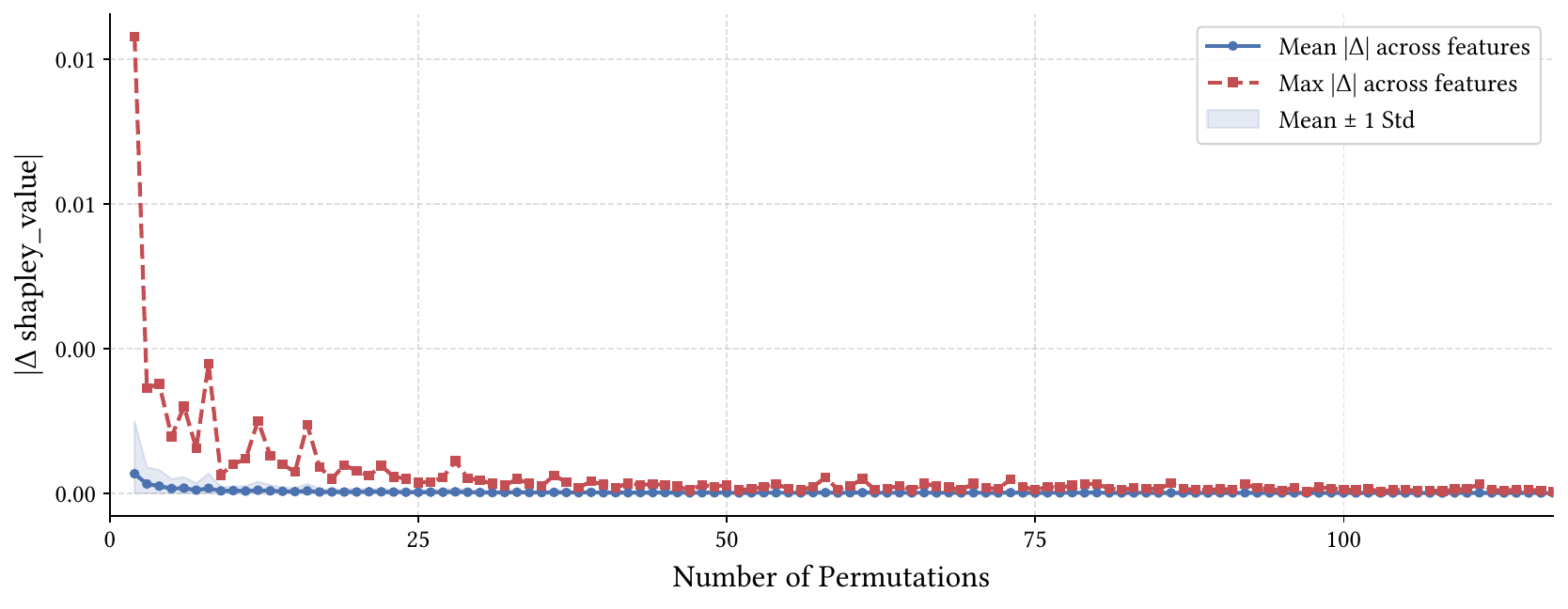}
        \caption{\texttt{rel-event}, \textit{user-repeat} (RelGT)}
    \end{subfigure}

    \caption{Empirical convergence curves for the Monte Carlo Shapley approximation on the \texttt{rel-f1} and \texttt{rel-event} datasets.}
    \label{fig:app_convergence_grid_1}
\end{figure*}

\begin{figure*}[htbp]
    \centering
    
    \begin{subfigure}[b]{0.32\linewidth}
        \centering
        \includegraphics[width=\linewidth]{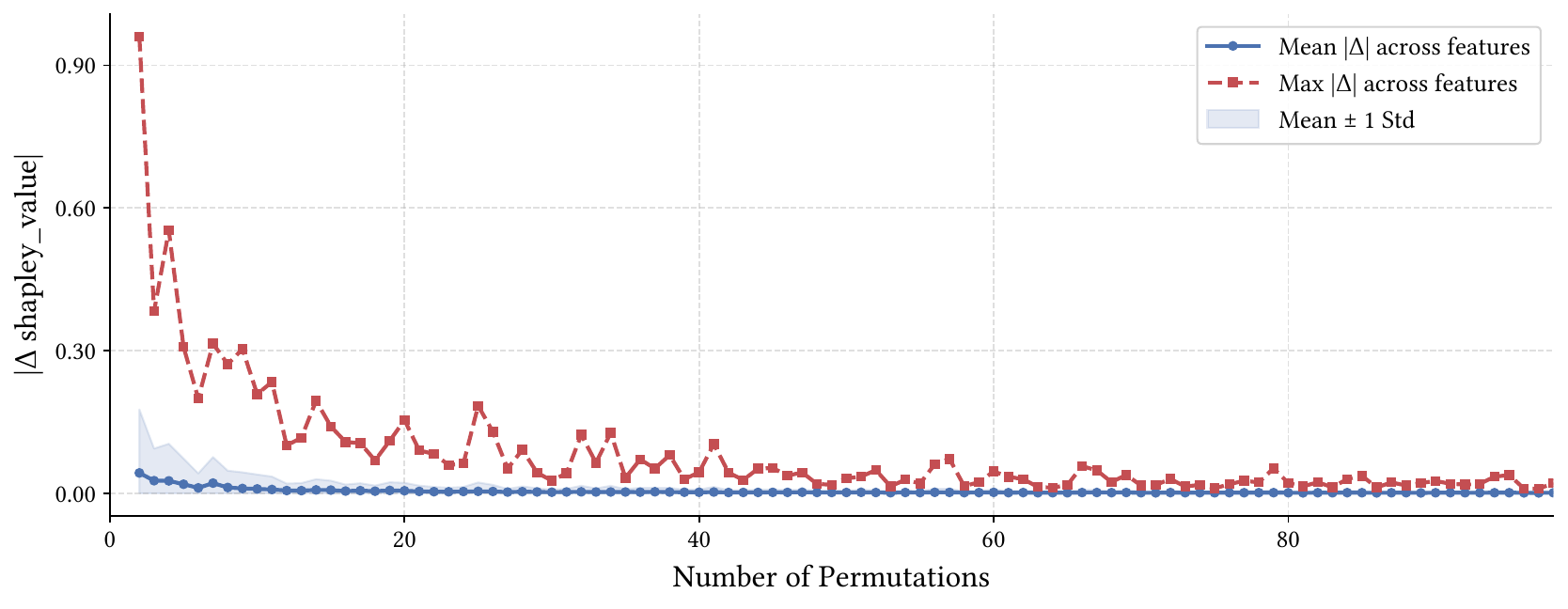}
        \caption{\texttt{rel-trial}, \textit{study-adverse} (RelGT)}
    \end{subfigure}
    \hfill
    \begin{subfigure}[b]{0.32\linewidth}
        \centering
        \includegraphics[width=\linewidth]{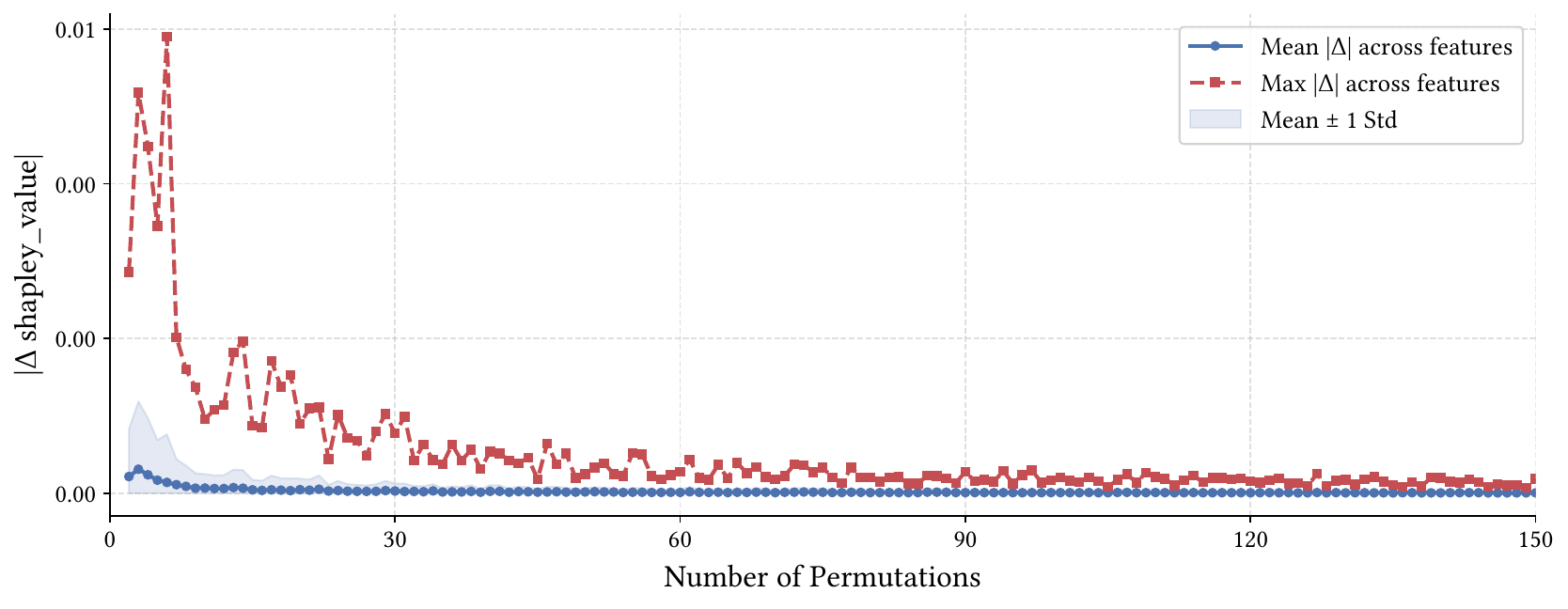}
        \caption{\texttt{rel-trial}, \textit{study-outcome} (HGS)}
    \end{subfigure}
    \hfill
    \begin{subfigure}[b]{0.32\linewidth}
        \centering
        \includegraphics[width=\linewidth]{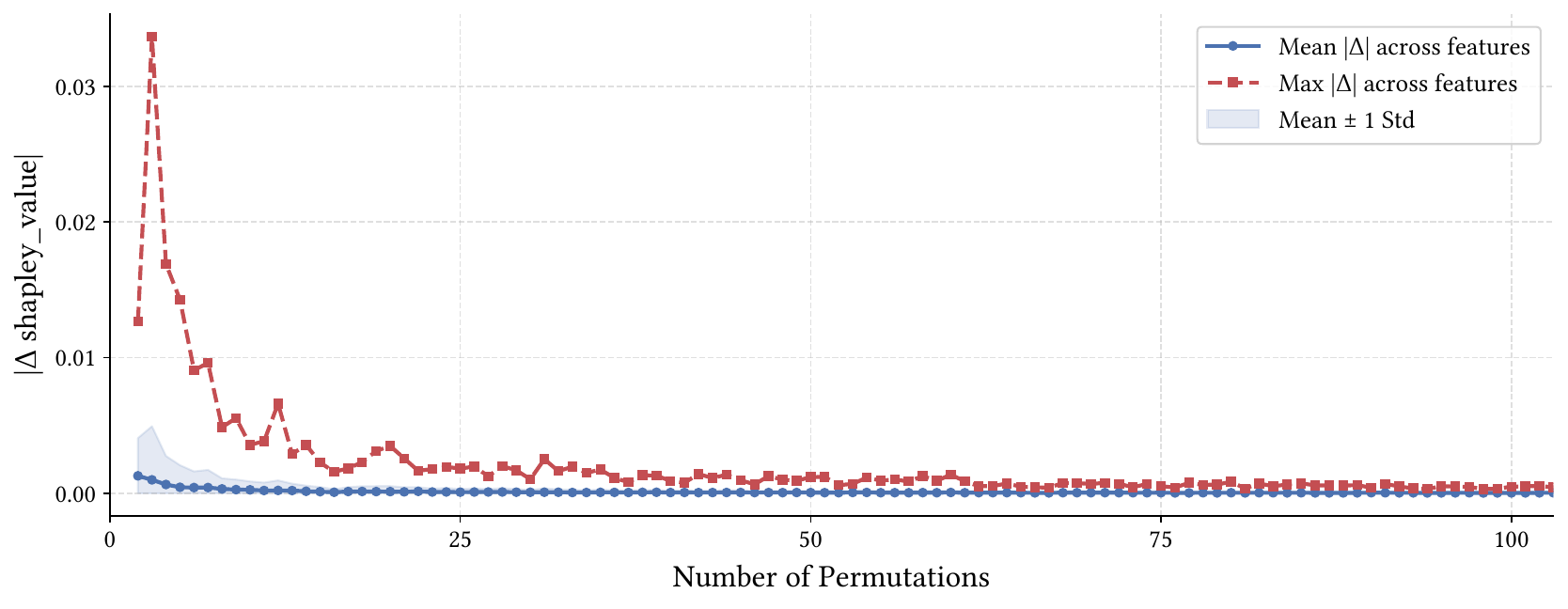}
        \caption{\texttt{rel-trial}, \textit{study-outcome} (RelGT)}
    \end{subfigure}

    \vspace{1em}

    \begin{subfigure}[b]{0.24\linewidth}
        \centering
        \includegraphics[width=\linewidth]{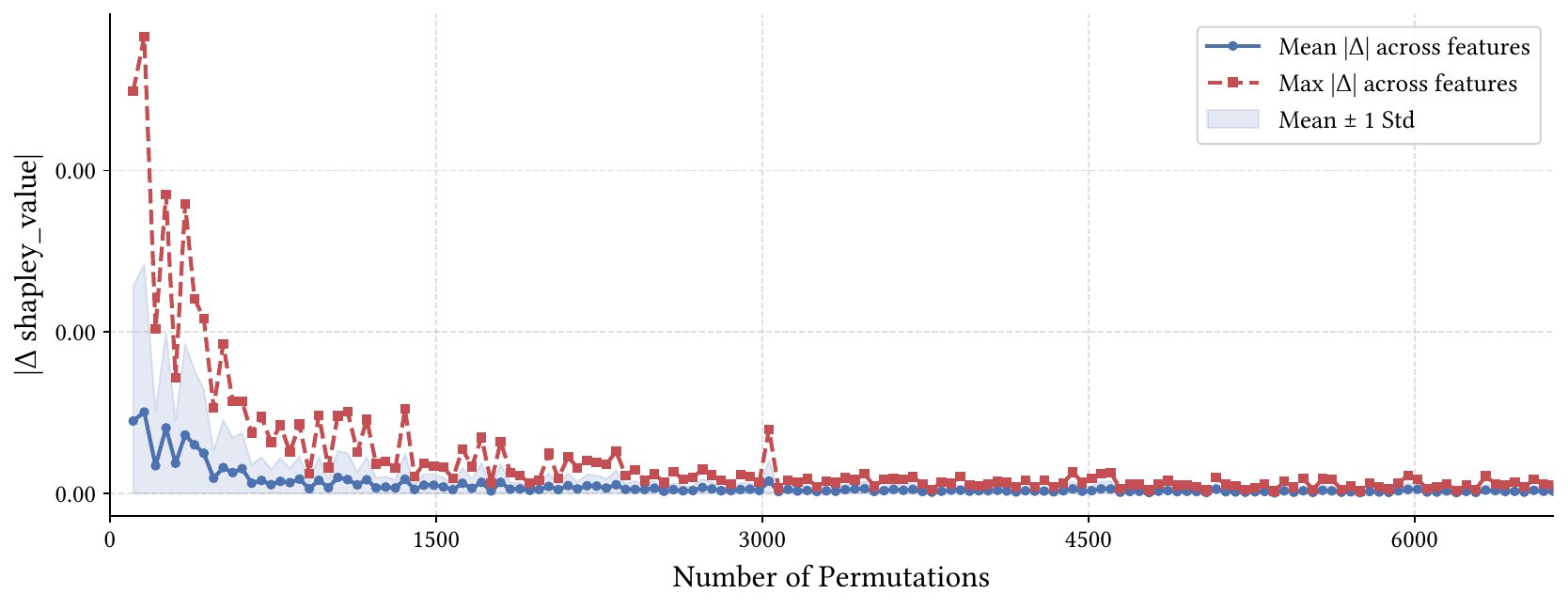}
        \caption{\texttt{rel-arxiv}, \textit{paper-citation} (HGS)}
    \end{subfigure}
    \hfill
    \begin{subfigure}[b]{0.24\linewidth}
        \centering
        \includegraphics[width=\linewidth]{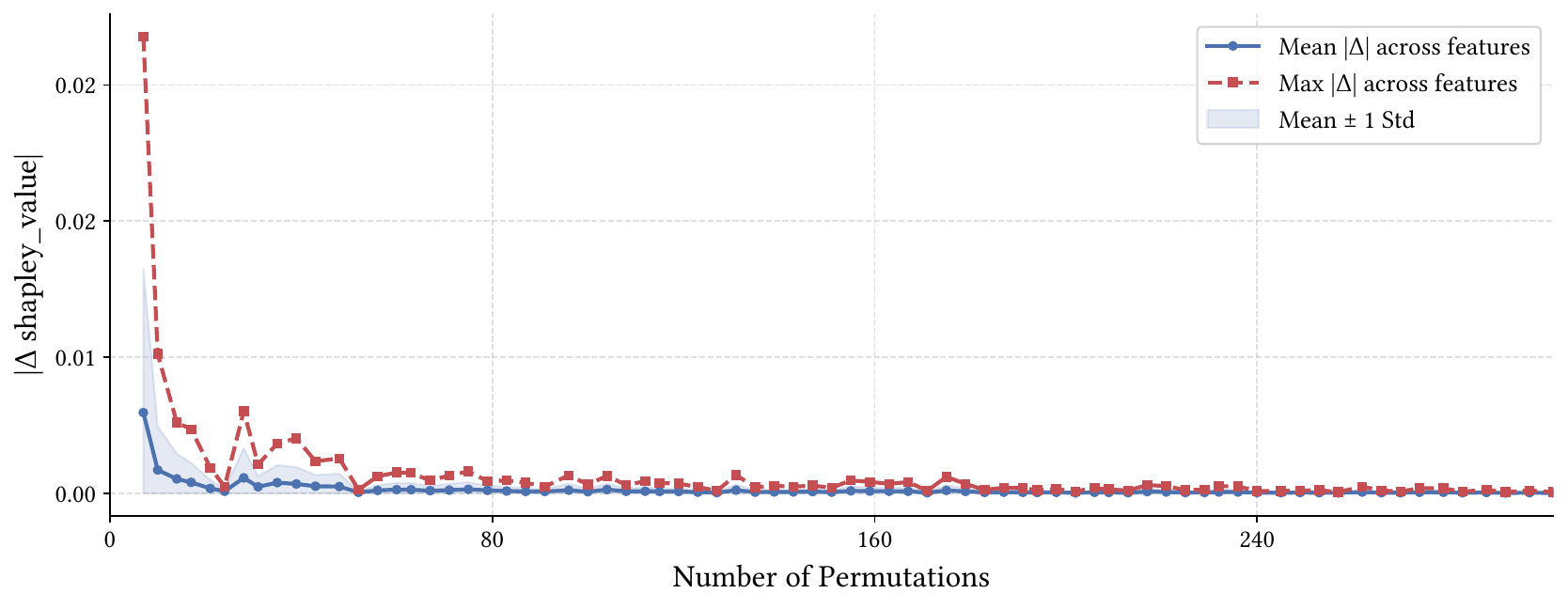}
        \caption{\texttt{rel-arxiv}, \textit{paper-citation} (RelGT)}
    \end{subfigure}
    \hfill
    \begin{subfigure}[b]{0.24\linewidth}
        \centering
        \includegraphics[width=\linewidth]{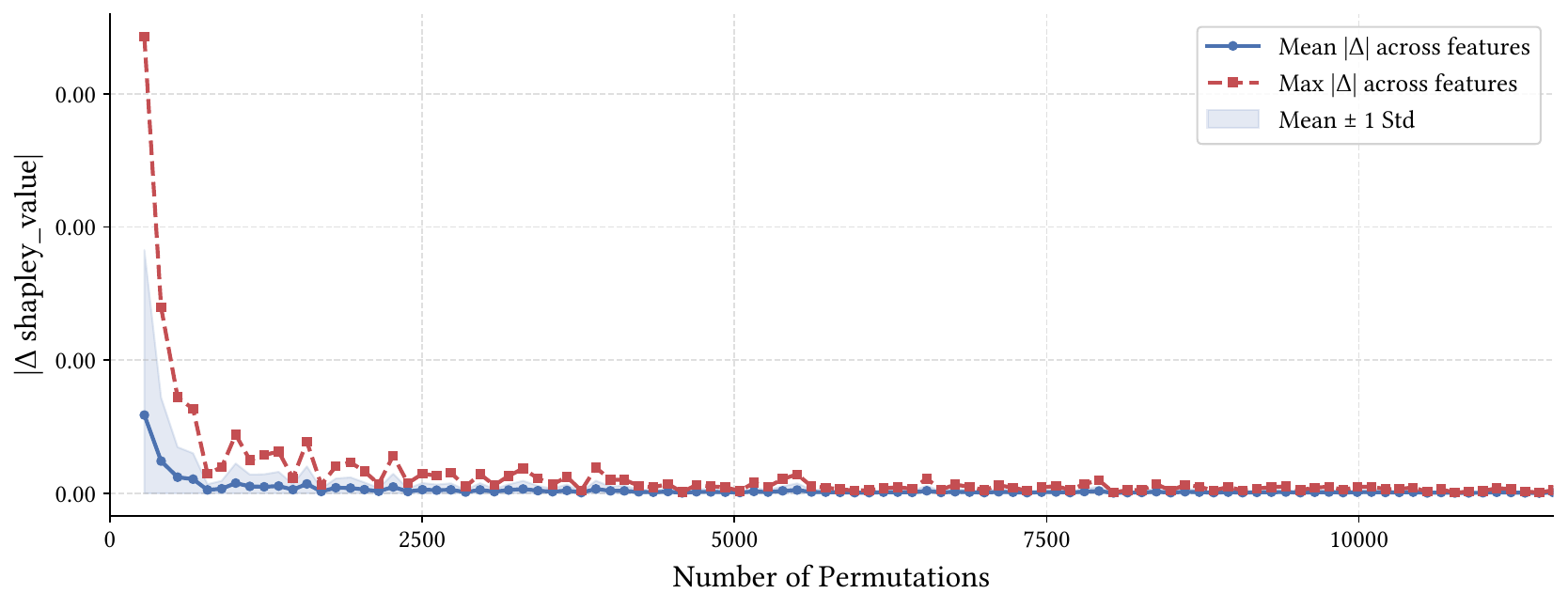}
        \caption{\texttt{rel-arxiv}, \textit{author-publication} (HGS)}
    \end{subfigure}
    \hfill
    \begin{subfigure}[b]{0.24\linewidth}
        \centering
        \includegraphics[width=\linewidth]{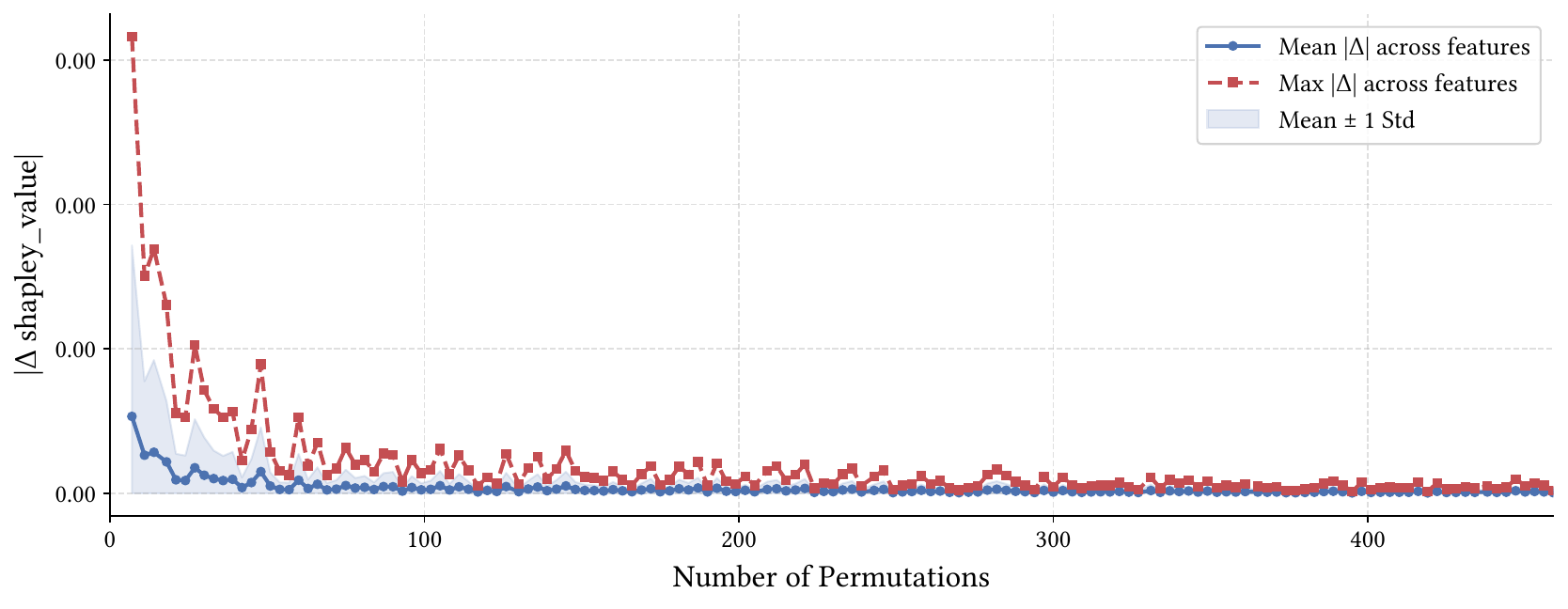}
        \caption{\texttt{rel-arxiv}, \textit{author-publication} (RelGT)}
    \end{subfigure}

    \caption{Empirical convergence curves for the Monte Carlo Shapley approximation on the \texttt{rel-trial} and \texttt{rel-arxiv} datasets.}
    \label{fig:app_convergence_grid_2}
\end{figure*}

\Cref{fig:app_convergence_grid_1,fig:app_convergence_grid_2} present the empirical convergence curves for all remaining tasks and architectures not shown in the main text. Each plot tracks the absolute difference in estimated scores between consecutive permutation checkpoints ($|\Delta \text{shapley\_value}|$).

\section{Kernel SHAP Convergence Plots}
\label{app:kernel_convergence}

As discussed in Section~\ref{subsec:rq1}, to verify that the ranking discrepancies between \kernel and \shapley are not merely artifacts of insufficient sampling, we empirically validated the convergence of the \kernel method across all tasks. 

\begin{figure*}[htbp]
    \centering
    
    \begin{subfigure}[b]{0.24\linewidth}
        \centering
        \includegraphics[width=\linewidth]{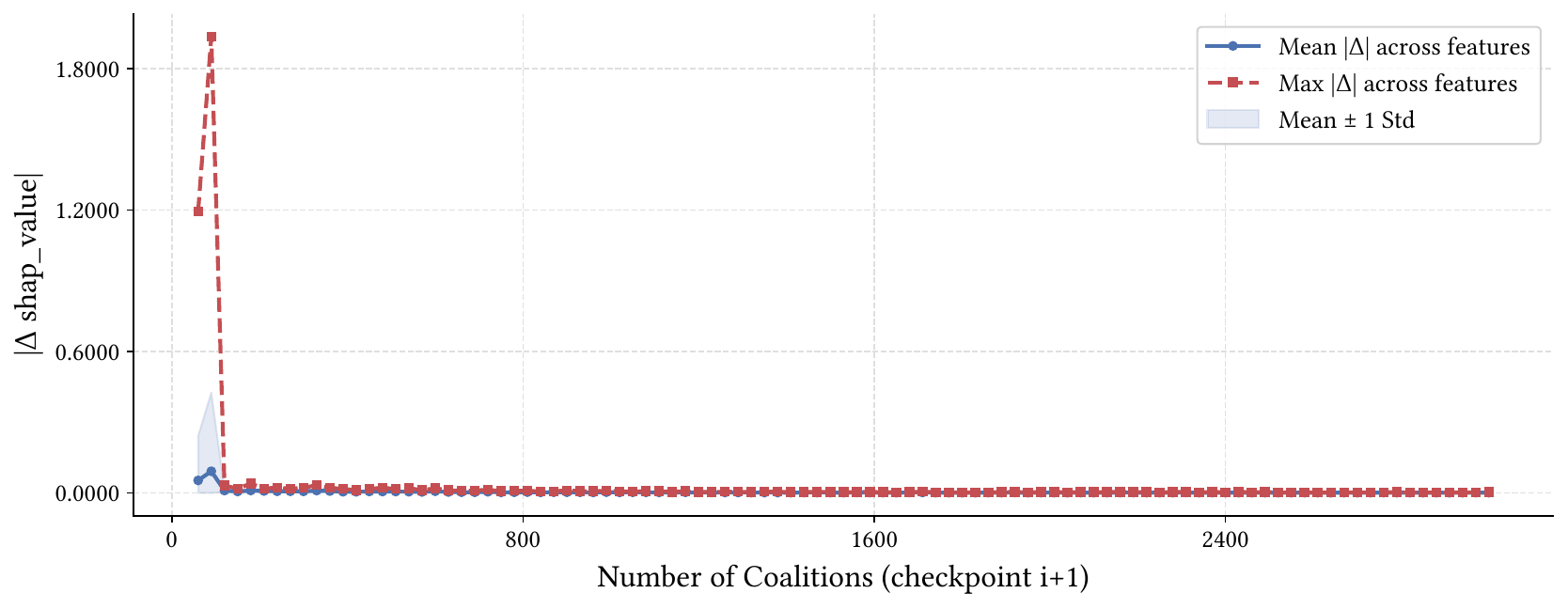}
        \caption{\texttt{rel-f1}, \textit{driver-position} (HGS)}
    \end{subfigure}
    \hfill
    \begin{subfigure}[b]{0.24\linewidth}
        \centering
        \includegraphics[width=\linewidth]{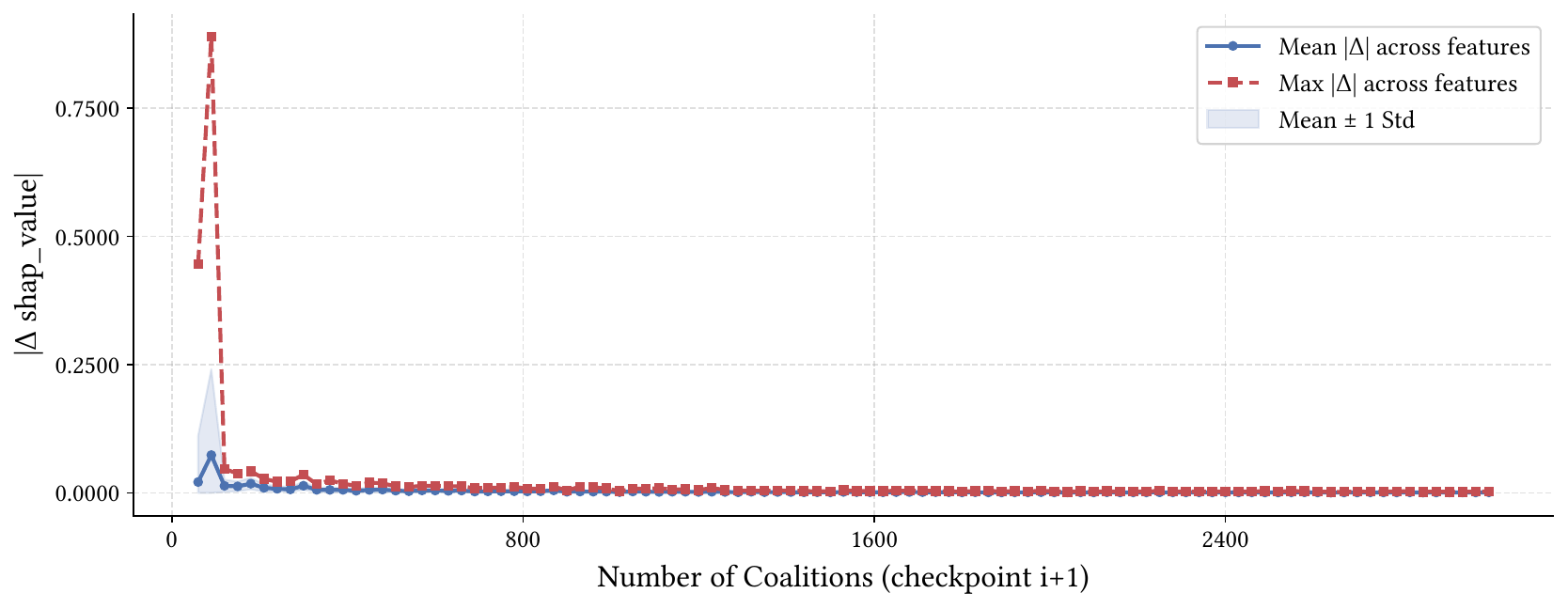}
        \caption{\texttt{rel-f1}, \textit{driver-position} (RelGT)}
    \end{subfigure}
    \hfill
    \begin{subfigure}[b]{0.24\linewidth}
        \centering
        \includegraphics[width=\linewidth]{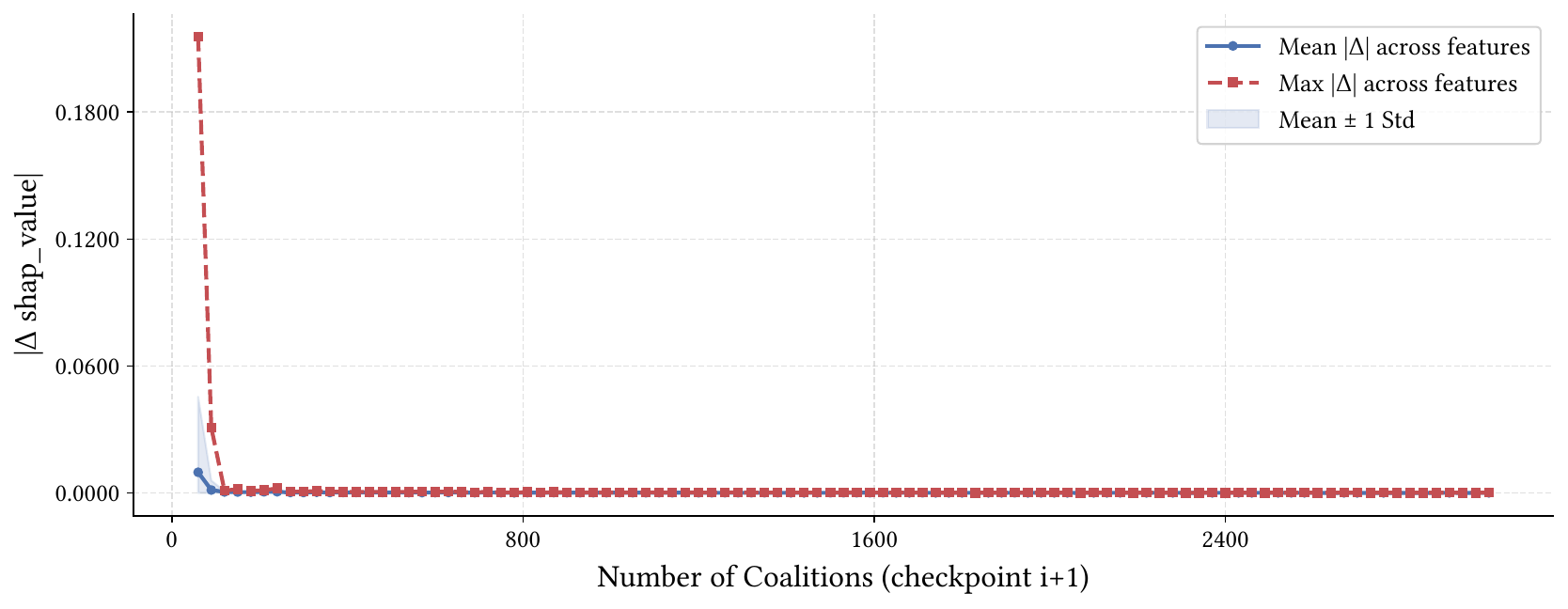}
        \caption{\texttt{rel-f1}, \textit{driver-dnf} (HGS)}
    \end{subfigure}
    \hfill
    \begin{subfigure}[b]{0.24\linewidth}
        \centering
        \includegraphics[width=\linewidth]{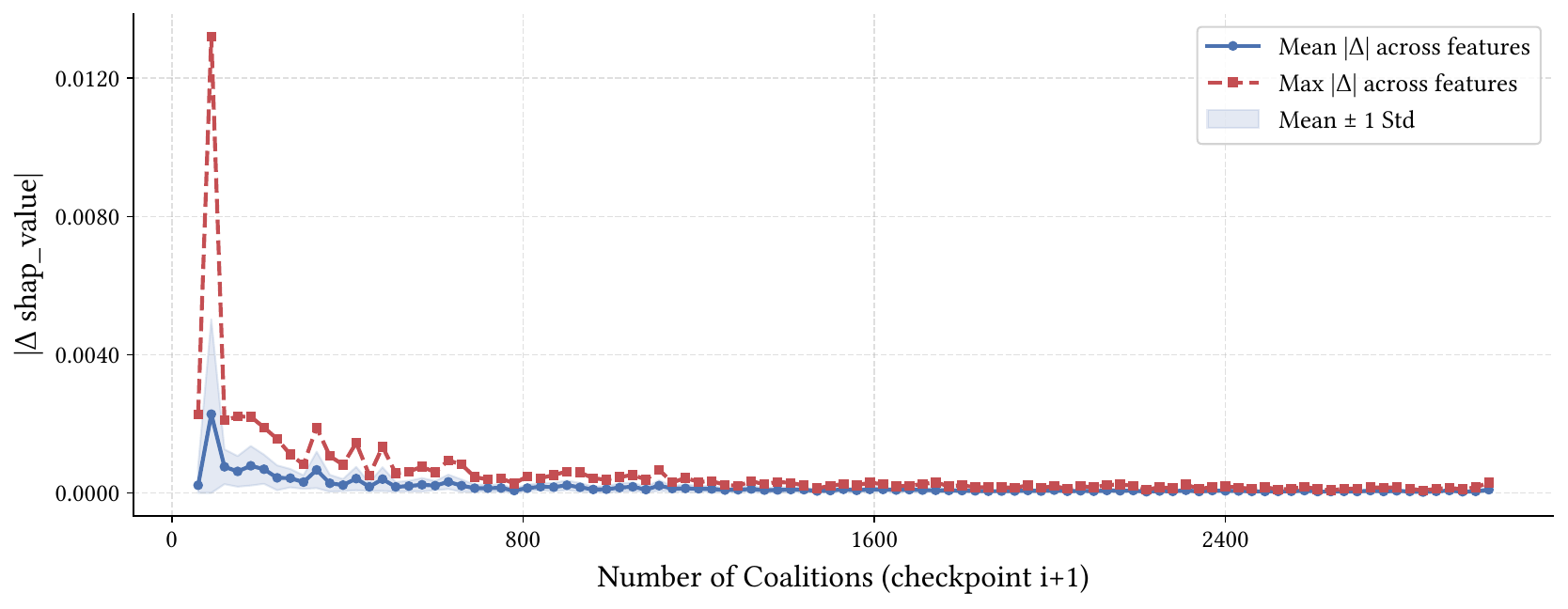}
        \caption{\texttt{rel-f1}, \textit{driver-dnf} (RelGT)}
    \end{subfigure}
    
    \vspace{1em}

    \begin{subfigure}[b]{0.24\linewidth}
        \centering
        \includegraphics[width=\linewidth]{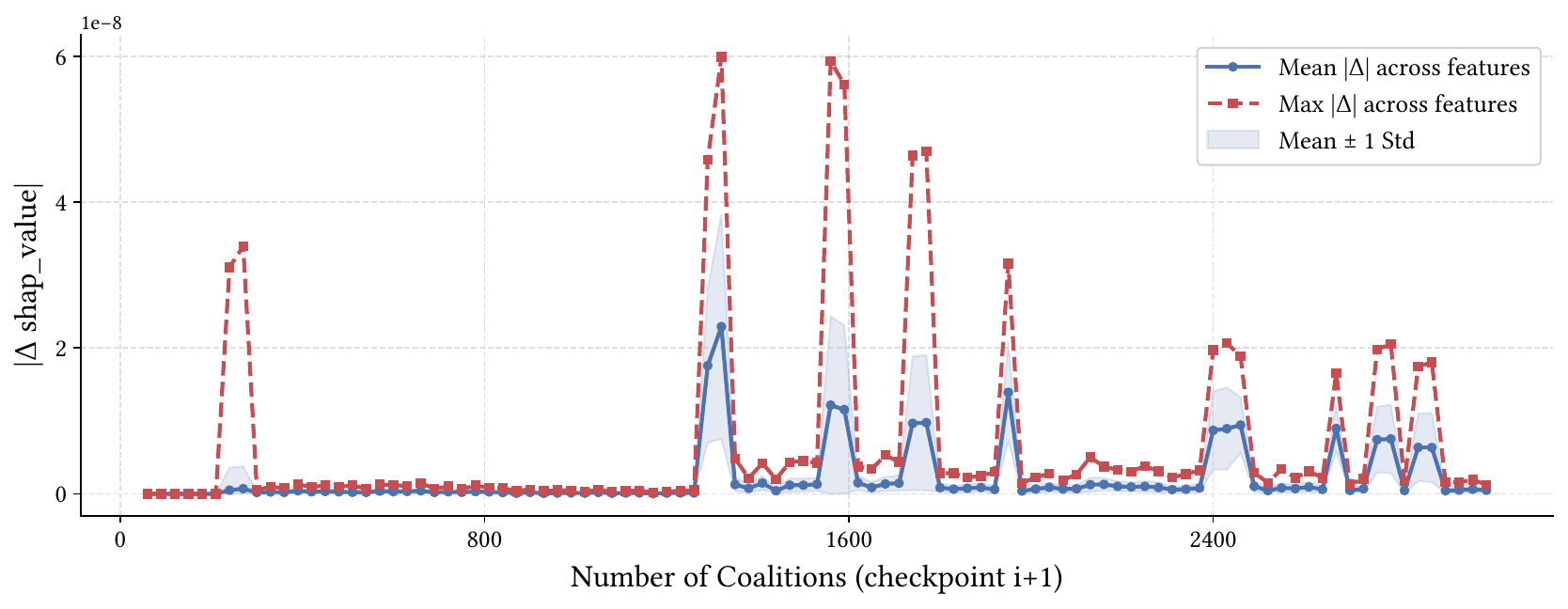}
        \caption{\texttt{rel-event}, \textit{user-attendance} (HGS)}
    \end{subfigure}
    \hfill
    \begin{subfigure}[b]{0.24\linewidth}
        \centering
        \includegraphics[width=\linewidth]{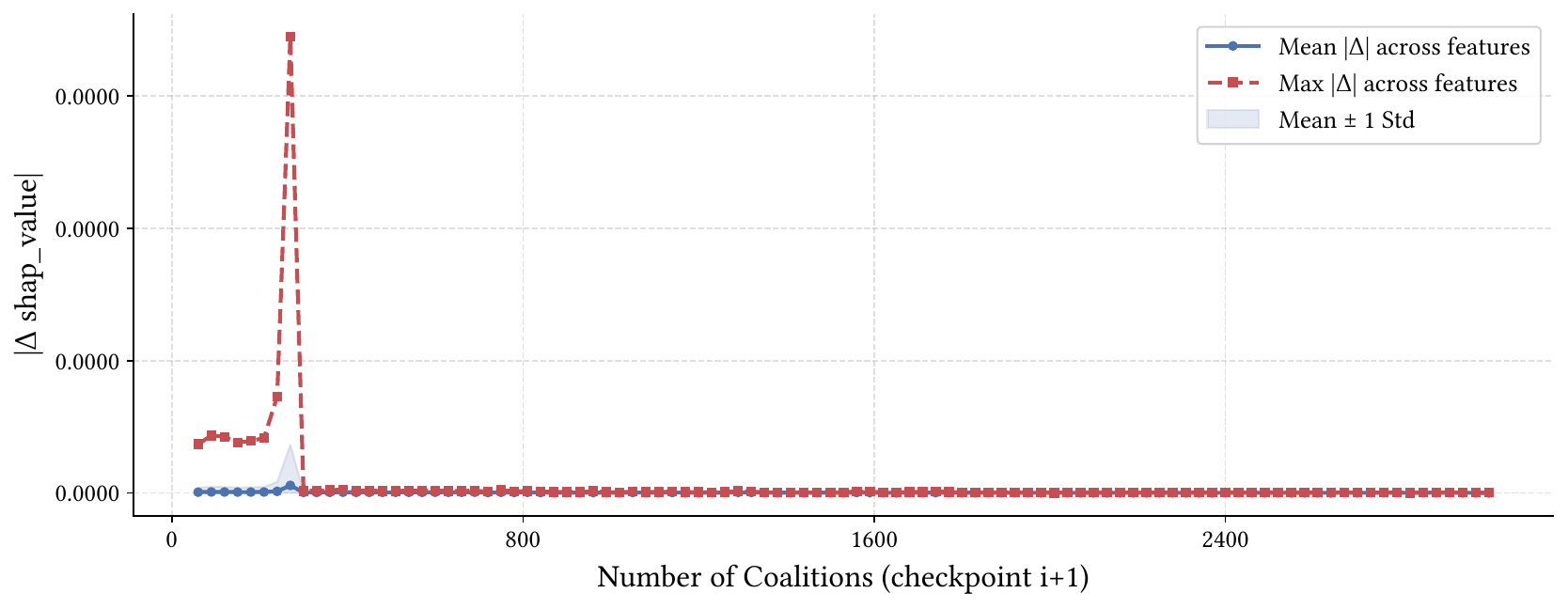}
        \caption{\texttt{rel-event}, \textit{user-attendance} (RelGT)}
    \end{subfigure}
    \hfill
    \begin{subfigure}[b]{0.24\linewidth}
        \centering
        \includegraphics[width=\linewidth]{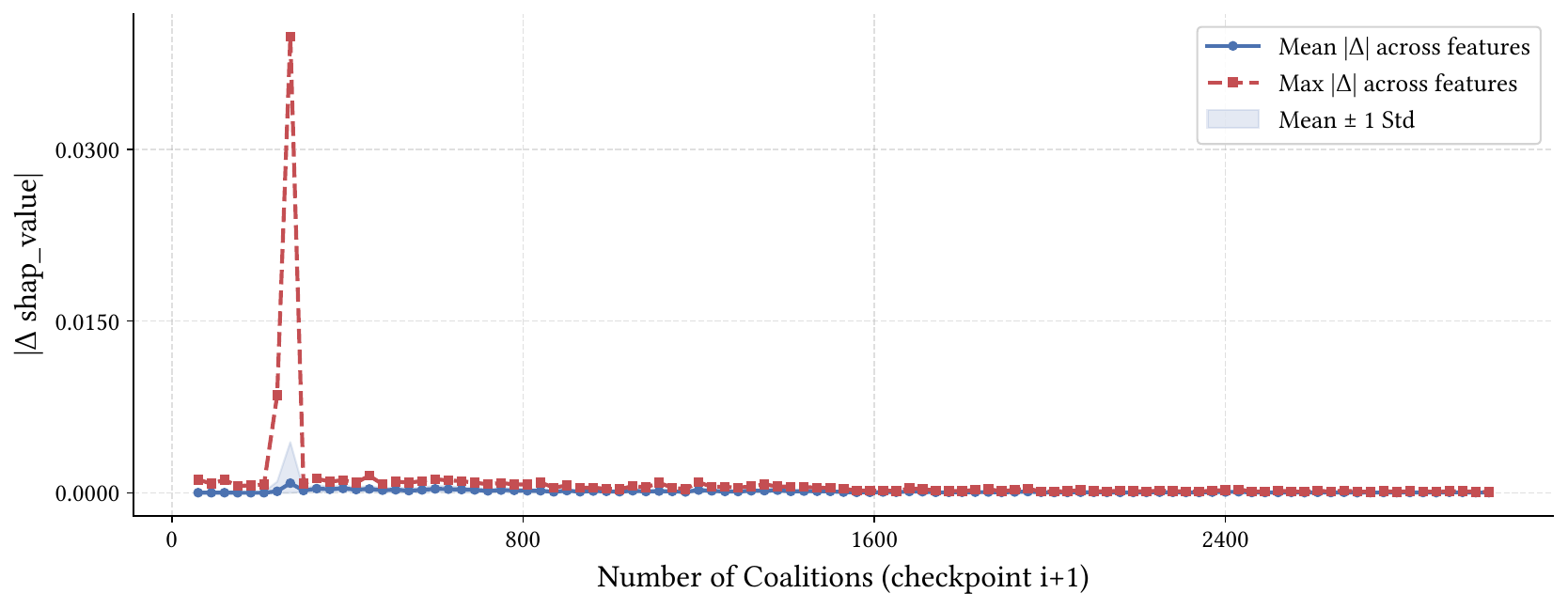}
        \caption{\texttt{rel-event}, \textit{user-repeat} (HGS)}
    \end{subfigure}
    \hfill
    \begin{subfigure}[b]{0.24\linewidth}
        \centering
        \includegraphics[width=\linewidth]{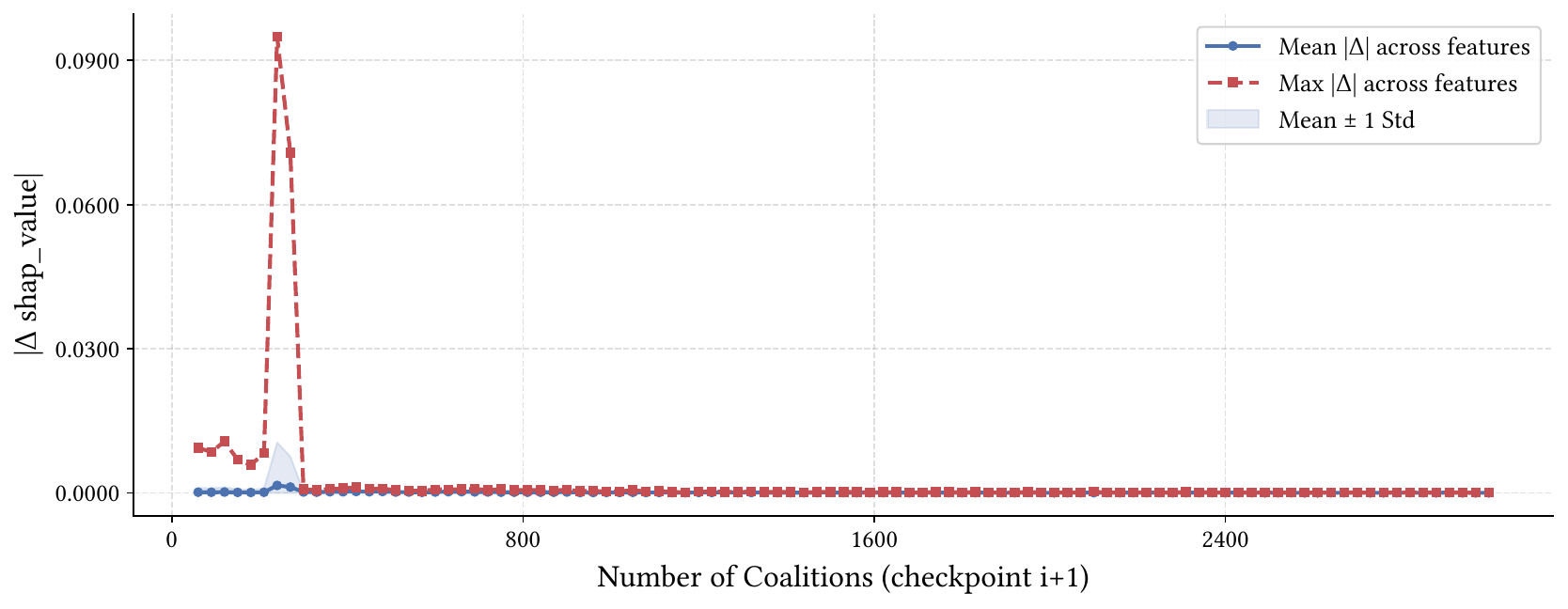}
        \caption{\texttt{rel-event}, \textit{user-repeat} (RelGT)}
    \end{subfigure}

    \caption{Empirical convergence curves for the Kernel SHAP approximation on the \texttt{rel-f1} and \texttt{rel-event} datasets.}
    \label{fig:app_kernel_convergence_grid_1}
\end{figure*}

\begin{figure*}[htbp]
    \centering
    
    \begin{subfigure}[b]{0.24\linewidth}
        \centering
        \includegraphics[width=\linewidth]{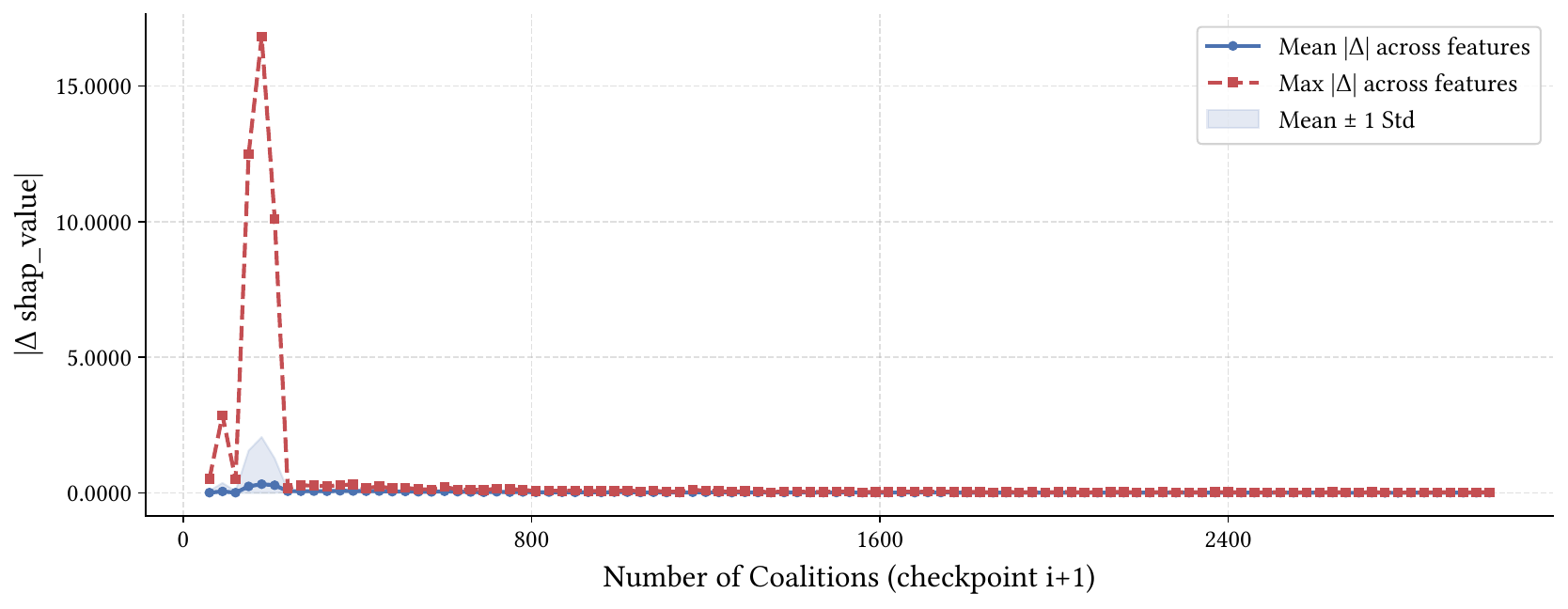}
        \caption{\texttt{rel-trial}, \textit{study-adverse} (HGS)}
    \end{subfigure}
    \hfill
    \begin{subfigure}[b]{0.24\linewidth}
        \centering
        \includegraphics[width=\linewidth]{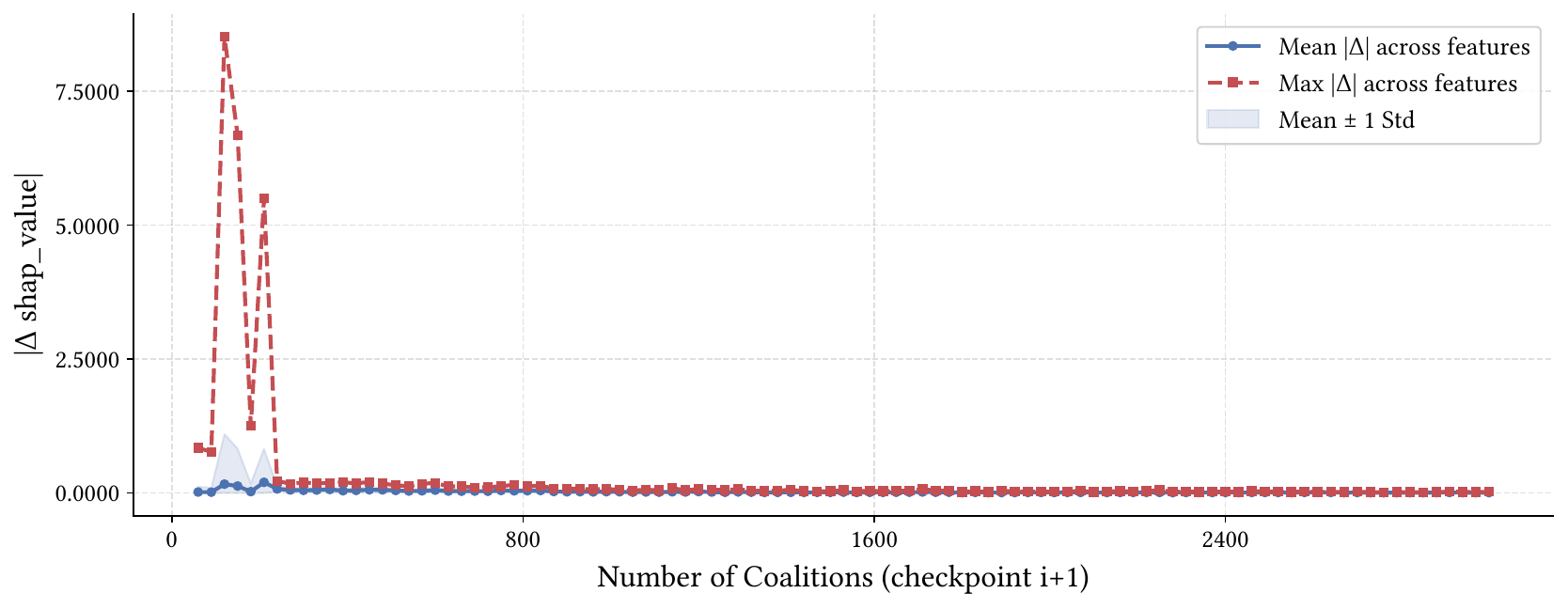}
        \caption{\texttt{rel-trial}, \textit{study-adverse} (RelGT)}
    \end{subfigure}
    \hfill
    \begin{subfigure}[b]{0.24\linewidth}
        \centering
        \includegraphics[width=\linewidth]{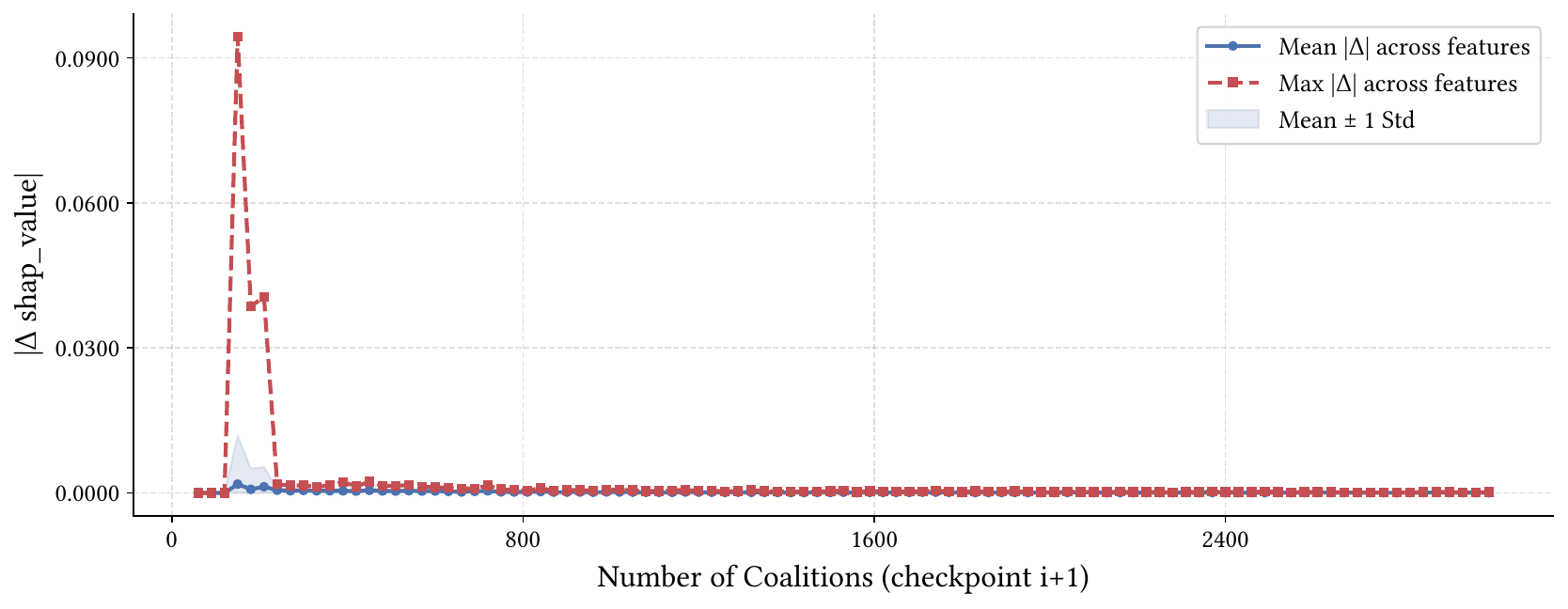}
        \caption{\texttt{rel-trial}, \textit{study-outcome} (HGS)}
    \end{subfigure}
    \hfill
    \begin{subfigure}[b]{0.24\linewidth}
        \centering
        \includegraphics[width=\linewidth]{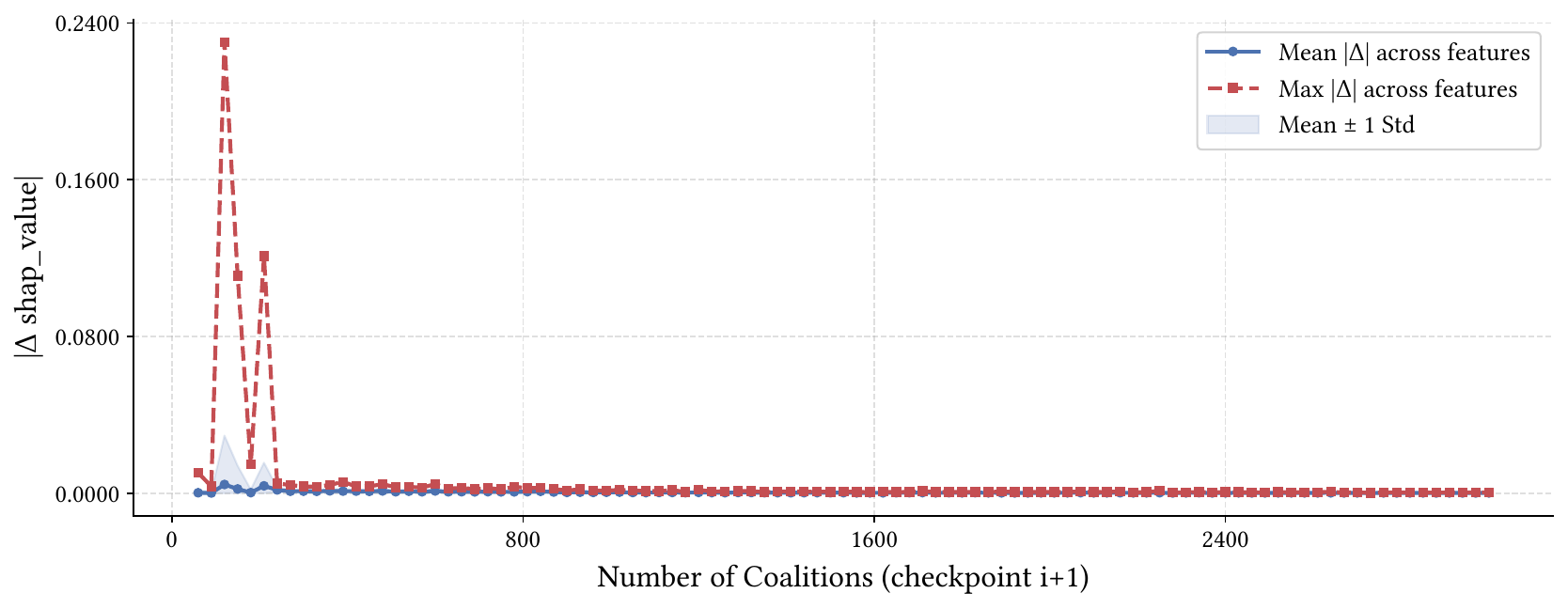}
        \caption{\texttt{rel-trial}, \textit{study-outcome} (RelGT)}
    \end{subfigure}
    
    \vspace{1em}

    \begin{subfigure}[b]{0.24\linewidth}
        \centering
        \includegraphics[width=\linewidth]{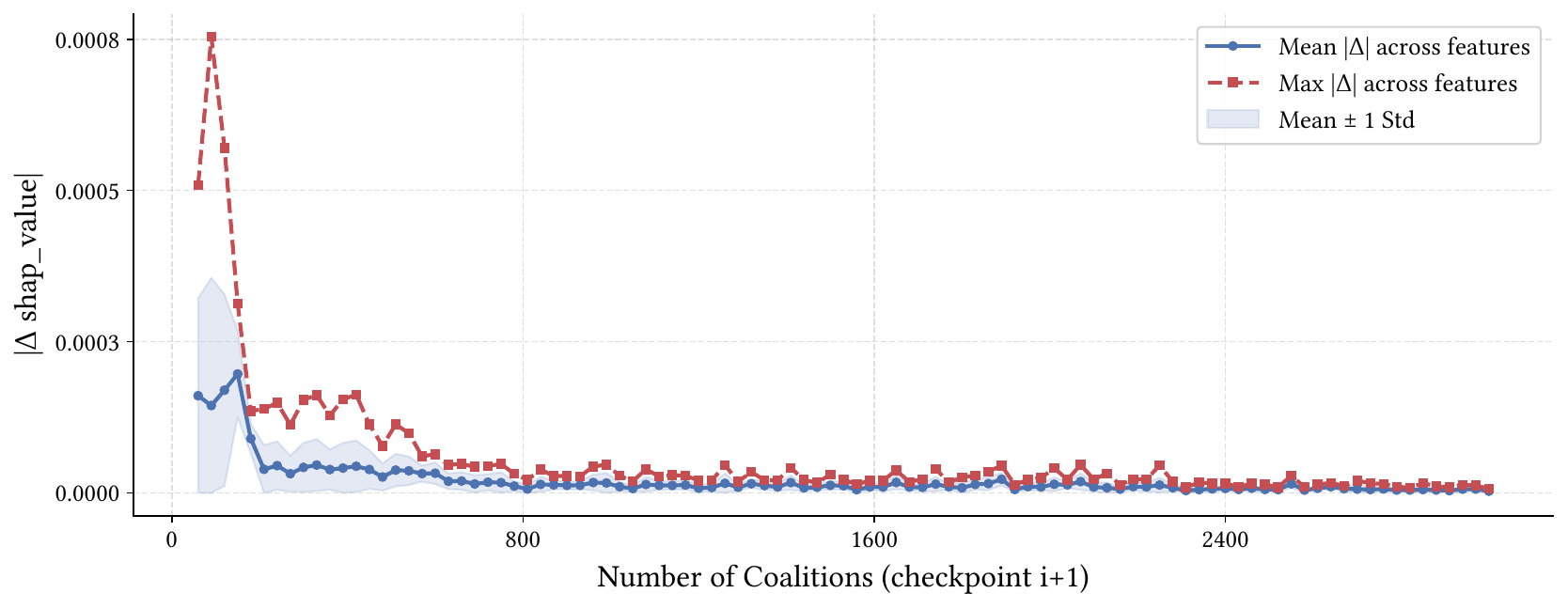}
        \caption{\texttt{rel-arxiv}, \textit{paper-citation} (HGS)}
    \end{subfigure}
    \hfill
    \begin{subfigure}[b]{0.24\linewidth}
        \centering
        \includegraphics[width=\linewidth]{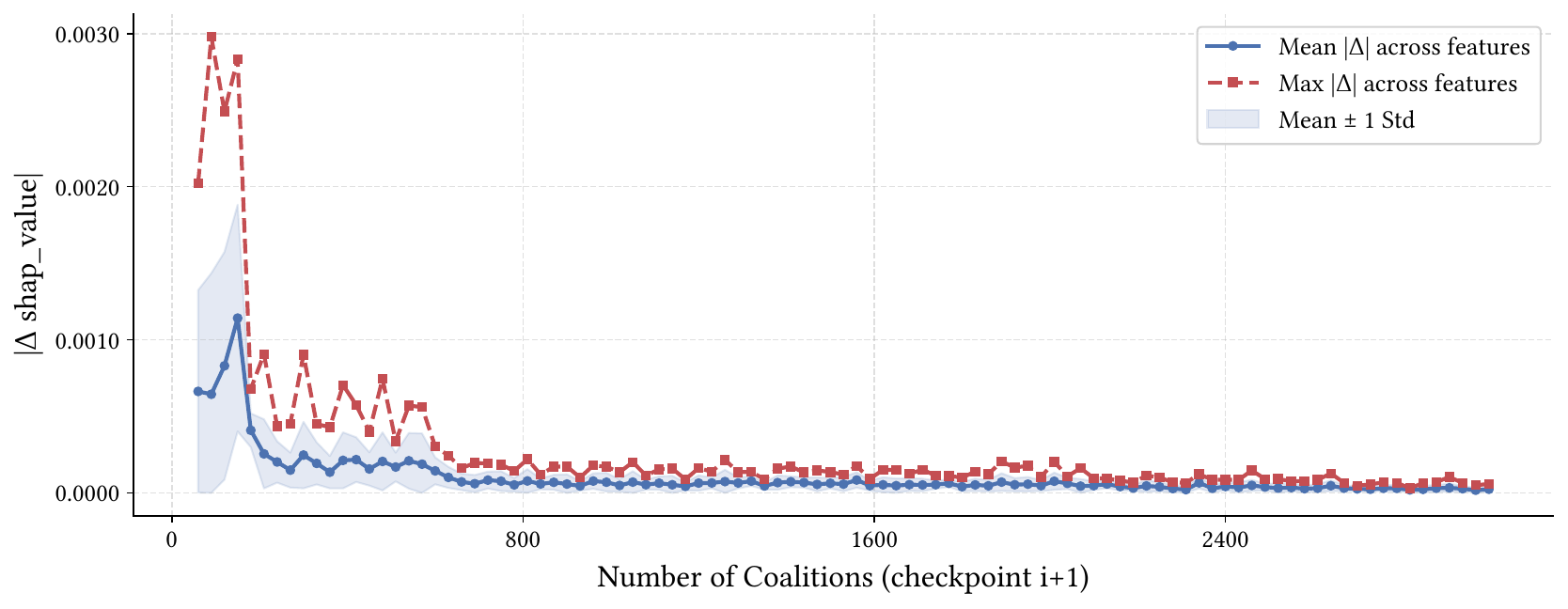}
        \caption{\texttt{rel-arxiv}, \textit{paper-citation} (RelGT)}
    \end{subfigure}
    \hfill
    \begin{subfigure}[b]{0.24\linewidth}
        \centering
        \includegraphics[width=\linewidth]{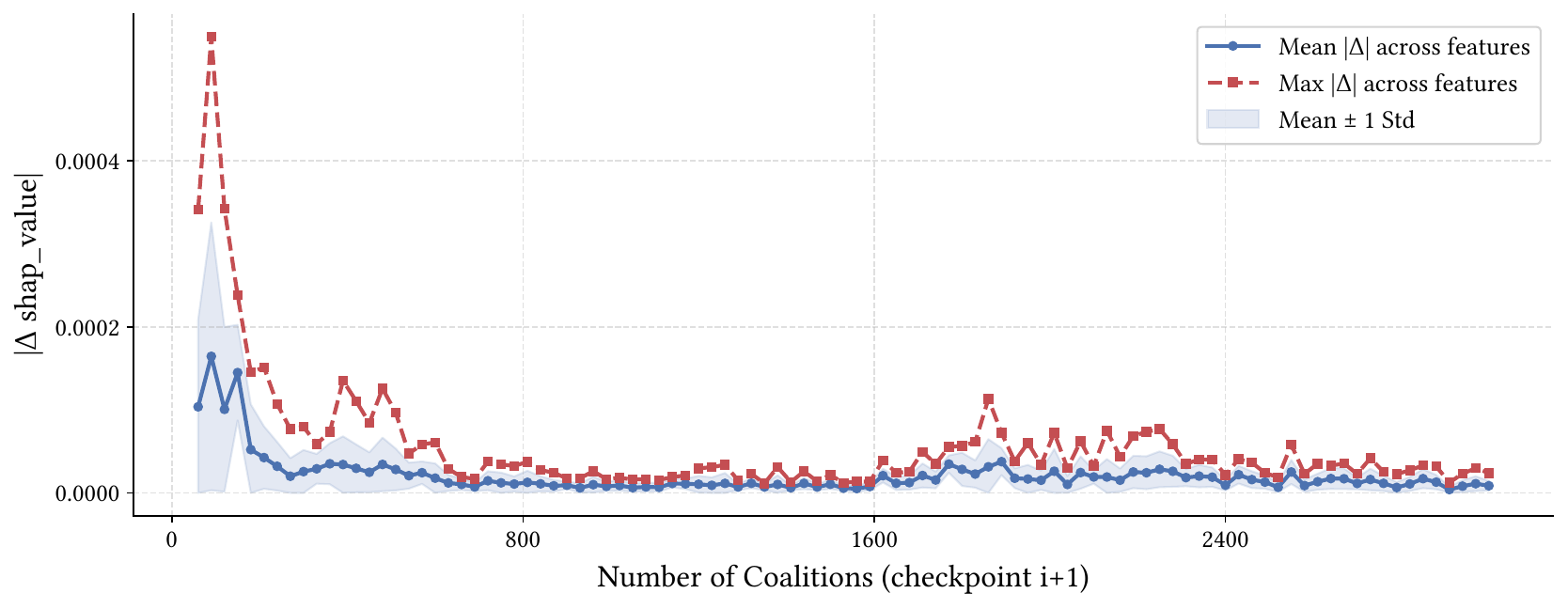}
        \caption{\texttt{rel-arxiv}, \textit{author-publication} (HGS)}
    \end{subfigure}
    \hfill
    \begin{subfigure}[b]{0.24\linewidth}
        \centering
        \includegraphics[width=\linewidth]{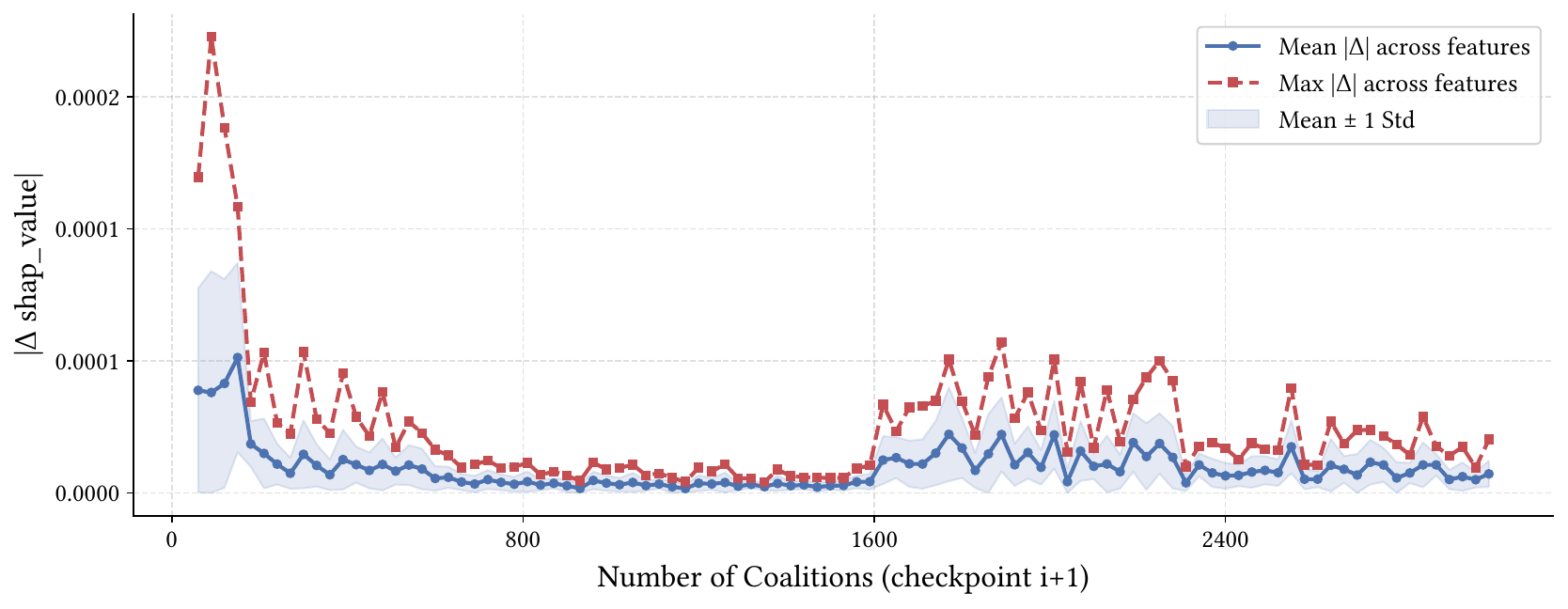}
        \caption{\texttt{rel-arxiv}, \textit{author-publication} (RelGT)}
    \end{subfigure}

    \caption{Empirical convergence curves for the Kernel SHAP approximation on the \texttt{rel-trial} and \texttt{rel-arxiv} datasets.}
    \label{fig:app_kernel_convergence_grid_2}
\end{figure*}

\Cref{fig:app_kernel_convergence_grid_1,fig:app_kernel_convergence_grid_2} present the empirical convergence curves for \kernel across all datasets and architectures. Similar to the Monte Carlo evaluation, each plot tracks the absolute difference in the estimated importance scores between consecutive sampling checkpoints ($|\Delta \text{shap\_value}|$). The stabilization of these curves confirms that the \kernel estimates have empirically converged within the allocated evaluation budget.

%% file: appendix_rt.tex
\section{Additional Architecture: Relational Transformer (RT)}\label{app:rt}

To verify that our findings generalize beyond GNN-based (HGS) and graph-transformer (RelGT) architectures, we replicate our experimental evaluation on a third relational deep learning model, the \textbf{Relational Transformer (RT)}~\cite{ranjan2026rtjlargescalepretrainingrelational}. We evaluate RT across all eight RelBench tasks using the three removal-based explainers (\single, \shapley, and \kernel), following the experimental protocol of \Cref{sec:experiments} with the model-specific adaptations detailed in \Cref{app:rt_setup}. We do not apply \mask to RT, as the masking intervention is specific to graph-based architectures. Overall, the results on RT closely corroborate the conclusions drawn from HGS and RelGT across all three research questions.

\subsection{Model Setup and Adaptations}\label{app:rt_setup}

Unlike HGS and RelGT, which operate over a heterogeneous tuple graph whose initial node embeddings are computed by a shared tuple-level feature encoder, RT represents the database as a collection of cell tokens. For each prediction, it samples a context of cells around the target entity, comprising the target row, foreign-key-linked rows, and earlier rows of the task table. We fine-tune the pretrained RT-J checkpoint on each task's training split and select the best model using early stopping on the validation split.

This difference in representation is also why we do not apply \mask to RT. For graph-based architectures, the masks are applied to the per-attribute embeddings within the tuple-level feature encoder, and the perturbation is defined tuple-wise~\cite{rissaki2025database}. RT's representation is not a graph, and its unit of representation is the individual cell rather than the tuple, which does not agree with our tuple-level masking. The removal-based explainers, in contrast, require only black-box access to the model and can thus be applied to RT with the adaptations described below.

To ensure comparability across architectures, we evaluate RT over the same attribute set and type-specific baseline statistics (mean, most frequent value, median, or mean embedding) used for HGS and RelGT. Because RT also observes earlier rows of the task table in its context, we expose the entity's historical target and timestamp columns as two additional \emph{task-history} attributes (visible in \Cref{fig:rt_consensus_grid}).

Since RT operates on sampled cell contexts rather than a fixed graph representation, we perform ablation by replacing the corresponding cell values directly inside each sampled context prior to the forward pass (leaving the target query row intact) and mapping the baseline statistics into RT's input encoding. To prevent context-sampling noise from affecting the attribution scores, we fix the random seed so that the original and all ablated predictions are evaluated on the exact same sampled context, using a single context sample per instance to keep \kernel and \shapley computationally feasible.

\subsection{Explainer Ranking Alignment ($Q_1$)}

\begin{figure*}[t]
    \centering
    \begin{subfigure}[b]{0.485\linewidth}
        \centering
        \includegraphics[width=\linewidth]{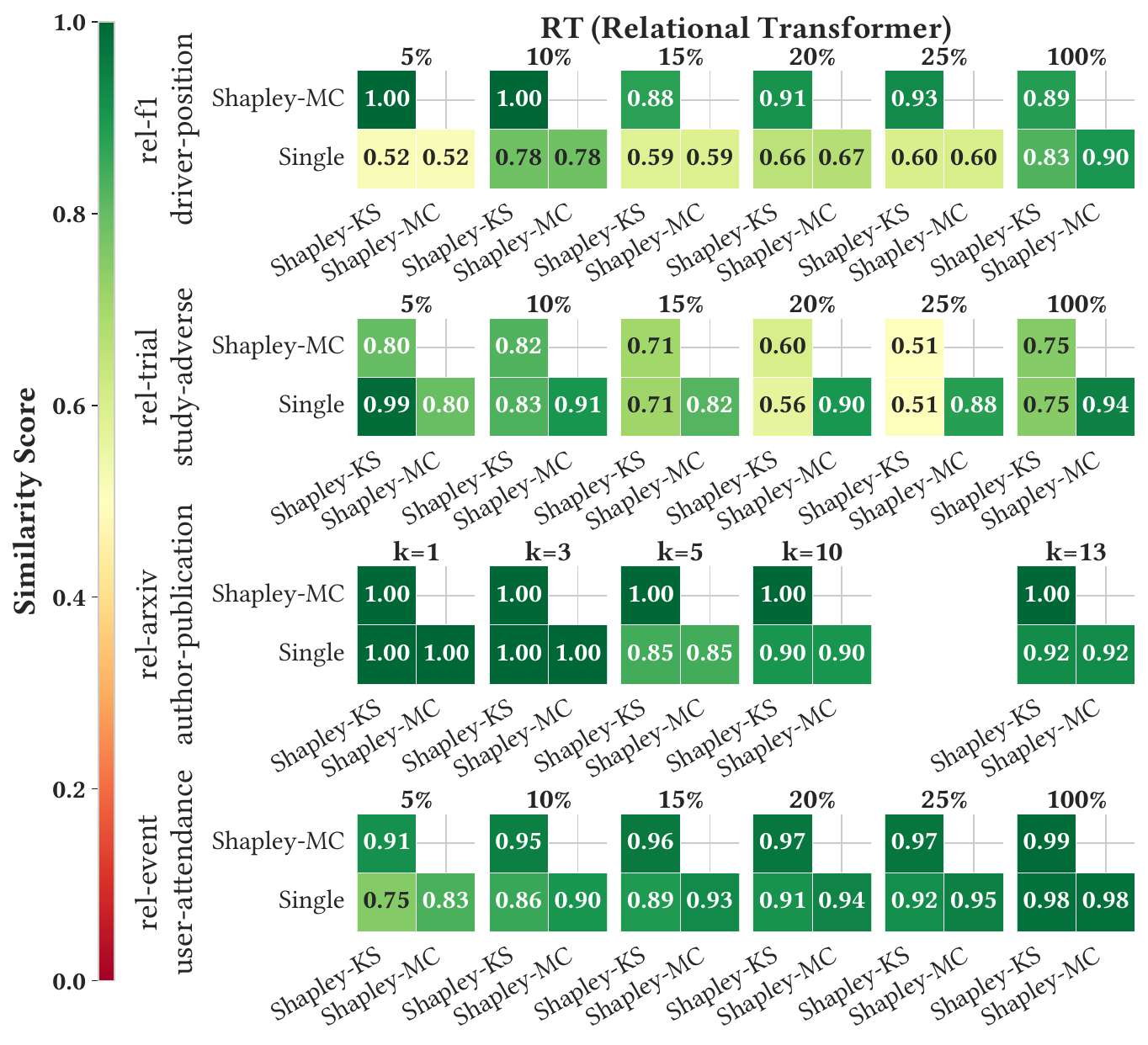}
        \caption{Regression tasks.}
        \label{fig:rt_heatmaps_regression}
    \end{subfigure}
    \hfill
    \begin{subfigure}[b]{0.485\linewidth}
        \centering
        \includegraphics[width=\linewidth]{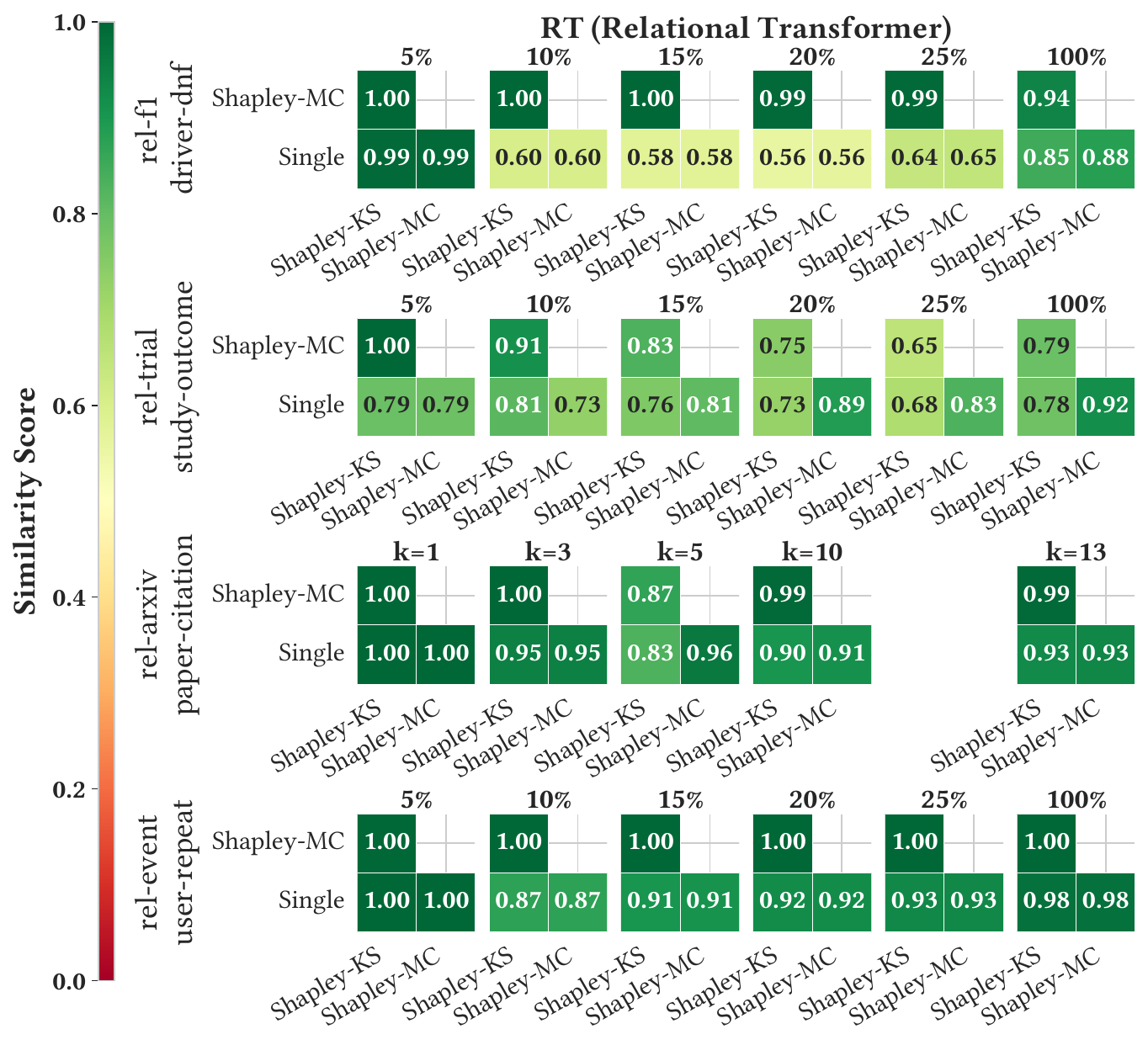}
        \caption{Classification tasks.}
        \label{fig:rt_heatmaps_classification}
    \end{subfigure}
    \caption{Pairwise explainer similarity for RT across regression and classification tasks over different prefixes ($k$).}
    \label{fig:rt_heatmaps}
\end{figure*}

\Cref{fig:rt_heatmaps} shows pairwise ranking similarity heatmaps for RT. Consistent with our main findings (\Cref{subsec:rq1}), \kernel and \shapley agree near-perfectly across most tasks and prefix sizes—e.g., reaching $1.00$ for all $k$ in \texttt{author-publication} and \texttt{user-repeat}, and $0.91$--$0.99$ in \texttt{user-attendance}. 
In RT, the most notable divergences between \kernel and \shapley at intermediate $k$ values appear in \texttt{study-adverse} and \texttt{study-outcome}, consistent with our broader observation in \Cref{subsec:rq1} that the two Shapley approximations can diverge as $k$ increases due to their distinct estimation mechanisms.

Similarly, \single aligns closely with both Shapley methods in high-consensus tasks such as \texttt{user-repeat} ($0.87$--$1.00$), \texttt{author-publication} ($0.85$--$1.00$), \texttt{user-attendance} ($0.75$--$0.98$), and \texttt{paper-citation} ($0.83$--$1.00$), while showing lower similarity at intermediate $k$ in complex tasks such as \texttt{driver-dnf} and \texttt{driver-position} before converging at larger $k$.

\subsection{Explainer Score Distribution ($Q_2$)}

\begin{figure*}[t]
    \centering
    \begin{subfigure}[b]{0.48\linewidth}
        \centering
        \includegraphics[width=\linewidth]{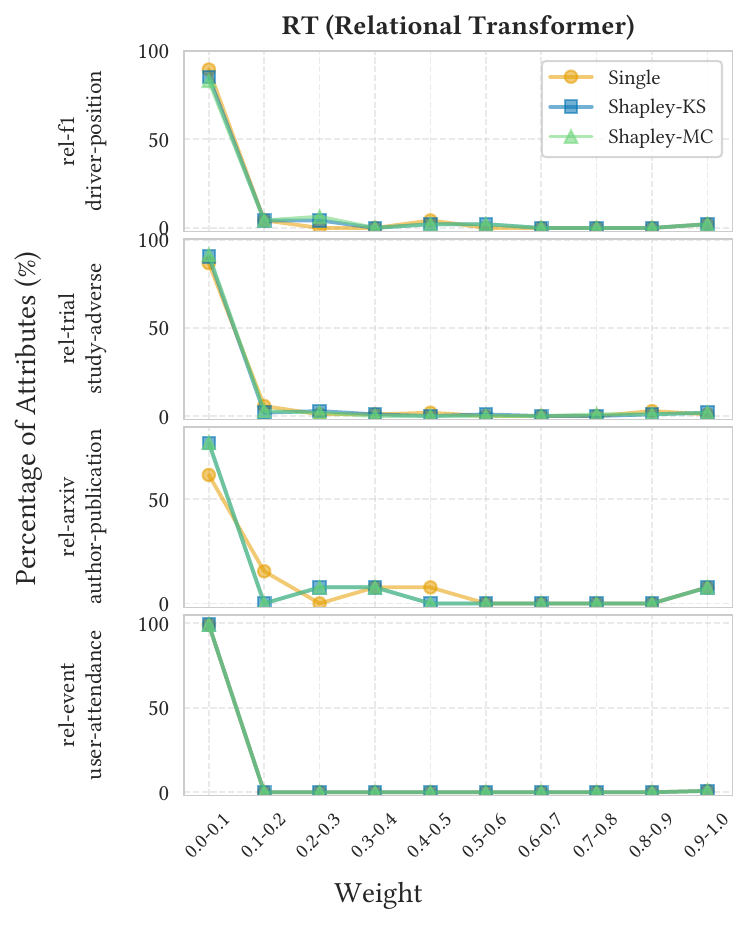}
        \caption{Regression tasks.}
        \label{fig:rt_dist_regression}
    \end{subfigure}
    \hfill
    \begin{subfigure}[b]{0.48\linewidth}
        \centering
        \includegraphics[width=\linewidth]{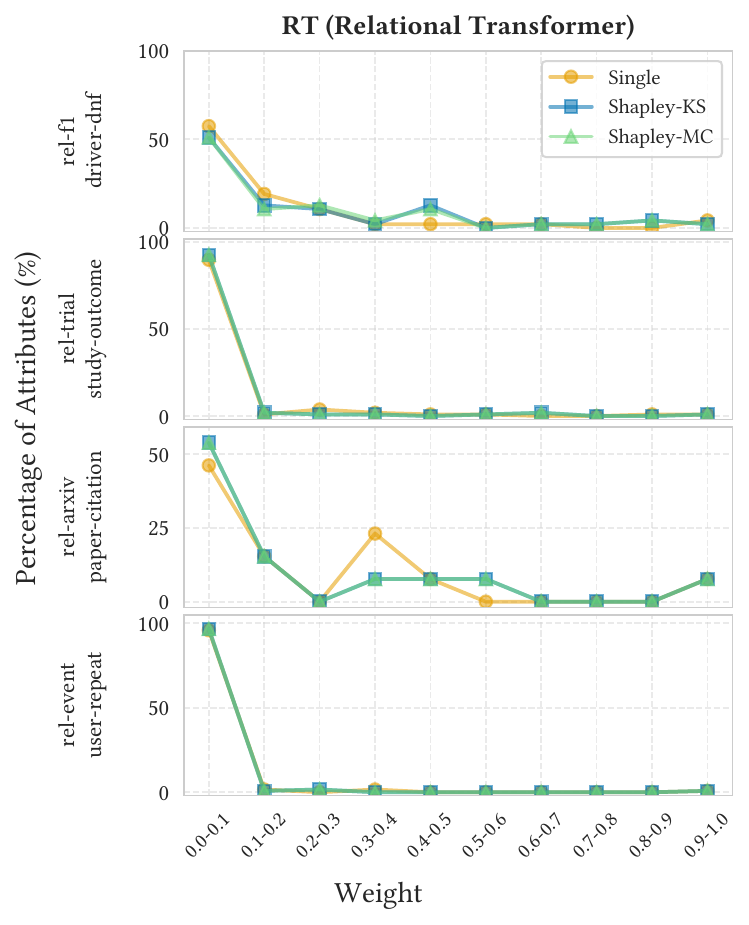}
        \caption{Classification tasks.}
        \label{fig:rt_dist_classification}
    \end{subfigure}
    \caption{Score distribution (PDF) of normalized attribution weights for RT.}
    \label{fig:rt_distribution_grid}
\end{figure*}

\begin{figure*}[t]
    \centering
    \begin{subfigure}[b]{0.48\linewidth}
        \centering
        \includegraphics[width=\linewidth]{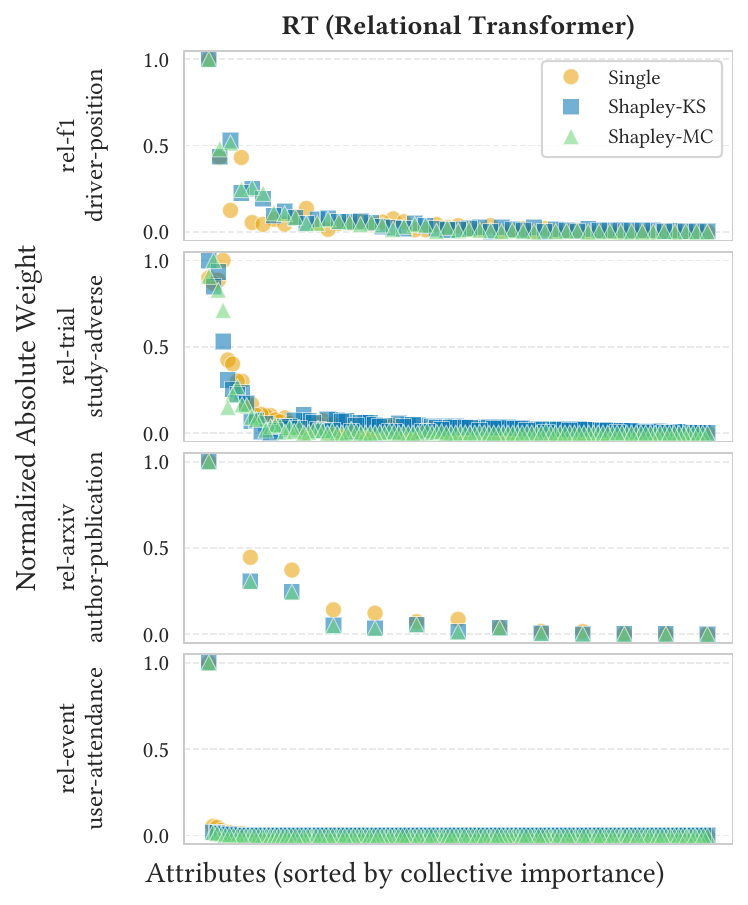}
        \caption{Regression tasks.}
        \label{fig:rt_scatter_regression}
    \end{subfigure}
    \hfill
    \begin{subfigure}[b]{0.48\linewidth}
        \centering
        \includegraphics[width=\linewidth]{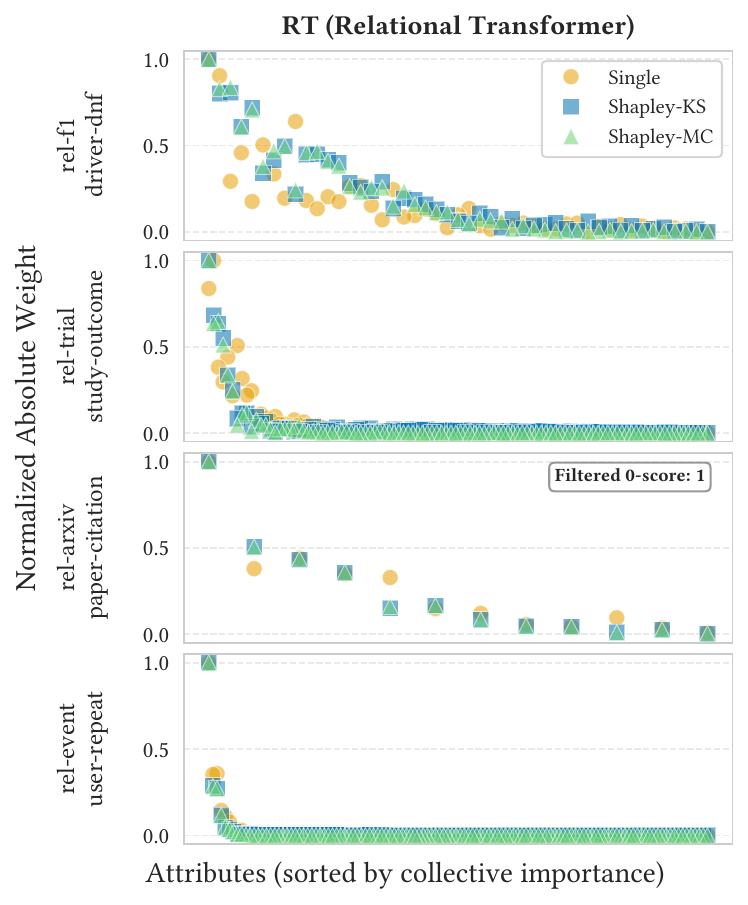}
        \caption{Classification tasks.}
        \label{fig:rt_scatter_classification}
    \end{subfigure}
    \caption{Normalized score scatter grid for RT. Attributes along the x-axis are sorted by collective importance.}
    \label{fig:rt_scatter_grid}
\end{figure*}

\begin{figure*}[t]
    \centering
    \begin{subfigure}[b]{0.485\linewidth}
        \centering
        \includegraphics[width=\linewidth]{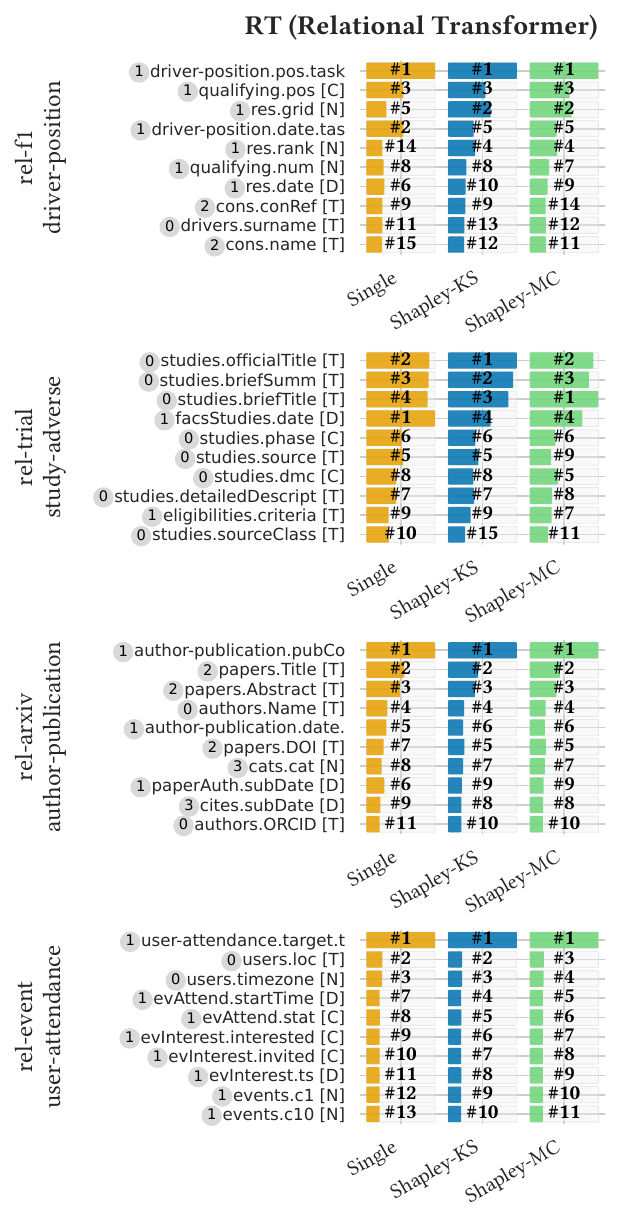}
        \caption{Regression tasks.}
        \label{fig:rt_consensus_regression}
    \end{subfigure}
    \hfill
    \begin{subfigure}[b]{0.485\linewidth}
        \centering
        \includegraphics[width=\linewidth]{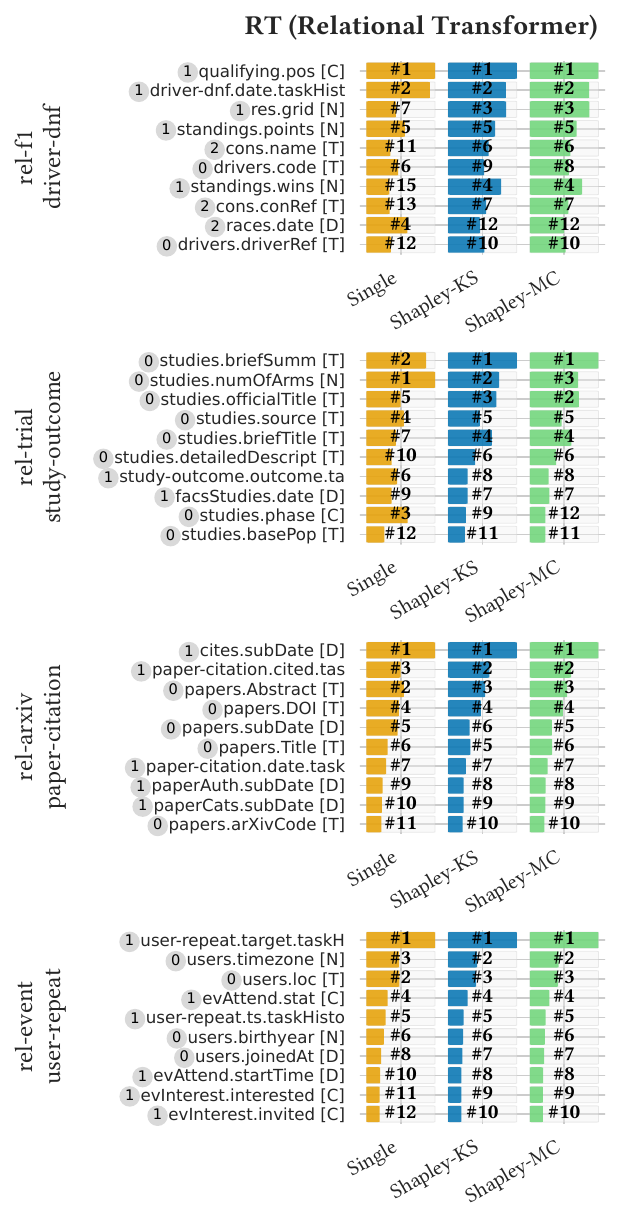}
        \caption{Classification tasks.}
        \label{fig:rt_consensus_classification}
    \end{subfigure}
    \caption{Consensus min-max normalized score grids for RT (top-15 evaluation, displaying top 10 consensus attributes after filtering out attributes with normalized scores below $0.001$ across all explainers).}
    \label{fig:rt_consensus_grid}
\end{figure*}

The score distribution PDFs (\Cref{fig:rt_distribution_grid}) and scatter plots (\Cref{fig:rt_scatter_grid}) mirror the distributional patterns observed in \Cref{subsec:rq2}. Across both task types, all three explainers concentrate the vast majority of attributes in the lowest weight bin ($0.0$--$0.1$) and assign high importance to only a small subset of attributes. In regression tasks (\Cref{fig:rt_scatter_regression}), the explainers track closely together along a steep decay curve, whereas in classification tasks—most notably \texttt{driver-dnf} (\Cref{fig:rt_scatter_classification})-the normalized scores are more dispersed across the top attributes.

The consensus grids (\Cref{fig:rt_consensus_grid}) further confirm strong agreement among the removal-based explainers on the leading attributes (with unanimous agreement on the top-ranked attribute in nearly every task). As with HGS and RelGT, RT draws its most influential attributes from across multiple tables and hop distances—combining $0$-hop target attributes with $1$-, $2$-, and (in \texttt{author-publication}) $3$-hop relational attributes, alongside historical target signals.

\subsection{Fidelity Metrics Evaluation ($Q_3$)}

\begin{figure*}[t]
    \centering
    \includegraphics[width=0.75\linewidth]{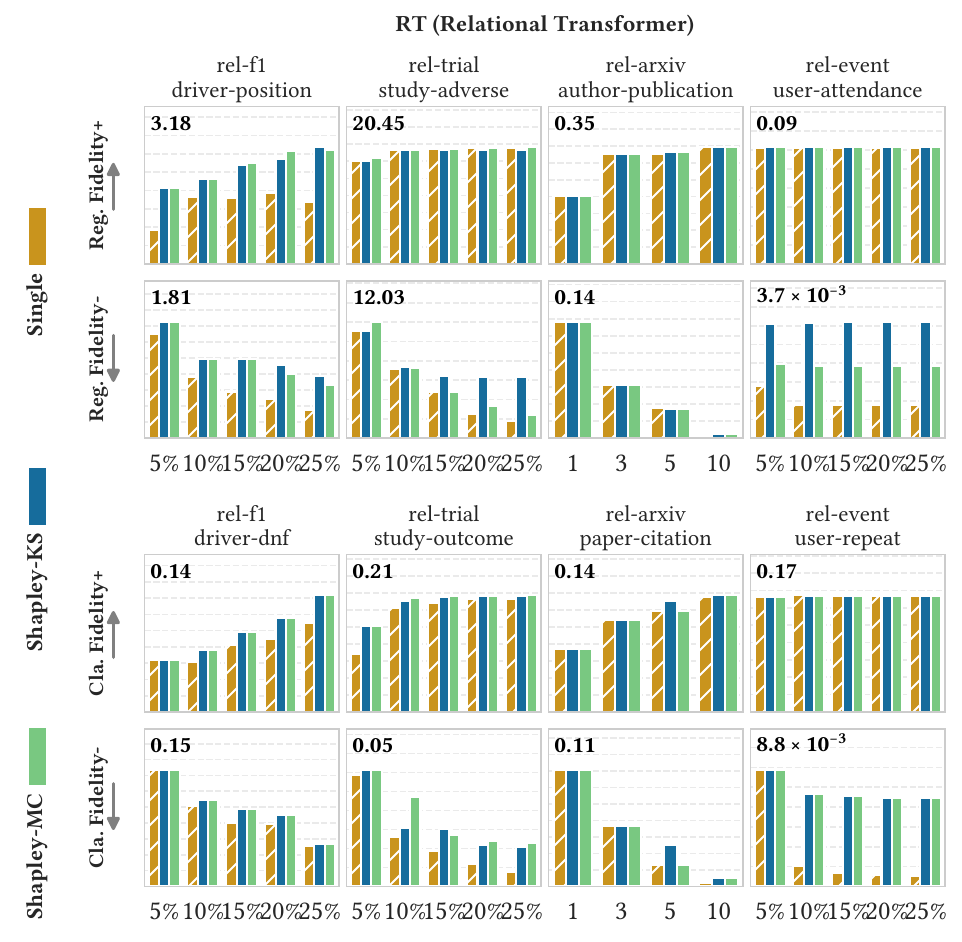}
    \caption{Explainer fidelity for RT in regression (top 2 rows) and classification (bottom 2 rows) tasks. Bars represent $\text{Fidelity}^+$ (necessity, higher is better) and $\text{Fidelity}^-$ (sufficiency, lower is better) across varying top-$k$ attribute subsets.}
    \label{fig:rt_fidelity}
\end{figure*}

\Cref{fig:rt_fidelity} shows the $\text{Fidelity}^+$ and $\text{Fidelity}^-$ results for RT. Consistent with \Cref{subsec:rq3}, fidelity scores broadly reflect the ranking alignment: in tasks with high explainer consensus (such as \texttt{author-publication} and \texttt{paper-citation}), all three methods exhibit closely matched fidelity profiles, whereas in tasks where \single diverges in ranking (such as \texttt{driver-position} and \texttt{driver-dnf}), its fidelity values differ accordingly from \kernel and \shapley. At the same time, cases such as \texttt{user-repeat} and \texttt{user-attendance} $\text{Fidelity}^-$, where \single achieves noticeably lower sufficiency error than \kernel despite strong ranking alignment, reinforce our main conclusion that fidelity evaluations alone do not capture the full picture and should be paired with direct ranking and distributional comparisons.